\documentclass[preprint]{elsarticle}
\usepackage[margin=1in]{geometry}

\usepackage{hyperref}

\journal{arXiv.org}

\biboptions{authoryear}
\let\cite\citep

\usepackage[caption=false,font=footnotesize]{subfig}
\usepackage{threeparttable} 
\usepackage{multirow}
\usepackage{calc} 
\usepackage{amsmath}
\usepackage{bm} 
\usepackage{amsfonts}
\usepackage{xfrac}
\usepackage{mathtools}
\usepackage{textgreek}
\usepackage{dsfont}
\usepackage{booktabs}
\usepackage{xcolor}
\usepackage{siunitx}
\definecolor{lightgray}{gray}{0.95}
\usepackage[ruled,vlined]{algorithm2e}
\SetAlFnt{\small}
\makeatletter
\newcommand{\algrule}[1][.01pt]{\par\vskip.5\baselineskip\hrule height #1\par\vskip.5\baselineskip}
\makeatother
\makeatletter
\newcommand{\StatexIndent}[1][3]{
  \setlength\@tempdima{\algorithmicindent}
  \Statex\hskip\dimexpr#1\@tempdima\relax}
\makeatother

\DeclarePairedDelimiter{\ceil}{\lceil}{\rceil}
\DeclareMathOperator*{\argmax}{arg\,max}
\DeclareMathOperator*{\argmin}{arg\,min}
\DeclareMathOperator*{\median}{median}
\DeclarePairedDelimiter\abs{\lvert}{\rvert}
\DeclarePairedDelimiter\norm{\lVert}{\rVert}

\newcommand{\ARTa}{ART\textsubscript{a}}
\newcommand{\ARTb}{ART\textsubscript{b}}
\newcommand{\Fin}{F\textsubscript{1}}
\newcommand{\Fout}{F\textsubscript{2}}
\newcommand{\Foa}{$\text{F}_\text{1}^\text{a}$}
\newcommand{\Fta}{$\text{F}_\text{2}^\text{a}$}

\newcommand{\Ftb}{$\text{F}_\text{2}^\text{b}$}
\newcommand{\Fab}{F\textsuperscript{ab}}

\begin{document}

\begin{frontmatter}

\title{A Survey of Adaptive Resonance Theory Neural Network Models for Engineering Applications}

\author[address1,address2]{Leonardo Enzo Brito da Silva\corref{mycorrespondingauthor}}
\cortext[mycorrespondingauthor]{Corresponding author}
\ead{leonardoenzo@ieee.org}

\author[address1]{Islam Elnabarawy}

\author[address1]{Donald C. Wunsch II}

\address[address1]{Applied Computational Intelligence Laboratory, Missouri University of Science and Technology, Rolla, MO 65409, USA.}

\address[address2]{CAPES Foundation, Ministry of Education of Brazil, Bras\'{i}lia 70040-020, Brazil.}

\begin{abstract}
This survey samples from the ever-growing family of adaptive resonance theory (ART) neural network models used to perform the three primary machine learning modalities, namely, unsupervised, supervised and reinforcement learning.  It comprises a representative list from classic to modern ART models, thereby painting a general picture of the architectures developed by researchers over the past 30 years. The learning dynamics of these ART models are briefly described, and their distinctive characteristics such as code representation, long-term memory and corresponding geometric interpretation are discussed. Useful engineering properties of ART (speed, configurability, explainability, parallelization and hardware implementation) are examined along with current challenges. Finally, a compilation of online software libraries is provided. It is expected that this overview will be helpful to new and seasoned ART researchers.
\end{abstract}


\end{frontmatter}

\tableofcontents

\section{Introduction}

Adaptive Resonance Theory (ART)~\cite{Grossberg1976a, Grossberg1976b, Grossberg1982, Grossberg2013} is a biologically-plausible theory of how a brain learns to consciously attend, learn and recognize patterns in a constantly changing environment. The theory states that resonance regulates learning in neural networks with feedback (recurrence). Thus, it is more than a neural network architecture, or even a family of architectures. However, it has inspired many  neural network architectures that have very attractive properties for applications in science and engineering, such as being fast and stable incremental learners with relatively small memory requirements and straightforward algorithms~\cite{wunsch2009}. In this context, fast learning refers to the ability of the neurons' weight vectors to converge to their asymptotic values directly with each input sample presentation. These, and other properties, make ART networks attractive to many researchers and practitioners, as they have been used successfully in a variety of science and engineering applications.

ART addresses the problem of \textit{stability vs. plasticity}~\cite{Grossberg1982,Carpenter1987}. Plasticity refers the ability of a learning algorithm to adapt and learn new patterns. In many such learning systems plasticity can lead to instability, a situation in which learning new knowledge leads to the loss or corruption of previously-learned knowledge, also known as catastrophic forgetting. Stability, on the other hand, is defined by the condition that no prototype vector can take on a previous value after it has changed, and that an infinite presentation of inputs results in forming a finite number of clusters~\cite{xu2009,moore1989}. ART addresses this stability-plasticity dilemma by introducing the ability to learn arbitrary input patterns in a fast and stable self-organizing fashion without suffering from catastrophic forgetting. 

There have been some previous studies with similar objectives of surveying the ART neural network literature~\cite{Du.2010a, Amorim.2011a, Jain.2014a, RamaKrishna.2014a}. This survey expands on those works, compiling a broad and informative sampling of ART neural network architectures from the ever-growing machine learning literature. It captures a representative set of examples of various ART architectures in the unsupervised, supervised and reinforcement learning modalities, as well as some models that cross these boundaries and/or combine multiple learning modalities. The overarching goal of this survey is to provide researchers with an accessible coverage of these models, with a focus on their motivations, interpretations for engineering applications and a discussion of open problems for consideration. It is not meant as a comparative assessment of these models but rather a roadmap to assess options.

The remainder of this paper is organized as follows. Section~\ref{Sec:UL} presents a sampling of unsupervised learning (UL) ART models, divided into elementary, topological, hierarchical, biclustering and data fusion architectures. Section~\ref{Sec:SL} discusses supervised learning (SL) ART models for both classification and regression. Reinforcement learning (RL) ART models are discussed in Section~\ref{Sec:RL}. Sections~\ref{Sec:Advantages} and \ref{Sec:open_problems} discuss some of the useful properties of ART architectures and open problems in this field, respectively. Section~\ref{Sec:Resources} provides links to some repositories of ART neural network code, and Section~\ref{Sec:Conclusion} concludes the paper.

\section{ART models for unsupervised learning}
\label{Sec:UL}

\subsection{Elementary architectures}
\label{Sec:UL_units}

At their core, the elementary ART models are predominantly used for unsupervised learning applications. However, they also lay the foundation to build complex ART-based systems capable of performing all three machine learning modalities (Secs.~\ref{Sec:UL}, \ref{Sec:SL}, and \ref{Sec:RL}). This section describes the main characteristics of ART family members in terms of their code representation, long-term memory unit, system dynamics (which encompasses activation, match, resonance and learning) and user-defined parameters. For clarity, Table~\ref{Tab:Notation} summarizes the common notation used in the following subsections. 

\begin{figure}[!b]
\centerline{
\includegraphics[width=0.8\textwidth]{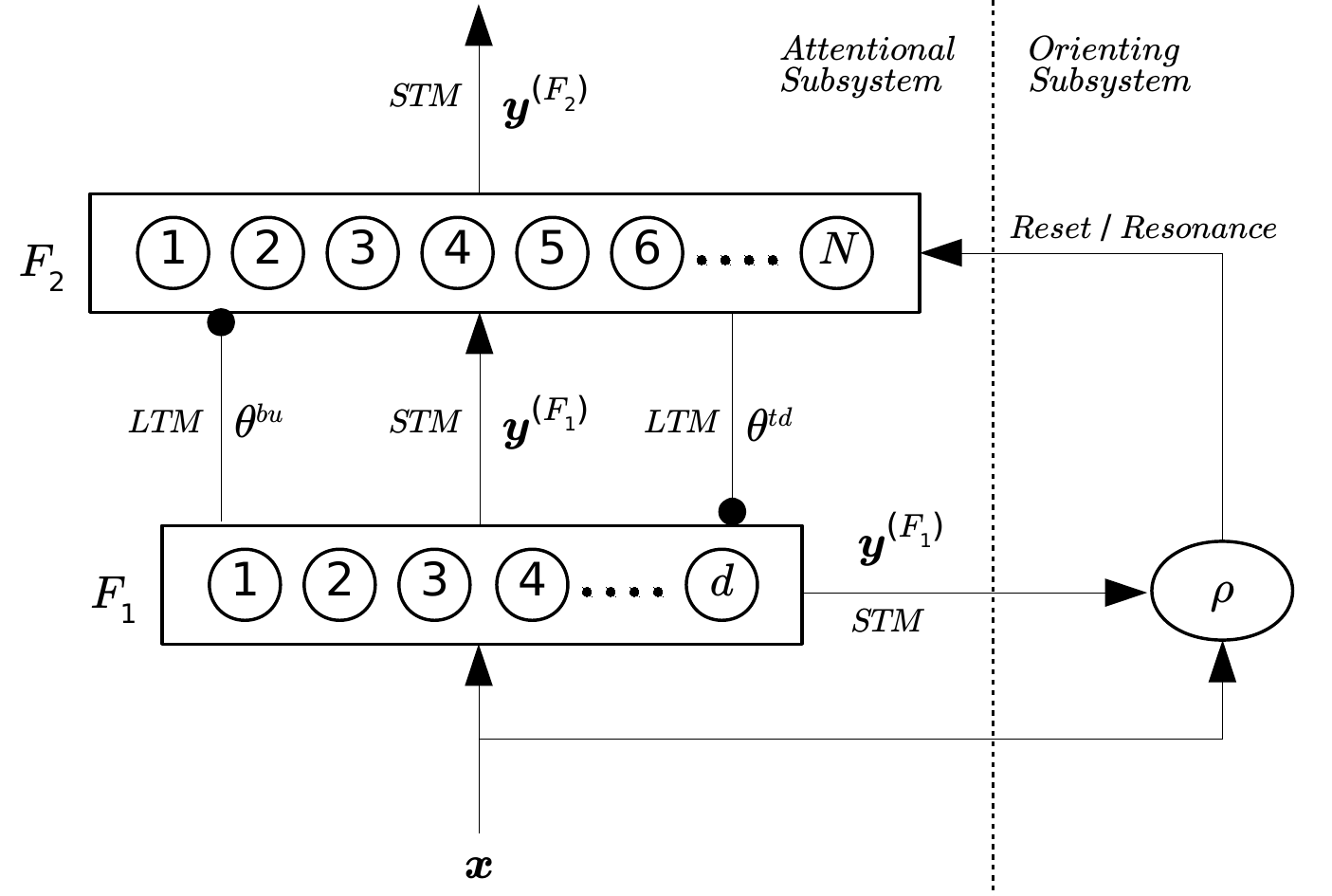}}
\caption{Elementary ART model underlying various designs. The orienting subsystem uses the vigilance threshold to regulate whether ART can go into resonance or if it must reset.}
\label{Fig:gen_ART}
\end{figure}

An elementary ART neural network model (Fig.~\ref{Fig:gen_ART}) usually consists of two fully connected layers as well as a system responsible for its decision-making capabilities:
\begin{itemize}
\item Feature representation field \Fin: this is the input layer.
In feedforward mode, the output $\bm{y}^{(F_1)}$ of this layer, or short-term memory (STM), simply propagates the input samples $\bm{x} \in \mathbb{R}^d$ to the \Fout\ layer via the bottom-up long-term memory units (LTMs) $\bm{\theta}^{bu}$. In feedback mode, the \Fin\ layer works as a comparator, in which~$\bm{x}$ and the \Fout's expectation (in the form of a top-down LTM $\bm{\theta}^{td}$) are compared and the outcome $\bm{y}^{(F_1)}$ is sent to the orienting subsystem. Hence, \Fin\ is also known as comparison layer.
\item Category representation field \Fout: this layer yields the network output $\bm{y}^{(F_2)}$ (STM). It is also known as recognition or competitive layer. Neurons, prototypes, categories and templates will be used interchangeably when referring to the \Fout\ nodes. The LTM associated with a category $j$ is $\bm{\theta}_j=\{\bm{\theta}^{bu}_j,\bm{\theta}^{td}_j\}$, $j=1,...,N$. Note that not all elementary ART models discussed in this survey have independent bottom-up and top-down LTM parts; however, $\bm{\theta}$ is always used to indicate the LTM (or set of adaptive parameters) of a given category.
\item Orienting subsystem: this is a system that regulates both the search and learning mechanisms by inhibiting or allowing categories to resonate.
\end{itemize}

\begin{table}[!b]
\centering
\caption{Unsupervised ART models notation.}
\begingroup\setlength{\fboxsep}{0pt}
\colorbox{lightgray}{
\begin{tabular*}{\columnwidth}{@{}ll@{\extracolsep{\fill}}}
\toprule
Notation & Description \\
\midrule
\midrule
$\bm{x}$            & input sample ($\bm{x} \in \bm{X}$) \\
$d$                 & original data dimensionality ($\bm{x} \in \mathbb{R}^d$)\\
\Fin                & feature representation field \\
\Fout               & category representation field \\
$N$                 & number of categories \\
$\bm{y}^{(F_1)}$    & \Fin \ activity/output (STM) \\
$\bm{y}^{(F_2)}$    & \Fout \ activity/output (STM) \\
$c$                 & a category \\
$\bm{\theta}$       & category parameters (LTM unit) \\
$T$                 & activation function \\
$M$                 & match function \\
$J$                 & chosen category index (via WTA) \\
$\rho$              & vigilance parameter \\
$VR$                & vigilance region \\
\bottomrule
\end{tabular*}
}\endgroup
\label{Tab:Notation}
\end{table}

Note that some ART models represent pre-processing procedures of the input samples by another layer preceding \Fin, namely the Input field F\textsubscript{0}. In this survey, it is assumed that the inputs to an ART network have already gone through the required transformations, and thus this layer is omitted from the discussion. 

ART models are competitive, self-organizing, dynamic and modular networks. When a sample $\bm{x}$ is presented, a winner-takes-all (WTA) competition takes place over its categories at the output layer \Fout. Then, the neuron $J$ that optimizes that model's \textit{activation function} across the nodes is chosen, e.g., the neuron that maximizes some similarity measure $T$ to the presented sample
\begin{equation}
J = \argmax\limits_j(T_j).
\label{Eq:intro_1}
\end{equation}

A category represents a hypothesis. Therefore, a hypothesis test cycle, commonly referred to as a vigilance test, is performed by the orienting subsystem to determine the adequacy of the selected category, i.e., the winner category must satisfy a match criterion (or several match criteria). If the confidence on such a hypothesis is larger than the minimum threshold (namely, the vigilance parameter $\rho$), the neural network enters in a resonance state and learning (i.e., adaptation of the long-term memory (LTM) units) is allowed. Otherwise, category $J$ is inhibited, the next highest ranked category is selected, and the search resumes. If no category satisfies the required resonance condition(s), then a new one is created to encode the presented input sample. This ability to reject a hypothesis/category via a two-way similarity measure, i.e. \textit{permissive clustering}~\cite{seiffertt2010}, makes ART stand out from other methods, such as k-means~\cite{MacQueen.1967a}. A vigilance region ($VR$) for a given network category $j$ can be defined in the data space as 
\begin{equation}
VR_j = \{\bm{x} : M_j(\bm{x}) \mbox{ satisfies the resonance constraint}\},
\label{Eq:intro_2}
\end{equation}
\noindent where $M_j$ is the \textit{match function}, which yields the confidence on hypothesis $j$. In other words, it is the region in the input space containing the set of all points such that the resonance criteria is met. Therefore satisfying (or not) the vigilance test for sample $\bm{x}$ can be modeled using
\begin{equation}
\mathds{1}_{VR_j}(\bm{x}) =
\begin{cases} 
1, & \mbox{if } \bm{x} \in VR_j \\ 
0, & \mbox{otherwise}
\end{cases},
\label{Eq:intro_3}
\end{equation}
\noindent where $\mathds{1}_{\{\cdot\}}$ is the indicator function.

The resonance constraint in Eq.~(\ref{Eq:intro_2}) is depends on the vigilance parameter $\rho$, which regulates the granularity of the network as ART maps samples to categories. Particularly, lower vigilance encourages generalization~\cite{vigdor2007}. Selecting the vigilance parameter is a difficult task in clustering problems. Concretely, the problem of choosing the number of clusters is traded for the problem of choosing the vigilance value.

Distinct ART models feature specific LTM units, activation and match functions, vigilance criteria and learning laws. Algorithm~\ref{Alg:ART} summarizes the dynamics of an elementary ART model.

\begin{algorithm}[!ht]
\DontPrintSemicolon
\SetKwInOut{Input}{Input}\SetKwInOut{Output}{Output}
\BlankLine
\Input{$\bm{x}$, $\{\bm{\alpha}, \bm{\beta}, \bm{\gamma}, \bm{\rho}, \bm{\lambda}\}$ (parameters).} 
\Output{$\bm{y}^{(F_2)}$.}
\algrule
\tcc{Notation\;
$\mathcal{C}$: set of ART nodes. \;
$\Lambda$: subset of highly active nodes ($\Lambda \subseteq \mathcal{C}$). \;
$\bm{\theta}$: LTM unit.\; 
$\bm{\alpha}$: activation function parameter(s).\; 
$\bm{\beta}$: learning function parameter(s).\; 
$\bm{\gamma}$: match function parameter(s).\; 
$\bm{\rho}$: vigilance parameter(s).\; 
$\bm{\lambda}$: initialization parameter(s).\; 
$f_T(\cdot)$: activation function.\;
$f_M(\cdot)$: match function.\;
$f_L(\cdot)$: learning function.\;
$f_V(\cdot)$: vigilance function (e.g., $f_V = \bigwedge\limits_{k} \mathds{1}_{VR_J}^k(\bm{x})$).\;
$f_N(\cdot)$: initialization function.
}
\nl \label{step1}Present input sample: $\bm{x} \in \bm{X}$. \;
\nl Compute activation function(s): $T_j = f_T(\bm{x}, \bm{\theta}_j, \bm{\alpha}),~\forall j \in \mathcal{C}$.\;
\nl \label{search}Perform WTA competition: $J = \argmax\limits_{j \in \Lambda}{(T_j)}$. \;
\nl Compute match function(s): $M_J^k = f_M^k(\bm{x}, \bm{\theta}_J, \bm{\gamma}),~\forall k,~k \geq 1$. \;
\nl Perform vigilance test(s): $V_J = f_V(\mathds{1}_{VR_J}^1(\bm{x}),...,\mathds{1}_{VR_J}^k(\bm{x}))$. \;
\nl \uIf{$V_J$ is TRUE}{
\nl Update category $J$: $\theta_J^{new} = f_L(\bm{x},\bm{\theta}_J^{old}, \bm{\beta})$.\;
}
\nl \Else{
\nl Deactivate category $J$: $\Lambda \leftarrow \Lambda - \{J\}$.\;
\nl \uIf{$\Lambda \neq \{ \emptyset \}$}{
\nl Go to step~\ref{search}.\;
}
\nl \Else{
\nl Set $J = |\mathcal{C} |+1$.\;
\nl Create new category: $\mathcal{C} \leftarrow \mathcal{C} \cup \{J\}$.\; 
\nl Initialize new category: $\theta_{J}^{new} = f_N(\bm{x}, \bm{\lambda})$.\;

}
}
\nl Set output:
$y_j^{(F_2)} = 
\begin{cases} 
1, & \mbox{if } j=J \\ 
0, & \mbox{otherwise}
\end{cases}$.\;
\nl Go to step~\ref{step1}.\;
\caption{Elementary ART algorithm.}
\label{Alg:ART}  
\end{algorithm}

\subsubsection{ART 1}
\label{Sec:ART1}

The ART 1 neural network~\cite{Carpenter1987} was the seminal implementation of the theory championed by Grossberg used for engineering applications. It relies on crisp set theoretic operators to cluster binary input samples using a similarity measure based on Hamming distance~\cite{Gotarredona.1998a}. 

\textbf{LTM.} 
ART 1 categories are parameterized with bottom-up and top-down adaptive weight vectors \mbox{$\bm{\theta} = \{\bm{w}^{bu},\bm{w}^{td}\}$}. 

\textbf{Activation.} 
When a sample $\bm{x}$ is presented to ART~1, the activation function of each category $j$ is computed as
\begin{equation}
T_j = \| \bm{x}  \cap \bm{w}_j^{bu} \|_1 \doteq \langle \bm{w}_j^{bu},\bm{x} \rangle = \sum\limits_{i=1}^d x_i w_{ji}^{bu}, 
\label{Eq:ART1_1}
\end{equation}
\noindent where $\bm{x}$ is a binary input, $\cap$ is a binary logic AND, $\bm{w}^{bu}$ is the bottom-up weight vector, $\| \cdot \|_1$ is the $L_1$ norm, and $\langle \cdot, \cdot \rangle$ is an inner product. 

When a given node $J$ is selected via the WTA competition, the output of the F\textsubscript{2} activity (short-term memory - STM) becomes
\begin{equation}
y_j^{(F_2)} = 
\begin{cases} 
1, & \mbox{if } j=J \\ 
0, & \mbox{otherwise}
\end{cases},
\label{Eq:ART1_2}
\end{equation}
\noindent moreover, the F\textsubscript{1} activity (short-term memory - STM) is defined as
\begin{equation}
\bm{y}^{(F_1)} = 
\begin{cases} 
\bm{x}, & \mbox{if $F_2$ is inactive} \\ 
\bm{x} \cap \bm{w}_J^{td}, & \mbox{otherwise}
\end{cases}. 
\label{Eq:ART1_1b}
\end{equation}

Note that the WTA competition always include one uncommitted node, which is is guaranteed to satisfy the vigilance criterion following Eq.~(\ref{Eq:ART1_3}).

\textbf{Match and resonance.} 
The highest activated node $J$ is tested for resonance using 
\begin{equation}
M_J =  \frac{\|\bm{y}^{(F_1)} \|_1}{\|\bm{x}\|_1} = \dfrac{ \| \bm{x}  \cap \bm{w}_J^{td} \|_1 }{\| \bm{x} \|_1},
\label{Eq:ART1_3}
\end{equation}
\noindent where $VR_J = \{\bm{x} : M_J(\bm{x}) \geq \rho\}$ and $\rho \in [0,1]$. The vigilance criterion checks if $\mathds{1}_{VR_J}(\bm{x})$ is true, and, in the affirmative case, the category is allowed to learn.

\textbf{Learning.} 
When the system enters a resonant state, learning is ensued as
\begin{equation}
\bm{w}_J^{td}(new) = \bm{x}  \cap \bm{w}_J^{td}(old),
\label{Eq:ART1_4}
\end{equation}
\begin{equation}
\bm{w}_J^{bu}(new) = \dfrac{L}{L - 1 + \| \bm{w}_J^{td}(new) \|_1} \bm{w}_J^{td}(new), 
\label{Eq:ART1_5}
\end{equation}
\noindent where $L>1$ is a user-defined parameter (larger values of $L$ bias the selection of uncommitted nodes over committed ones). Note that the bottom-up weight vectors are normalized versions of their top-down counterparts. If an uncommitted node is selected to learn sample $\bm{x}$, then another one is created and initialized as
\begin{equation}
\bm{w}^{td} = \Vec{\bm{1}},    
\end{equation}
\begin{equation}
\bm{w}^{bu} = \dfrac{L}{L-1+d}\bm{w}^{td}. 
\end{equation}  

ART 1 features the following appealing properties thoroughly discussed in~\cite{Gotarredona.1998a}: ``\textit{vigilance or variable coarseness, self-scaling, self-stabilization in a small number of iterations, online learning, capturing rate events, direct assess to familiar input patterns, direct assess to subset and superset patterns, biasing the network to form new categories.}''

\subsubsection{ART 2}
\label{Sec:ART2}

ART~2~\cite{Carpenter.1987a} and 2-A \cite{Carpenter1991b} represent the initial effort toward extending ART~1 (Sec.~\ref{Sec:ART1}) applications to real valued data. They were largely supplanted by Fuzzy ART (Sec.~\ref{Sec:FA}) which has since become one of the most widely used and referenced foundational building block for ART networks. This was followed by other architectures such as the ART~3~\cite{Carpenter1990} hierarchical architecture, Exact ART~\cite{Raijmakers.1997a} (which is a complete ART network based on ART~2) and Correlation-based ART~\cite{Yavas.2009a} along with its hierarchical variant~\cite{Yavas.2012a} which use correlation analysis methods for category matching. Particularly, the ART~2-A \cite{Carpenter1991b} architecture was developed following ART~2 with the same properties and a much faster speed.

\textbf{LTM.}
The internal category representation in ART 2-A consists of an adaptive scaled weight vector \mbox{$\theta = \{\bm{w}\}$}.

\textbf{Activation.}
The activation function of each category~$j$ in response to a normalized input sample $\bm{x}$ is computed as
\begin{equation}
T_j = 
\begin{cases}
\alpha \sum_i \bm{x}_i, &  \text{ if } j \text{ is uncommitted} \\
\bm{x} \bm{w}_j, & \text{ if } j \text{ is committed} \end{cases},
\end{equation}
where $\alpha \le \frac{1}{\sqrt{d}}$ is the choice parameter.

\textbf{Match and resonance.}
The category with the highest activation value is chosen via winner-takes-all selection. Its match function is computed as
\begin{equation}
M_J = T_J, 
\end{equation}
and the vigilance test is performed to determine whether resonance occurs using the following: $M_J \geq \rho$, where $0 \le \rho \le 1$ is the vigilance threshold.

If the winning category passes the vigilance test, resonance occurs, and the category is allowed to learn this input pattern. If the category fails the vigilance test, a reset signal is triggered for this category, and the category with the next highest activation is selected for the same process.

\textbf{Learning.} 
When resonance occurs, the weights of the winning category are updated as
\begin{equation}
\bm{w}_J(new) = 
\begin{cases}
\bm{x}, &  \text{ if } J \text{ is uncommitted} \\
\beta \bm{x} + (1-\beta)\bm{w}_J(old), &  \text{ if } j \text{ is committed} 
\end{cases},
\end{equation}
where $0 < \beta \le 1$ is the learning rate.

\subsubsection{Fuzzy ART}
\label{Sec:FA}

Fuzzy ART (FA)~\cite{Carpenter1991} is arguably the most widely used ART model. It extends the capabilities of ART 1 (Sec.~\ref{Sec:ART1}) to process real-valued data by incorporating fuzzy set theoretic operators~\cite{Zadeh1965}. Typically, samples are pre-processed by applying complement coding~\cite{Carpenter1991a, carpenter1992}. This transformation doubles the original input dimension while imposing a constant norm  ($\bm{x} \leftarrow [\bm{x}, \Vec{\bm{1}} - \bm{x}]$):
\begin{equation}
\|\bm{x}\|_1 = \sum\limits_{i=1}^{2d} x_i = \sum\limits_{i=1}^{d} x_i + \sum\limits_{i=1}^{d} ( 1 - x_i) = d.
\label{Eq:FA_0}
\end{equation}

This process encodes the degree of presence and absence of each data feature. The augmented input vector prevents a category proliferation type due to weight erosion~\cite{Carpenter.1997a}. 

\textbf{LTM.} 
Each category LTM unit is a weight vector $\bm{\theta} = \{ \bm{w} \}$. If complement coding is employed, then \mbox{$\bm{w}=[\bm{u},\bm{v}^c]$}, and the geometric interpretation of a category is a hyperrectangle (or hyperbox), in the data space, with lower left corner $\bm{u}$ and upper right corner $\bm{v}^c$ representing features ranges (minimum and maximum data statistics).

\textbf{Activation.} 
The activation function of a category $j$ is defined as (Weber law)
\begin{equation}
T_j = \dfrac{ \norm{ \bm{x}  \wedge \bm{w}_j }_1 }{\alpha + \| \bm{w}_j \|_1},
\label{Eq:FA_1}
\end{equation}
\noindent where $\wedge$ is a component-wise fuzzy AND/intersection (minimum), $\alpha>0$ is the choice parameter which is related to the system's complexity (it can be seen as a regularization parameter that penalizes large weights). Its role has been thoroughly investigated in~\cite{Georgiopoulos1996}. The activation function measures the degree to which $\bm{x}$ is a fuzzy subset of $\bm{w}_j$ and is biased towards smaller categories. The F\textsubscript{1} activity is defined as
\begin{equation}
\bm{y}^{(F_1)} = 
\begin{cases} 
\bm{x}, & \mbox{if $F_2$ is inactive} \\ 
\bm{x} \wedge \bm{w}_J, & \mbox{otherwise}
\end{cases}. 
\label{Eq:FA_2}
\end{equation}

\textbf{Match and resonance.} 
When the winner node $J$ is selected, the F\textsubscript{2} activity is
\begin{equation}
y_j^{(F_2)} = 
\begin{cases} 
1, & \mbox{if } j=J \\ 
0, & \mbox{otherwise}
\end{cases}. 
\label{Eq:FA_7}
\end{equation}
\noindent and a hypothesis testing cycle is conducted using
\begin{equation}
M_J = \frac{\|\bm{y}^{(F_1)} \|_1}{\|\bm{x}\|_1} = \frac{\|\bm{x} \wedge \bm{w}_J\|_1}{\|\bm{x}\|_1},
\label{Eq:FA_3}
\end{equation}
\noindent where $VR_J = \{\bm{x} : M_J(\bm{x}) \geq \rho\}$ and $\rho \in [0,1]$ is the vigilance parameter. The vigilance criterion checks if $\mathds{1}_{VR_J}(\bm{x})$ is true, and, in the affirmative case, the category is allowed to learn. An uncommitted category will always satisfy the match criterion. Fuzzy ART vigilance regions are hyperoctagons and thoroughly discussed in~\cite{Anagnostopoulos.2002a, Verzi2006, meng2015}. The match function ensures that if learning takes place, the updated category will not exceed the maximum allowed size. Specifically, category $j$'s size is measured as
\begin{equation}
 R_j  = \| \bm{v}_j - \bm{u}_j \|_1  = \sum\limits_{i=1}^{d} \left[ (1 - w_{j,d+i}) - w_{j,i} \right]  = d - \| \bm{w}_j \|_1,  
\label{Eq:FA_5}
\end{equation}
\noindent where, considering the complement coded inputs,  $-d \leq  R_j  \leq d$ (for an uncommitted category: $R_j  = -d$). Particularly, the match function measures the size of the category if it is allowed to learn the presented sample. Thus, the vigilance criterion imposes an upper bound to the category size defined by the vigilance parameter ($\rho$) 
\begin{equation}
 R_J \oplus \bm{x}  = d - \norm{ \bm{x}  \wedge \bm{w}_j }_1 \leq d(1-\rho),     
\label{Eq:FA_4}
\end{equation}
\noindent where $R_J \oplus \bm{x}$ represents the smallest hyperrectangle capable of enclosing both $R_J$ and the presented sample $\bm{x}$.

\textbf{Learning.} 
If the vigilance test fails, then the winner category is inhibited, and the search continues until another one is found or created. When the vigilance criterion in met by category $J$, it adapts using
\begin{equation}
\bm{w}_J(new) = (1-\beta)\bm{w}_J(old) + \beta(\bm{x} \wedge \bm{w}_J(old)),
\label{Eq:FA_6}
\end{equation}
\noindent where $\beta \in (0,1]$ is the learning parameter. If an uncommitted node is recruited to learn sample $\bm{x}$, then another one is created and initialized as $\bm{w} = \Vec{\bm{1}}$. According to Eq.~(\ref{Eq:FA_6}), the norm of a weight vector is monotonically non-increasing during learning since categories can only expand~\cite{vigdor2007}. 

\subsubsection{Fuzzy Min-Max}
\label{sec:fuzzy-min-max-art}

The Fuzzy Min-Max neural network~\cite{Simpson1993} is an unsupervised learning network that uses fuzzy set theory to build clusters using a hyperbox representation discovered via the fuzzy min-max learning algorithm. Each category in Fuzzy Min-Max is represented explicitly as a hyperbox, with the minimum and maximum points of the hyperbox as well as a value for the membership function that measures the degree to which each input pattern falls within this category. The category hyperboxes are adjusted to fit each input sample using a contraction and expansion algorithm that expands the hyperbox of the winning category to fit the input sample and then contracts any other hyperboxes that are found to overlap with the new hyperbox boundaries.

\subsubsection{Distributed ART}
\label{Sec:dART}

The distributed ART (dART)~\cite{Carpenter.1996a,Carpenter.1996b, Carpenter.1997a} features distributed code representation for activation, match and learning processes to improve noise robustness and memory compression in a system that features fast and stable learning. Particularly, in WTA mode, distributed ART reduces in functionality to fuzzy ART (Sec.~\ref{Sec:FA}).

\textbf{LTM.} 
The distributed ART LTM units consist of bottom-up ($\bm{\tau}^{bu}$) and top-down ($\bm{\tau}^{td}$) adaptive thresholds (\mbox{$\bm{\theta} = \{\bm{\tau}^{bu}, \bm{\tau}^{td}\}$}), which are initialized as small random values and $\Vec{\bm{0}}$, respectively. When employing complement coding, the geometric interpretation of a category $j$ is a family of hyperrectangles nested by the activation levels \mbox{$y_j^{(F_2)} \in [0,1]$}. The edges of hyperrectangle $R_j(y_j^{(F_2)})$ are defined, for each input dimension $i$, as the bounded interval $\left[[y_j^{(F_2)} - \tau_{j,i}^{bu}]^+, 1 - [y_j^{(F_2)} - \tau_{j,d+i}^{bu}]^+\right]$ --- where \mbox{$[\xi]^+ = \max(0, \xi)$} is a rectifier operator. Note that the $R_j$ size decreases as $y_j^{(F_2)}$ increases. Particularly, setting $y_j^{(F_2)}=1$ yields the smallest hyperrectangle $R(1)$, and the substitution \mbox{$\bm{w}_j = (1 - \bm{\tau}^{bu})$} corresponds to fuzzy ART's LTM.

\textbf{Activation.} 
The activation function can be defined as a choice-by-difference~\cite{Gjaja.1994a} ($T_j \in [0, d]$) variant
\begin{equation}
T_j  = \norm{ [\bm{x} \wedge (1 - \bm{\tau}_j^{bu}) - \bm{\Delta}_j ]^+}_1 
+ (1 - \alpha) \norm{ [\bm{\tau}_j^{bu} - \bm{\delta}_j ]^+}_1~,~0 < \alpha < 1,
\label{Eq:dART_2}
\end{equation}
\noindent or a Weber law~\cite{Carpenter1987} ($T_j \in [0, 1]$) variant
\begin{equation}
T_j = \dfrac{\norm{ [\bm{x} \wedge (1 - \bm{\tau}_j^{bu}) - \bm{\Delta}_j ]^+}_1}{\alpha + d - \norm{ [\bm{\tau}_j^{bu} - \bm{\delta}_j ]^+ }_1}~,~\alpha >0,
\label{Eq:dART_3}
\end{equation}
\noindent where $[\bm{\xi}]^+$ is a component-wise rectifier operator (i.e., $[\xi_k]^+ = \max(0, \xi_k)$ for each component $k$ of vector $\bm{\xi}$), and $\bm{\Delta}$ and $\bm{\delta}$ are the medium-term memory (MTM) depletion parameters. After the nodes' activations are computed, the \Fout\ activity can be obtained by employing the increased-gradient content-addressable-memory (IG CAM) rule:
\begin{equation}
y_j^{(F_2)} = 
\begin{cases}
\dfrac{(T_j)^p}{\sum\limits_{\lambda \in \Lambda} (T_\lambda)^p}, & \mbox{ if } j \in \Lambda  \\
0, & \mbox{ otherwise}
\end{cases},
\label{Eq:dART_1}
\end{equation}
\noindent such that $\| \bm{y}^{(F_2)} \|_1 = 1$ and $p>0$. The subset $\Lambda$ consists of the nodes such that $T_J \geq T_j$ for $J \in \Lambda$ and $j \notin \Lambda$. Examples are the Q-max rule (see Sec.~\ref{Sec:AMIC}) or greater than average activations (i.e., $\Lambda = \{j : T_j \geq T_{avg}\}$, $T_{avg} = 1/N \sum_{j=1}^N T_j$). Note that the power law $f(\zeta) = \zeta^p$ converges to WTA when $p \to +\infty$. 

\textbf{Match and Resonance.} 
The distributed ART's match function is defined as
\begin{equation}
M = \dfrac{\|\bm{y}^{{(F_1)}}\|_1}{\|\bm{x}\|_1},
\label{Eq:dART_6}
\end{equation}
\noindent where the $F_1$ activity is given by 
\begin{equation}
\bm{y}^{{(F_1)}} = \bm{x} \wedge \bm{\sigma},
\label{Eq:dART_4}
\end{equation}
\noindent and
\begin{equation}
\sigma_i = \sum\limits_{j=1}^N [y_j^{{(F_2)}} - \tau_{ji}^{td}]^+~,~\sigma_i \in [0,1].
\label{Eq:dART_5}
\end{equation}

Resonance occurs if $\mathds{1}_{VR}(\bm{x})=1$, where \mbox{$VR = \{\bm{x} : M(\bm{x}) \geq \rho\}$} and $\rho \in [0,1]$. Otherwise, the MTM depletion parameters are updated as
\begin{equation}
\Delta_{ji}(new) = \Delta_{ji}(old) \vee (x_i \wedge [y_j - \tau_{ji}^{bu}]^+),
\label{Eq:dART_7}
\end{equation}
\begin{equation}
\delta_{ji}(new) = \delta_{ji}(old) \vee (y_j \wedge \tau_{ji}^{bu}), \label{Eq:dART_8}
\end{equation}
\noindent and the distributed dynamics continue by recomputing Eqs.~(\ref{Eq:dART_1}) through~(\ref{Eq:dART_6}). Note that the depletion parameters $\bm{\Delta}$ and $\bm{\delta}$ are (re)set to $\Vec{\bm{0}}$ at the beginning of every input sample presentation.

\textbf{Learning.} 
When the system enters a resonant state, distributed learning takes place according to the nodes' activation levels. Specifically, the top-down adaptive thresholds are updated using the distributed outstar learning law~\cite{Carpenter.1994a}:
\begin{equation}
\tau_{ji}^{td}(new) = \tau_{ji}^{td}(old) + \beta\dfrac{\left[\sigma_i - x_i\right]^+}{\sigma_i} \left[y_j^{(F_2)} - \tau_{ji}^{td}(old) \right]^+,
\label{Eq:dART_9}
\end{equation}
\noindent whereas the bottom-up adaptive thresholds are updated using the distributed instar learning law~\cite{Carpenter.1997a}:
\begin{equation}
\tau_{ji}^{bu}(new) = \tau_{ji}^{bu}(old) + \beta \left[y_j^{(F_2)} - \tau_{ji}^{bu}(old) - x_i \right]^+,
\label{Eq:dART_10}
\end{equation}
\noindent where $\beta \in [0,1]$ is the learning rate. The adaptive thresholds' components, $\in [0,1]$, start near zero and monotonically increase during the learning process. After learning takes place, the depletion parameters $\bm{\Delta}$ and $\bm{\delta}$ are both reset to their initial values ($\Vec{\bm{0}}$). In WTA mode, the distributed instar and outstar learning laws become the instar~\cite{Grossberg.1972} and outstar~\cite{Grossberg.1968a, Grossberg.1969a} laws, respectively, and thus distributed ART reduces to fuzzy ART (Sec.~\ref{Sec:FA}).

\subsubsection{Gaussian ART} 
\label{Sec:GA}

Gaussian ART~\cite{williamson1996} was developed to reduce category proliferation in noisy environments and to provide a more efficient category LTM unit. 

\textbf{LTM.} 
Each category $j$ is a Gaussian distribution composed by mean $\bm{\mu}_j \in \mathbb{R}^d$, standard deviation $\bm{\sigma}_j \in \mathbb{R}^d$ and instance counting $n_j$ (i.e., the number of samples encoded by category $j$ used to compute its a priori probability). Therefore, a category is geometrically interpreted as a hyperellipse in the data space.

\textbf{Activation.} 
Gaussian ART is rooted in Bayes' decision theory, and as such its activation function is defined as:
\begin{equation}
T_j = \hat{p}(c_j | \bm{x}) = \frac{\hat{p}(\bm{x} | c_j) \hat{p}(c_j)}{\hat{p}(\bm{x})},   
\label{Eq:GA_1}
\end{equation}
\noindent where the likelihood is estimated as
\begin{equation}
\hat{p}(\bm{x} | c_j) =  \dfrac{exp\left[-\dfrac{1}{2} \left( \bm{\mu}_j - \bm{x} \right)^T \bm{\Sigma}_j^{-1} \left( \bm{\mu}_j - \bm{x} \right)\right]}{\sqrt{\left( 2 \pi \right)^{d} \det(\bm{\Sigma}_j})},
\label{Eq:GA_2}
\end{equation}
\noindent and the prior as
\begin{equation}
\hat{p}(c_j) = \dfrac{n_j}{\sum\limits_{i=1}^N n_i}.
\label{Eq:GA_3}
\end{equation}

Note that the evidence $\hat{p}(\bm{x})$ is neglected in the computations (since it is equal for all categories $c_j$), and feature independence is assumed, i.e., $\bm{\Sigma}_j$ is a diagonal matrix ($\bm{\Sigma}_j=diag(\sigma_{j,1}^2, ..., \sigma_{j,d}^2)$). Therefore, since it assumes uncorrelated features, it cannot capture covarying data. A category $J$ is then chosen following the maximum a posteriori (MAP) criterion:
\begin{equation}
J = \argmax\limits_j(T_j) = \argmax\limits_j \left[ \hat{p}(c_j | \bm{x}) \right].
\label{Eq:GA_4}
\end{equation}

\textbf{Match and Resonance.} 
The match function is defined as a normalized version of $\hat{p}(\bm{x} | c_j)$:
\begin{equation}
M_J = exp\left[-\dfrac{1}{2} \left( \bm{\mu}_J - \bm{x} \right)^T \bm{\Sigma}_J^{-1} \left( \bm{\mu}_J - \bm{x} \right) \right],
\label{Eq:GA_5}
\end{equation}
\noindent which is then compared to the vigilance parameter threshold $\rho \in (0,1]$. Note that in the original Gaussian ART paper~\cite{williamson1996}, a log discriminant is used to reduce the computational burden in both the activation (Eq.~(\ref{Eq:GA_1})) and match (Eq.~(\ref{Eq:GA_5})) functions. 

\textbf{Learning.} 
When the vigilance criterion is met, learning is ensued for the resonating category $J$ as
\begin{equation}
n_J(new) = n_J(old) + 1, 
\label{Eq:GA_6}
\end{equation}
\begin{equation}
\hat{\bm{\mu}}_J(new) =  \left(1-\frac{1}{n_J(new)}\right)\hat{\bm{\mu}}_J(old) +\frac{1}{n_J(new)} \bm{x},
\label{Eq:GA_7}
\end{equation}
\begin{equation}
\sigma_{J,i}^2(new)   = \left(1-\frac{1}{n_J(new)}\right)\sigma_{J,i}^2(old)  + \frac{1}{n_J(new)} \left( \mu_{J,i}(new) - x_i \right)^2.  
\label{Eq:GA_8}
\end{equation}

If a new category is created, then it is initialized with $n_{N+1}=1$, $\bm{\mu}_{N+1} = \bm{x}$, and $\bm{\Sigma}_{N+1} = \sigma_{init}^2 \bm{I}$ (isotropic). The initial standard deviation $\sigma_{init}$ in Gaussian ART directly affects the number of categories created.

\subsubsection{Hypersphere ART}
\label{Sec:HA}

The Hypersphere ART (HA)~\cite{anagnostopoulos2000} architecture was designed as a successor for Fuzzy ART (Section~\ref{Sec:FA}) that inherits its advantageous qualities while utilizing fewer categories and having a more efficient internal knowledge representation.

\textbf{LTM.}
Each category is represented as $\bm{\theta} = \{R, \bm{m}\}$, where $\bm{m}_j \in \mathbb{R}^{d}$ and $R_j \in \mathbb{R}$ are the centroid and radius, respectively. Since it does not require complement coding of input samples, it uses $d+1$ memory per category, which is a smaller memory requirement than fuzzy ART, which uses $2d$ memory to represent the hyperrectangular categories. Naturally, categories are hyperspheres in the data space.

\textbf{Activation.}
The category activation function $T_j$ for each $F_2$ category $j$ is calculated as:
\begin{equation}\label{eq:ha-choice}
T_j = \dfrac{\bar{R} - \max(R_j, ||\bm{x} - \bm{m}_j||_2)}{\bar{R} - R_j + \alpha},
\end{equation} 
where $||\cdot||_2$ is the $L_2$ (Euclidean) norm, $\alpha \in (0, \infty)$ is the choice parameter and $\bar{R} \in [R_{\text{max}}, \infty)$ is the radial extend parameter which controls the maximum possible category size achieved during training. The lower-bound $R_{\text{max}}$ is defined as:

\begin{equation}\label{eq:ha-rmax}
R_{\text{max}} = \dfrac{1}{2} \max_{i,j}{||\bm{x_i} - \bm{x_j}||_2}
\end{equation}

\textbf{Match and resonance.}
The winning category $J$ is selected using WTA competition, and the match function is computed as
\begin{equation}\label{eq:ha-vigilance}
 M_J = 1 - \dfrac{\max(R_j, || \bm{x} - \bm{m_j} ||_2)}{\bar{R}},
\end{equation}
\noindent where the vigilance criterion is $M_J \geq \rho$.

\textbf{Learning.}
If the winning category satisfies the vigilance test, then resonance occurs, and the radius $R_J$ and centroid $\bm{m}_J$ of the winning node are updated as follows:
\begin{equation}\label{eq:ha-update-radius}
R_J^{\text{new}}  = R_J^{\text{old}}  
+ \dfrac{\beta}{2} \left[\max\left(R_J^{\text{old}}, || \bm{x} - \bm{m}_J^{\text{old}} ||_2\right) - R_J^{\text{old}}\right],
\end{equation}
\begin{equation}\label{eq:ha-update-centroid}
\bm{m}_J^{\text{new}}  = \bm{m}_J^{\text{old}} + \dfrac{\beta}{2} \left(\bm{x} - \bm{m}_J^{\text{old}}\right) \\ 
\left[1 - \dfrac{\min\left(R_J^{\text{old}}, || \bm{x} - \bm{m}_J^{\text{old}} ||_2\right)}{|| \bm{x} - \bm{m}_J^{\text{old}} ||_2}\right], 
\end{equation}
\noindent where $\beta \in (0, 1]$ is the learning rate parameter.

If the winning category fails the vigilance test, it is reset, and the process is repeated. Eventually, either a category succeeds or a new one is created with its radius and centroid initialized as $R_J = 0$ and $\bm{m}_J = \bm{x}$, respectively.

\subsubsection{Ellipsoid ART}
\label{Sec:EA}

Ellipsoid ART (EA)~\cite{anagnostopoulos2001,anagnostopoulos2001b} is a generalization of hypersphere ART that uses hyperellipses instead of hyperspheres to represent the categories. These require $2d + 1$ memory and are subjected to two distinct constraints during training: (1)~maintain a constant ratio between the lengths of their major and minor axes, and~(2) maintain a fixed direction of their major axis once it is set. These restrictions, however, can pose some limitations to the categories discovered by ellipsoid ART depending on the order in which the input samples are presented.

\textbf{LTM.}
A category $j$ in ellipsoid ART is described by its parameters $\bm{\theta}_j = \{\bm{m}_j, \bm{d}_j, R_j\}$, where $\bm{m}_j$ is the centroid of the category's hyperellipses, $\bm{d}_j$ is the direction of the category's major axis and $R_j$ is the category's radius (or half the length of its major axis).

\textbf{Activation.}
The distance between an input sample and a category $j$ is calculated as:
\begin{equation}
dis(\bm{x} , \bm{m}_j ) = 
\begin{cases}
\dfrac{1}{\mu} \sqrt{|| \mathbf{x} - \mathbf{m}_j ||_2^2 - \left(1 - \mu^2 \right) \left[ \mathbf{d}_j^T \left(\mathbf{x} - \mathbf{m}_j \right) \right]^2 } & \mbox{ if } \mathbf{d}_j \neq \mathbf{0}  \\
\qquad || \mathbf{x} - \mathbf{m}_j ||_2 & \mbox{ if } \mathbf{d}_j = \mathbf{0}
\end{cases},
\end{equation}
where $||\cdot||_2$ is the $L_2$ (Euclidean) vector norm and \mbox{$\mu \in (0, 1]$} is a user-specified parameter that defines the ratio between a category's major and minor axes. The category activation function $T_j$ for each category $j$ is then calculated as:
\begin{equation}\label{eq:ea-choice}
T_j = \dfrac{\bar{R} - R_j - \max \left\lbrace R_j, dis(\bm{x} , \bm{m}_j ) \right\rbrace }{\bar{R}- 2 R_j + \alpha},
\end{equation}
where $\alpha \in (0, +\infty)$ is the choice parameter, and \mbox{$\bar{R} \ge \frac{1}{\mu} \max\limits_{p,q} \norm{ \bm{x}_p - \bm{x}_q }_2 $} is a user-specified parameter.

\textbf{Match and resonance.}
The match function of the winning category $J$ selected using winner-takes-all is given by
\begin{equation}\label{eq:ea-activation}
M_J = 1 - \dfrac{R_J + \max \left\lbrace R_J, dis(\bm{x} , \bm{m}_J) || \right\rbrace }{\bar{R}},
\end{equation}
where $\rho \in (0, 1]$ is the vigilance parameter. 

\textbf{Learning.}
If the winning category $J$ satisfies \mbox{$M_J \geq \rho$}, then resonance occurs, and it is updated as follows:
\begin{equation}\label{eq:ea-update-radius}
R_J^{\text{new}}  = R_J^{\text{old}}  
+ \dfrac{\beta}{2} \left[ \max \left\lbrace R_J^{\text{old}}, dis(\bm{x} , \bm{m}_J^{\text{old}}) \right\rbrace - R_J^{\text{old}} \right],
\end{equation}
\begin{equation}\label{eq:ea-update-centroid}
\bm{m}_J^{\text{new}}  = \bm{m}_J^{\text{old}} + \dfrac{\beta}{2} \left(\bm{x} - \bm{m}_J^{\text{old}}\right)  
\left[1 - \dfrac{\min \left\lbrace R_J^{\text{old}}, dis(\bm{x} , \bm{m}_J^{\text{old}}) \right\rbrace }{dis(\bm{x} , \bm{m}_J^{\text{old}}) }\right], 
\end{equation}
\begin{equation}\label{eq:ea-update-direction}
\bm{d}_j = \dfrac{\bm{x}_{(2)} - \bm{m}_J}{|| \bm{x}_{(2)} - \bm{m}_J ||_2},
\end{equation}
where $\beta \in (0, 1]$ is the learning rate, and $\bm{x}_{(2)}$ represents the second input sample to be encoded by this category. When a new category is created, its major axis direction $\bm{d}_J$ is initially set to the zero vector $\Vec{\bm{0}}$, and then Eq.~(\ref{eq:ea-update-direction}) is used to update it when the second pattern is committed to the category. The hyperellipse's major axis direction stays fixed after that.

If the winning category fails the vigilance check, then it is inhibited, and the entire process is repeated until a winner category satisfies the resonance criterion. If no existing category succeeds, then a new category is created with its weights initialized with $R_J = 0$, $\bm{m}_J = \bm{x}$, and $\bm{d}_J = \Vec{\bm{0}}$. 

\subsubsection{Quadratic neuron ART}
\label{Sec:QuadART}

The quadratic neuron ART model~\cite{ChunSu2002,ChunSu2005} was developed in the context of a multi-prototype-based clustering framework that integrates dynamic prototype generation and hierarchical agglomerative clustering to retrieve arbitrarily-shaped data structures.

\textbf{LTM.}
A category $j$ is a quadratic neuron~\cite{DeClaris.1991a, DeClaris.1992a, Su.1997a, ChunSu2001} parameterized by \mbox{$\bm{\theta}_j = \{s_j, \bm{W}_j, \bm{b}_j \}$}, where $s_j$, $\bm{W}_j = [w^{(j)}_{k,i}]_{d \times d}$, and $\bm{b}_j$ are the adaptable LTMs. Particularly, these neurons are hyperellipsoid structures in the multidimensional data space.

\textbf{Activation.} The activation of a quadratic neuron~$j$ is given by
\begin{equation}
T_j = exp\left(-s_j^2 \norm{\bm{z}_j - \bm{b}_j}^2_2 \right), 
\label{Eq:QuadART_1}
\end{equation}
\noindent where $\bm{z}_j$ is a linear transformation of the input $\bm{x}$
\begin{equation}
\bm{z}_j = \bm{W}_j \bm{x}. 
\label{Eq:QuadART_2}
\end{equation}

\textbf{Match and resonance.} After the winning node $J$ is selected using WTA competition, the system will enter a resonant state if node $J$'s response is larger than or equal to the vigilance parameter $\rho$, i.e., if \mbox{$M_J \geq \rho$}, where the match function is equal to the activation function (Eq.~(\ref{Eq:QuadART_1})).

\textbf{Learning.} If the vigilance criterion is satisfied for node~$J$, then its parameters $\bm{p} \in \{s_j, \bm{W}_j, \bm{b}_j \}$ are adapted using gradient ascent
\begin{equation}
\bm{p}(new) = \bm{p}(old) + \eta \dfrac{\partial T_J}{\partial \bm{p}(old)}, 
\label{Eq:QuadART_3a}
\end{equation}
\noindent where $\eta$~is the learning rate. Specifically,
\begin{equation}
b_{J,i}(new)  = b_{J,i}(old) + \eta_b \left[ 2s_J^2 T_J \left(z_{J,i} - b_{J,i}\right)\right],
\label{Eq:QuadART_3}
\end{equation}
\begin{equation}
w^{(J)}_{k,i}(new) =  w^{(J)}_{k,i}(old) + \eta_w \left[ -2s_J^2 T_J \left( z_{J,k} - b_{J,k}\right)x_i\right],
\label{Eq:QuadART_4}
\end{equation}
\begin{equation}
s_J(new) = s_J(old) + \eta_s \left( -2s_J T_J \norm{\bm{z}_J - \bm{b}_J}^2_2 \right),
\label{Eq:QuadART_5}
\end{equation}
\noindent where $\eta_b$, $\eta_w$ and $\eta_s$ are the learning rates. Otherwise, a new category is created and initialized with $\bm{b}_{N+1}=\bm{x}$, $\bm{W}_{N+1} = \bm{I}_{d \times d}$, and $s_{N+1} = s_{init}$, where $s_{init} \in \mathbb{R}$ is a user-defined parameter.

\subsubsection{Bayesian ART} 
\label{Sec:BA}

\textbf{LTM.} 
Bayesian ART (BA)~\cite{vigdor2007} is another architecture using multidimensional Gaussian distributions to parameterize the categories: \mbox{$\bm{\theta} = \{\mathcal{N}(\bm{\mu}, \bm{\Sigma}), p\}$}, where $\bm{\mu}$, $\bm{\Sigma}$ and $p$ are the mean, covariance matrix, and prior probability, respectively. The latter parameter is computed using the number of samples $n$ learned by a category. 

\textbf{Activation.} 
Like Gaussian ART (Sec.~\ref{Sec:GA}), Bayesian ART also integrates Bayes decision theory in its framework. Thus, its activation function is given by the posterior probability of category~$j$:
\begin{equation}
T_j = \hat{p}(c_j | \bm{x}) = \dfrac{ \hat{p}(\bm{x}|c_j) \hat{p}(c_j) }{\sum\limits_{l=1}^{N} \hat{p}(\bm{x}|c_l) \hat{p}(c_l)}, 
\label{Eq:BA_1}
\end{equation}
\noindent where $\hat{p}(\bm{x}|c_j)$ is the same as Eq.~(\ref{Eq:GA_2}) but uses a full covariance matrix (instead of diagonal), and $\hat{p}(c_j)$ is the estimated prior probability of category $j$ as in Eq.~(\ref{Eq:GA_3}).

\textbf{Match and Resonance.} 
After the WTA competition is performed and the winner category $J$ is selected using the maximum a posteriori probability (MAP) criterion (Eq.~(\ref{Eq:GA_4})), the match function is computed as 
\begin{equation}
M_J = \det(\bm{\Sigma}_J),  
\label{Eq:BA_2}
\end{equation}
\noindent such that the vigilance criterion is designed to limit category $J$'s hyper-volume. The vigilance test is defined as $M_J \leq \rho$, where $\rho$ represents the maximum allowed hyper-volume.

\textbf{Learning.} 
If the selected category resonates (i.e., the match criterion is satisfied), then learning occurs. The sample count and means are updated using Eq.~(\ref{Eq:GA_6}) and Eq.~(\ref{Eq:GA_7}), respectively. The covariance matrix is updated as:
\begin{equation}
\hat{\bm{\Sigma}}_J(new)  = \left(\frac{n_J(old)}{n_J(new)}\right)\hat{\bm{\Sigma}}_J(old)
+ \frac{1}{n_J(new)}(\bm{x} - \hat{\bm{\mu}}_J(new))(\bm{x} - \hat{\bm{\mu}}_J(new))^T \odot \bm{I},
\label{Eq:BA_3}
\end{equation}
\noindent which corresponds to the sequential maximum-likelihood estimation of parameters for a multidimensional Gaussian distribution~\cite{vigdor2007}. The Hadamard product $\odot$ is used when a diagonal covariance matrix is desired. Otherwise, a new category is created with $n_{N+1}=1$, $\bm{\mu}_{N+1} = \bm{x}$, and $\bm{\Sigma}_{N+1} = \bm{\Sigma}_{init}$. Naturally, the initial covariance matrix should satisfy the vigilance constraint (i.e., \mbox{$\bm{\Sigma}_{init} = \sigma_{init}^2\bm{I}$}, where $\sigma_{init}^2 \ll \rho^{1/d}$). In this ART model, categories can both grow and shrink.

\subsubsection{Grammatical ART}
\label{Sec:GramART}

The Grammatical ART (GramART) architecture~\cite{Meuth2009} represents a specialized version of ART designed to work with variable-length input patterns which are used to encode grammatical structure. It builds templates while adhering to a Backus-Naur form grammatical structure~\cite{Knuth.1964a}. 

\textbf{LTM.}
To allow for comparisons between variable-length input patterns, GramART uses a generalized tree representation to encode its internal categories. Each node in the tree for a category contains an array representing the distribution of the different possible grammatical symbols at that node.

\textbf{Activation.}
The activation function for a category $j$ is defined as a parallel to Fuzzy ART's activation function (Sec.~\ref{Sec:FA}), but GramART defines its own operator for calculating the intersection between a category and an input pattern. A tree in GramART is defined as an ordered pair $(N, R)$ where $N$ is a set of nodes and $R$ is a set of binary relations that describe the structure of the tree. For nodes $x$ and $y$:
\begin{equation}
R(x, y) = 
\begin{cases}
0, & \text{ if } y \text{ is not a successor of } x \\
> 0, & \text{ if } y \text{ is a successor of } x
\end{cases},
\end{equation}
The activation of a category $j$ in GramART is given  by
\begin{equation}
T_j = \dfrac{ | \bm{x}  \cap \bm{w}_j | }{ \| \bm{w}_j \| },
\end{equation}
where the intersection operator $| \bm{x}  \cap \bm{w}_j |$ is defined as:
\begin{equation}
| \bm{x}  \cap \bm{w}_j | = \sum_{i = 0}^{r} w_j [i, x_i],
\end{equation}
and $w_j[i, x_i]$ represents each of the values stored in $\bm{w}_j$ corresponding to the symbols present in the input pattern~$\bm{x}$.
The tree norm operator $\| \bm{w}_j \|$ is defined as the number of nodes in the tree.

\textbf{Match and resonance.}
The category with the highest activation value is chosen using winner-takes-all selection, and the following vigilance criterion is checked to determine whether the input pattern resonates with this category:
\begin{equation}
M_J = \dfrac{ | \bm{x}  \cap \bm{w}_J | }{ \| \bm{x} \| } > \rho.
\end{equation}
If this vigilance criterion is satisfied, resonance occurs and the category is allowed to learn this input pattern. Otherwise, it is reset, and the category with the next best activation is checked.

\textbf{Learning.}
When resonance occurs, the weight of the winning category is updated using the following learning rule:
\begin{equation}
w_j[i] = \dfrac{w_j[i] * N + \delta_{j}}{N + 1},
\end{equation}
where
\begin{equation}
\delta_{j} =
\begin{cases}
1, & \text{ if } x_i = j\\
0, & \text{ otherwise }
\end{cases}.
\end{equation}
The weights are updated recursively down the grammar tree, and they reflect the probability of a tree symbol occurring in the node representing this particular category.

\subsubsection{Validity index-based vigilance fuzzy ART}
\label{Sec:CVIFA}

The validity index-based vigilance fuzzy ART~\cite{leonardo2017} endows fuzzy ART with a second vigilance criterion based on cluster validity indices~\cite{xu2009}. The usage of this immediate reinforcement signal alleviates input order dependency and allows for a more a robust hyper-parameterization.

\textbf{LTM.} 
This is a fuzzy ART-based architecture. Therefore, categories are hyperrectangles as described in Sec.~\ref{Sec:FA}. 

\textbf{Activation.} 
The validity index-based vigilance fuzzy ART activation function is equal to fuzzy ART's and thus, is computed using Eq.~(\ref{Eq:FA_1}) in Sec.~\ref{Sec:FA}. 

\textbf{Match and Resonance.} 
After a winner $J$ is selected, the first match function ($M_J^1$) is identical to fuzzy ART's (Eq.~(\ref{Eq:FA_3}) in Sec.~\ref{Sec:FA}), whereas the second ($M_J^2$) is defined as
\begin{equation}
M_J^2 = \Delta f = f(\hat{\Omega}) - f(\Omega),
\label{Eq:CVIFA_1}
\end{equation}
\noindent which represents the penalty (or reward) incurred by assigning sample $\bm{x}$ to category $J$ and thereby changing the current clustering state of the data set from $\Omega$ to $\hat{\Omega}$ (if there is no change in assignment, then $M_J^2=0$). The function $f(\Omega)$ corresponds to a cluster validity index value given a partition $\Omega=\{\omega_1,...,\omega_k\}$ of disjointed clusters $\omega_i$ (defined by categories $i$), where $\bigcup\limits_{i=1}^{k} \omega_i = \bm{X}$. The second vigilance region is then $VR_J^2 = \{\bm{x} : M_J^2(\bm{x}) \geq \rho_2\}$, and $\rho_2 \in \mathbb{R}$. The vigilance criterion checks if $\mathds{1}_{VR_J}(\bm{x})=1$. In the affirmative case, the category is allowed to learn. Note that the discussion so far implies the maximization of a cluster validity index; naturally, when minimization is sought, the inequality in the definition of $VR_J^2$ should be reversed. This is a greedy algorithm that selects the best clustering assignment based on immediate feedback. Naturally, performance is biased toward the data structures favored by the selected cluster validity index.

\textbf{Learning.} 
If both vigilances are satisfied, then learning is ensued. Otherwise, the search resumes or a new category is created. The learning rules are identical to fuzzy ART's (Sec.~\ref{Sec:FA}). Note that the validity index-based vigilance fuzzy ART model learns in offline mode, given that the entire data is used for the computation of Eq.~(\ref{Eq:CVIFA_1}). 

\subsubsection{Dual vigilance fuzzy ART}
\label{Sec:DVFA}

The dual vigilance fuzzy ART (DVFA)~\cite{leonardo.2018b} seeks retrieve arbitrarily shaped clusters with low parameterization requirements via a single fuzzy ART module. This is accomplished by augmenting fuzzy ART with two vigilance parameters, namely, the upper bound ($\rho_{\text{UB}} \in [0, 1]$) and lower bound (\mbox{$0 \leq \rho_{\text{LB}} \leq \rho_{\text{UB}} \leq 1$}), representing quantization and cluster similarity, respectively. 

\textbf{LTM.} 
The categories of the dual vigilance fuzzy ART are hyperrectangles.

\textbf{Activation.} 
The activation function of the dual vigilance fuzzy ART is the same as fuzzy ART's (Eq.~(\ref{Eq:FA_1}) in Sec.~\ref{Sec:FA}).

\textbf{Match and resonance.} 
When a category  $J$ is chosen by the WTA competition, it is subjected to a dual vigilance mechanism. The first match function ($M_J^1$) uses $\rho_{\text{UB}}$ in Eq.~(\ref{Eq:FA_3}), whereas the second ($M_J^2$) is conducted using a more relaxed constraint; i.e., it uses $\rho_{\text{LB}}$ in Eq.~(\ref{Eq:FA_3}).

\textbf{Learning.} 
If the first vigilance criterion is satisfied, then learning proceeds as in fuzzy ART (Eq.~(\ref{Eq:FA_6})). Otherwise, the second test is performed, and, if satisfied, a new category is created and mapped to the same cluster as the category undergoing the dual vigilance tests via a mapping matrix \mbox{$\bm{M}_{map} = \left[m_{row,col}\right]_{N \times K}$} (where $N$ is the number of categories and $K$ is the number of clusters). Alternately, if both tests fail, then the search continues with the next highest ranked category; if there are none left, then a new node is created and the matrix $\bm{M}_{map}$ expands:
\begin{equation}
m_{r,c} = 
\begin{cases} 
1, & \mbox{if $row=N+1$ and $col=K+1$}\\ 
0, & \mbox{if $row=N+1$ and $col \neq K+1$}\\
0, & \mbox{if $row \neq N+1$ and $col=K+1$} \\
m_{r,c}, & \mbox{if $row \neq N+1$ and $col \neq K+1$}
\end{cases}.
\label{Eq:DVFA_1}
\end{equation}

The associations between categories and clusters are permanent in this incremental many-to-one mapping (multi-prototype representation of clusters), and they enable the data structures of arbitrary geometries to be detected by dual vigilance fuzzy ART's simple design.

\subsection{Topological architectures}

The ART models discussed in this section are designed to enable multi-category representation of clusters, thus capturing the data topology more faithfully. Generally, they are used to cluster data in which arbitrarily-shaped structures are expected (multi-prototype clustering methods). 

\subsubsection{Fuzzy ART-GL}
\label{Sec:FAGL}

Fuzzy ART with group learning (fuzzy ART-GL) model~\cite{Isawa2007} augments fuzzy ART (Sec.~\ref{Sec:FA}) with topology learning (inspired by neural-gas~\cite{Martinetz.1991a, Martinetz.1994a}) to retrieve clusters with arbitrary shapes. The code representation, LTMs and dynamics of fuzzy ART remain the same. However, when a sample is presented, a connection between the first and second resonating categories (if they both exist) is created by setting the corresponding entry of an adjacency matrix to one. This model also possesses an age matrix, which tracks the duration of such connections and whose dynamics are as follows: the entry related to the first and second current resonating categories is refreshed (i.e., set to zero) following a sample presentation, whereas all other entries related to the first resonating category are incremented by one. Connections with an age value above a certain threshold expire, i.e., they are pruned (note that the threshold varies deterministically over time). This procedure allows this model to dynamically create and remove connections between categories during learning (co-occurrence of resonating categories, thus following a Hebbian approach). Clusters are defined by groups of connected categories. 

The fuzzy ART combining overlapped category in consideration of connections (C-fuzzy ART) variant~\cite{isawa2008} was developed to mitigate category proliferation, which is accomplished by merging the first resonant category with another connecting and overlapping category. Another variant introduced in~\cite{Isawa2008b, Isawa2009} augments the latter model with individual and adaptive vigilance parameters to further reduce category proliferation.

\subsubsection{TopoART}
\label{Sec:TopoART}

Fuzzy topoART~\cite{Tscherepanow2010} is a model that combines fuzzy ART (Sec.~\ref{Sec:FA}) and topology learning (inspired by self-organizing incremental neural networks~\cite{Furao2006}). Specifically, it features the same representation, activation/match functions, vigilance test and search/learning mechanisms as fuzzy ART, while integrating noise robustness and topology-based learning. 

Briefly, the topoART model consists of two fuzzy ART-based modules (topoARTs A and B) that cluster, in parallel, the data in two hierarchical levels, while sharing the same complement coded inputs. Each category is endowed with an instance counting feature $n$ (i.e., sample count), such that every $\tau$ learning cycles (i.e., iterations) categories that encoded less than a minimum number of samples $\phi$ are dynamically removed. Once this threshold is reached, ``candidate'' categories become ``permanent'' categories, which can no longer be deleted. In this setup, module~A serves as a noise filtering mechanism for module~B. The propagation of a sample to module B depends on which type of module~A's category was activated. Specifically, a sample is fed to module~B if and only if the corresponding module~A's resonant category is ``permanent''; therefore, module~B will only focus on certain regions of the data space. Note that no additional information is passed from module A to B, and both can form clusters independently.

Regarding the hierarchical structure, the vigilance parameters of modules A and B are related by
\begin{equation}
\rho_b = \frac{1}{2}\left( \rho_a + 1 \right),
\end{equation}
\noindent such that module B's maximum category size is $50$\% smaller than module A's ($\rho_a$ and $\rho_b$ are module~A's and B's vigilance parameters, respectively), which implies that module B has a higher granularity ($\rho_b \geq \rho_a$) and thus yields a finer partition of the data set.

TopoART employs competitive and cooperative learning: not only the winner category $J_1$ but also the second winner $J_2$ is allowed to learn (naturally, both need to satisfy the vigilance criteria). The learning rates are set as $\beta_{J_2}<\beta_{J_1} =1$, such that the second winner partially learns to encode the presented sample. If the first and second winner both exist, then they are linked to establish a topological structure. These lateral connections are permanent, unless categories are removed via the noise thresholding procedure. Clusters are formed by the connected categories, thus better reflecting the data distribution and enabling the discovery of arbitrarily-shaped data structures (topoART is a graph-based multi-prototype clustering method). 

Finally, in prediction mode, the following activation function, which is independent of category size, is used: 
\begin{equation}
T_j = 1 - \frac{\norm{\left( \bm{x} \wedge \bm{w}_j \right) - \bm{w}_j}_1}{\norm{\bm{x}}_1},
\label{Eq:TopoART_1}
\end{equation}
\noindent the vigilance test is neglected, and only ``permanent'' nodes are allowed to be activated. 

A number of topoART variants have been developed in the literature, e.g., the hypersphere topoART \cite{Tscherepanow2012a}, which replaces fuzzy ART modules with hypersphere ARTs (Sec.~\ref{Sec:HA}); the episodic topoART \cite{Tscherepanow2012b} which incorporates temporal information (i.e., time variable and thus the order of input presentation) to build a spatio-temporal mapping throughout the learning process and generate ``episode-like'' clusters; and the \mbox{topoART-AM}~\cite{Tscherepanow2011b}, which builds hierarchical hetero-associative memories via a recall mechanism. 

\subsection{Hierarchical architectures}
\label{Sec:hierarchical}

Elementary ART modules have been used as building blocks to construct both bottom-up (agglomerative) and top-down (divisive) hierarchical architectures. Typically, these follow one of two designs~\cite{Massey2009}: (i) cascade (series connection) of ART modules in which the output of a preceding ART layer is used as the input of the succeeding one, or (ii)~parallel ART modules enforcing different vigilance criteria while having a common input layer. 

\subsubsection{ARTtree}
\label{Sec:ARTtree}

The ARTtree~\cite{Wunsch1993} is a way of building a hierarchy of ART neural modules in which an input sample is sent simultaneously to every module in every level of the tree. Each node in the ART tree hierarchy is connected to one of its parent's \Fout \ categories, and each of the \Fout \ categories in this node is connected to one of its children. The nodes in each layer of the tree hierarchy share a common vigilance value, and the vigilance typically increases further down the tree such that tiers of the tree that have more nodes are associated with higher vigilance values.

When an input sample is presented to the ARTtree hierarchy, all the ART nodes can be allowed to perform their match and activation functions, but only the node connected to its parent's winning \Fout \ category is allowed to resonate with and learn this pattern. Therefore, resonance only cascades down a single path in the ARTtree, and no other nodes outside that path are allowed to learn this sample. This can effectively allow ART to perform a type of varying-$k$-means clustering~\cite{Wunsch1993}.

The highly parallel nature of ARTtree lends itself well to hardware-based implementations, such as optoelectronic implementations~\cite{Wunsch1993} and massively parallel implementations via general purpose Graphics Processing Unit (GPU) acceleration~\cite{sejun2011}. The study presented in~\cite{sejun2011} performed this task using NVIDIA CUDA GPU hardware and an implementation of ARTtree that uses fuzzy ART units in the tree nodes. The results reported in the study show a massive speed boost for deep trees when compared to the CPU in terms of computing time, while smaller trees performed worse on the GPU due to the high data transfer penalties between the CPU and GPU memory.

\subsubsection{Self-consistent modular ART}
\label{Sec:SMART}

The self-consistent modular ART (SMART)~\cite{Bartfai1994} is a modular architecture designed to perform hierarchical divisive clustering (i.e., to represent different levels of data granularity in a top-down approach). It builds a self-consistent hierarchical structure via self-organization and uses ART~1 (Sec.~\ref{Sec:ART1}) as elementary units. In this architecture, a number of ART modules operate in parallel with different vigilance parameter values, while receiving the same input samples and connecting in a manner that makes the hierarchical cluster representation self-consistent. These connections are such that many-to-one mapping of specific to general categories is learned across such modules. Specifically, the hierarchy is explicitly represented via associative links between modules. 

Concretely, a two-level SMART architecture can be implemented using an ARTMAP (Sec.~\ref{Sec:AM}) in auto-associative mode; i.e., ARTMAP is used in an unsupervised manner by presenting the same input sample to both modules A and B with different vigilance parameters and forcing a hierarchical structure by making $\rho_A > \rho_B$, such that module B enforces its categorization (an internal supervision) on module A.

\subsubsection{ArboART}
\label{Sec:ArboART}

ArboART~\cite{Ishihara1995} is an agglomerative hierarchical clustering method based on ART. More specifically, it uses ART 1.5-SSS (small sample size)~\cite{Ishihara.1993a} (variant of ART 1.5~\cite{Levine.1990a}, which in turn is a variation of ART 2~\cite{Carpenter.1987a}), as a building block. Briefly, prototypes of one ART are the inputs to another ART with looser vigilance (similarity constraint). Therefore, prototypes obtained from a lower level (bottom part of the dendrogram) are fed to the next ART layer. ART modules on higher layers have decreasingly lower vigilance values, i.e., the similarity constraint is less strict. This enables the construction of a tree (hierarchical graph structure). One of the advantages over traditional hierarchical methods is that it does not require a full recomputation when a new sample is added, only partial recomputations are needed in ART (inside the specific clusters). ArboART uses several layers of ART as well as one pass learning. Concretely, it makes super-clusters of previous clusters in a hierarchical way, thereby making a generalization of categories in the process. 

\subsubsection{Joining hierarchical ART}
\label{Sec:HARTJ}

The joining hierarchical ART (HART-J)~\cite{Bartfai1996} 
is a hierarchical agglomerative clustering method (bottom-up approach) that uses ART~1 modules (Sec.~\ref{Sec:ART1}) as building blocks and follows a cascade design. Specifically, each layer of this multi-layer model corresponds to an ART~1 network that clusters the prototypes generated by the preceding layer. The input of layer $l$ is given by: 
\begin{equation}
\bm{x}_{l} = \bm{x}_{1} \cap \left(\bigcap\limits_{i=1}^{l-1} \bm{w}_{i,J}\right),~l=\{1,...,L\},
\end{equation}
\noindent where $L$ is the number of layers, $\bm{x}_{1}$ is equal to the input sample $\bm{x}$, and $\bm{w}_{i,J}$ is the resonant neuron $J$ of layer $i$. Interestingly, it is not imperative to reduce the vigilance values at higher layers to generate the hierarchy: the ``effective'' vigilance level of layer $l$ is given by:
\begin{equation}
\hat{\rho_{l}} = \prod\limits_{j=1}^l \rho_j.
\end{equation}
\noindent which decreases even if the vigilance increases with $l$ given that $\rho_l \in [0,1]~\forall l$. This fact is used to derive an upper bound for the maximum number of layers $L_{max}$. If all vigilance values are equal to $\rho$, then $L_{max}= \lfloor n+1 \rceil$, where $n$ is the minimum integer that satisfies
\begin{equation}
n > -\frac{\log K}{\log \rho},
\end{equation}
\noindent assuming that $\norm{\bm{x}_l}_1=constant=K$.

Naturally, succeeding networks can learn at most the number of prototypes from the previous layer. Learning can occur in sequential (waiting for stabilization before the next layer starts learning) or parallel (learning occurs in each layer in each presentation of inputs) modes. The former generates fewer categories but the training time, measured in number of epochs, is much smaller using the parallel approach.

HART-J is compared to SMART in~\cite{Bartfai1995}. Contrary to SMART, HART-J has no associative connection or feedback between hierarchical layers as a mechanism to enforce self-consistency. The constraint causing lower layers to have larger vigilance values than the higher layers guarantees consistency. In HART-J, the hierarchies ``emerge'' since there are no explicit links. It is reported that SMART builds a less compact model (larger number of categories) due to categorization forced by its internal feedback mechanism, whereas HART-J builds a simpler and more compact network.

\subsubsection{Hierarchical ART with splitting}
\label{Sec:HARTS}

The hierarchical ART with splitting (HART-S)~\cite{Bartfai1997} consists of a cascade of ART 1 (Sec.~\ref{Sec:ART1}) modules that performs incremental hierarchical divisive clustering (successive splitting in a top-down approach). A fuzzy HART-S~\cite{Bartfai1997b} variant uses a cascade of fuzzy ARTs, where each module clusters the difference between the input and the weight vector of the resonant category belonging the preceding layer. Specifically, the input to layer $l+1$ ($l=\{1,...,L\}$, where $L$ is the maximum number of layers) is given by:
\begin{equation}
\bm{x}_{l+1} = \bm{x}_{1} \wedge \bm{w}_{l,J}^c,  
\end{equation}
\noindent which recursively corresponds to
\begin{equation}
\bm{x}_{l+1} = \bm{x}_{1} \wedge \left( \bigwedge\limits_{i=1}^l \bm{w}_i^c \right), 
\end{equation}
\noindent where $\bm{x}_1 = \bm{x}$ is the data sample and $\bm{w}_{l,J}^c$ is the complement of the weight vector associated with the resonant neuron~$J$ of layer~$l$.

The hierarchy is explicitly represented by links between parent and children categories in a tree-like structure. These adaptive associative connections between consecutive modules ensure that only children of the preceding parent module can be activated. In its most general case, the fuzzy ART modules in each layer have their own set of parameters. Particularly, Fuzzy HART-S uses two global parameters: a resolution parameter $\epsilon$ to control the depth of the hierarchical tree (i.e., if $|x_k| < \epsilon S$, then there is no more splitting, where $S$ is the maximum size of root/global input $x_1$) and a feature threshold parameter to control the propagation of features throughout the layers. 

Strategies to prune and rebuild prototypes to improve HART-S in terms of network complexity (measured by the number of categories) are presented in~\cite{Bartfai1998}. During learning, the former strategy removes small clusters (and all their children if applicable) based on a cluster size threshold (percentage of the total number of samples), and the latter changes the components of a prototype weight vector  to better reflect the samples associated with them. 

\subsubsection{Distributed dual vigilance fuzzy ART}

The distributed dual vigilance fuzzy ART (DDVFA)~\cite{leonardo.2018c} is a dual vigilance-based ART model designed to improve memory compression and perform several ART-based hierarchical agglomerative clustering (HAC) methods online. It consists of a global ART module whose \Fout \ nodes are local fuzzy ARTs: the global module is used for decision making while the local module builds multi-prototype representations of clusters (many-to-one mappings). 

The activation of a global ART \Fout \ node $i$ ($T_i^{g}$) is a function of the activations of the $k$ \Fout \ nodes of its corresponding local fuzzy ART module:
\begin{equation}
T_i^{g} = f\left( T_1^{i}~,~...~,~T_j^{i}~,~...~,~T_k^{i} \right),
\label{Eq:T1}
\end{equation}
\noindent where $T_j^{i}$ is the activation function of the \Fout \ node $j$ of the local fuzzy ART module~$i$, which uses a higher order activation function defined as
\begin{equation}
T_j^{i}  = \left( \frac{\|\bm{x} \wedge \bm{w}^i_j\|_1}{\alpha +\|\bm{w}^i_j\|_1} \right)^\gamma,
\label{Eq:T2}
\end{equation}
\noindent and $\gamma \geq 1$ is a power parameter whose role is akin to a kernel width. Similarly, the match function of a global ART \Fout \ node $i$ ($M_i^g$) is defined as
\begin{equation}
M_i^g = g\left( M_1^i~,~...~,~M_j^i~,~...~,~M_k^i \right),
\label{Eq:M1}
\end{equation}
\noindent where $M_j^i$ is the match function of the \Fout \ node $j$ of the local fuzzy ART module~$i$, which uses the following normalized higher order match function
\begin{equation}
M_j^i = \left( \frac{\|\bm{w}^i_j\|_1}{\|\bm{x}\|_1}\right)^{\gamma^{*}}T_j^{i},
\label{Eq:M3}
\end{equation}
\noindent where $0 \leq \gamma^{*} \leq \gamma $ is the reference kernel width with respect to which the match function is normalized. Both functions $f(\cdot)$ and  $g(\cdot)$ are based on HAC methods, as listed in Table~\ref{Tab:DDpART_T_M}.

The DDVFA features a dual vigilance mechanism: when a sample $\bm{x}$ is presented and the \Fout \ node~$I$ of the global ART is the winner, then $VR_I = \{\bm{x} : M_J^g(\bm{x}) \geq \rho_{LB}\}$ and $\rho_{LB} \in [0,1]$. The vigilance criterion checks if $\mathds{1}_{VR_I^g}(\bm{x})$ is true. If not, the search continues, or a new local fuzzy ART module is created. If so, the corresponding local fuzzy ART module is allowed to learn. The local Fuzzy ART module  imposes a stricter constraint for its winner nodes: $VR_J^I = \{\bm{x} : M_J^I(\bm{x}) \geq \rho_{UB}\}$ and $ 0 \leq \rho_{LB} \leq \rho_{UB} \leq 1$. Again, the vigilance criterion checks if $\mathds{1}_{VR_J^I}(\bm{x})$ is true, and, if so, the category is allowed to learn. Otherwise, the search resumes or a new node is created following the standard ART dynamics.

When input order cannot be addressed via an offline pre-processing strategy (Sec.~\ref{Sec:ordering_effects}), then DDVFA should be used in conjunction with a Merge ART module to mitigate input order dependency in online learning applications. This module is connected to DDVFA in series, i.e., in a cascade design. The inputs to Merge ART are fuzzy ART modules with all their corresponding categories. Like DDVFA, Merge ART's \Fout \ nodes are also fuzzy ART modules. When a DDVFA's fuzzy ART node $l$ is fed to Merge ART, an activation matrix $\bm{T}_{k,l} = [t_{i,j}]_{R \times C}$ (where $R$ and $C$ are the number of categories in Merge ART node $k$ and DDVFA node $l$, respectively) is computed as
\begin{equation}
t_{i,j} = \left( \frac{\| \bm{w}_j^l \wedge \bm{w}_i^k \|_1}{\alpha + \| \bm{w}_i^k \|_1} \right)^\gamma,
\label{Eq:tij}
\end{equation}
\noindent where $\bm{w}_j^l$ is the weight vector of category~$j$ of DDVFA local fuzzy ART module~$l$, and $\bm{w}_i^k$ is the weight vector of category~$i$ of Merge ART module~$k$. The actual activation of Merge ART node~$k$ uses matrix $\bm{T}_{k,l}$ and follows one of the HAC forms as listed in Table~\ref{Tab:MergeART_T_M}. Assuming Merge ART's \Fout \ node $k$ is the winner, its match matrix \mbox{$\bm{M}_{k,l} = [m_{i,j}]_{R \times C}$} is computed as
\begin{equation}
m_{i,j} = \left( \frac{\|\bm{w}_i^k\|_1}{\|\bm{w}_j^l\|_1} \right)^{\gamma^{*}} t_{i,j},
\label{Eq:wij}
\end{equation}
\noindent where the actual match of Merge ART node $k$ uses matrix $\bm{M}_{k,l}$ and also uses one of the formulations listed in Table~\ref{Tab:MergeART_T_M}. If the vigilance constraint is satisfied (i.e., $M_k \geq \rho_{LB}$), then \mbox{$ART_K(new) \leftarrow ART_K(old) \cup ART_l$}, i.e., the weights of both ART modules are concatenated. To further reduce model complexity, the final step of Merge ART consists of feeding the weight vectors of each ART module to an independent fuzzy ART parameterized with $\rho = \rho_{UB}$, $\gamma$ and $\gamma^{*}$. Note that the Merge ART module can be run once or until convergence, where the latter is defined as no change in the Merge ART nodes between two consecutive iterations.

\begin{table}[!t]
\centering
\caption{DDVFA's activation and match functions.}
\begin{threeparttable}
\begingroup\setlength{\fboxsep}{0pt}
\colorbox{lightgray}{
\begin{tabular*}{\columnwidth}{@{\extracolsep{\fill}}lll@{}}
\toprule
HAC method 
& $T_i^g=f(\cdot)$ 
& $M_i^g=g(\cdot)$ \\
\midrule
\midrule
single 	  
& $\max\limits_{j}\left( T^i_j \right)$	
& $\max\limits_{j}\left( M^i_j \right)$ \\
complete  
& $\min\limits_{j}\left( T^i_j \right)$	
& $\min\limits_{j}\left( M^i_j \right)$ \\
median    
& $\median\limits_{j}\left( T^i_j \right)$	
& $\median\limits_{j}\left( M^i_j \right)$ \\
average\tnote{a} 
& $\dfrac{1}{k_i}\sum\limits_{j=1}^{k_i} T^i_j$	
& $\dfrac{1}{k_i}\sum\limits_{j=1}^{k_i} M^i_j$ \\
weighted\tnote{b} 
& $\sum\limits\limits_{j=1}^{k_i} p_j T^i_j$	
& $\sum\limits\limits_{j=1}^{k_i} p_j M^i_j$ \\
centroid\tnote{c}  
& $\left(\dfrac{\norm{\bm{x} \wedge \bm{w}_c}_1}{\alpha + \norm{\bm{w}_c}_1}\right)^\gamma$   
& $\left( \dfrac{\norm{\bm{w}_c}_1}{\norm{\bm{x}}_1}\right)^{\gamma^{*}}T_i^g$ \\
\bottomrule
\end{tabular*}
}\endgroup
\begin{tablenotes}[normal,flushleft]
\item[a,b]$k_i$ is the number of \Fout \ nodes in local fuzzy ART module~$i$.
\item[b]$p_j=\dfrac{n^i_j}{n_i^g}$, where $n^i_j$ is the number of samples encoded by category $j$ of local fuzzy ART module $i$, and $n_i^g=\sum\limits_jn^i_j$.
\item[c]$\bm{w}_c$ is a centroid, whose $l$ component is computed as $w_{c,l}=\min\limits_{j}\left( w_{j,l}\right)$ for \mbox{$l=\{1,...,2d\}$}.
\end{tablenotes}
\hrulefill
\end{threeparttable}
\label{Tab:DDpART_T_M}
\end{table}

\begin{table}[!t]
\centering
\caption{Merge ART's activation and match functions.}
\begin{threeparttable}
\begingroup\setlength{\fboxsep}{0pt}
\colorbox{lightgray}{
\begin{tabular*}{\columnwidth}{@{\extracolsep{\fill}}lll@{}}
\toprule
Method 
& $T_k=f(\cdot)$ 
& $M_k=g(\cdot)$ \\
\midrule
\midrule
single 	  
& $\max\limits_{i,j}\left( [t_{ij}] \right)$	
& $\max\limits_{i,j}\left( [m_{ij}] \right)$ \\
complete  
& $\min\limits_{i,j}\left( [t_{ij}] \right)$	
& $\min\limits_{i,j}\left( [m_{ij}] \right)$ \\
median    
& $\median\limits_{i,j}\left( [t_{ij}] \right)$	
& $\median\limits_{i,j}\left( [m_{ij}] \right)$ \\
average 
& $\dfrac{1}{RC}\sum\limits_{i=1}^{R}\sum\limits_{j=1}^{C} t_{ij}$	
& $\dfrac{1}{RC}\sum\limits_{i=1}^{R}\sum\limits_{j=1}^{C} m_{ij}$ \\
weighted\tnote{a}  
& $\sum\limits_{i=1}^{R}\sum\limits_{j=1}^{C} p_ip_jt_{ij}$		
& $\sum\limits_{i=1}^{R}\sum\limits_{j=1}^{C} p_ip_jm_{ij}$ \\
centroid\tnote{b}   
& $\left(\dfrac{\|\bm{w}^k_c \wedge \bm{w}^l_c\|_1}{\alpha +\|\bm{w}^k_c\|_1}\right)^\gamma$   
& $\left(\dfrac{\norm{\bm{w}^k_c}_1}{\norm{\bm{w}^l_c}_1}\right)^{\gamma^{*}}T_k$  \\
\bottomrule
\end{tabular*}
}\endgroup
\begin{tablenotes}[normal,flushleft]
\item[a]$p_i=\dfrac{n_i^k}{n_k}$ and $p_j=\dfrac{n_j^l}{n_l}$, where $n^i_k$ is the number of samples encoded by category $i$ of Merge ART node $k$, and \mbox{$n_k=\sum\limits_i n_i^k$}. The variables $n_j^l$ and $n_l$ refer to DDVFA node $l$ and are defined similarly.
\item[b]$\bm{w}^k_c$ and $\bm{w}^l_c$ are the centroids representing all categories of $ART^{(2)}_k$ and $ART^{(1)}_l$, respectively. Their components are given by $w^k_{c,n}=\min\limits_{j}\left( w^k_{j,n}\right)$ and $w^l_{c,n}=\min\limits_{j}\left( w^l_{j,n}\right)$, where $n=\{1,...,2d\}$.
\end{tablenotes}
\hrulefill
\end{threeparttable}
\label{Tab:MergeART_T_M}
\end{table}

\subsection{Biclustering and data fusion architectures}

\subsubsection{Fusion ART}
\label{Sec:Fusion_ART}

Fusion ART~\cite{tan2007} extends ART capabilities by augmenting it with multiple and independent \Fin\ layers (or input channels/field), all of which are connected to a shared \Fout\ layer. This model is then capable of learning mappings across multiple channels simultaneously.

\textbf{Activation.} 
The activation function of a category $j$ is a weighted sum of the activation functions of each input field
\begin{equation}
T_j = \sum\limits_{k=1}^K \gamma^k \dfrac{ \| \bm{x}^k  \wedge \bm{w}_j^k \|_1 }{\alpha^k + \| \bm{w}_j^k \|_1},
\label{Eq:fusion_1}
\end{equation}
\noindent where $\bm{x}^k$ is the complement coded input to the $k^{th}$ $F_1$ layer ($F_1^k$ or channel $k$), and $\gamma^k \in [0,1]$ and $\alpha^k \in (0,\infty)$ are the contribution and choice parameters of $F_1^k$, respectively. The variable $K$ is the total number of input channels such that \mbox{$\bm{x}=[\bm{x}^1,...,\bm{x}^k,...,\bm{x}^K]$} and category $j$'s LTM is \mbox{$\bm{w}_j=[\bm{w}_j^1,...,\bm{w}_j^k,...\bm{w}_j^K]$}.

\textbf{Match and resonance.} 
When category $J$ is selected by the WTA competition, one match function is computed for each channel
\begin{equation}
M_J^k = \frac{\|\bm{y}^{(F_1^k)} \|_1}{\|\bm{x}^k\|_1} = \frac{\|\bm{x}^k \wedge \bm{w}_J^k\|_1}{\|\bm{x}^k\|_1},
\label{Eq:fusion_2}
\end{equation}
\noindent where $VR_J^k = \{\bm{x} : M_J^k(\bm{x}) \geq \rho^k\}$, and $\rho^k \in [0,1]$ is $F_1^k$'s vigilance parameter. The vigilance test must be satisfied for all input fields simultaneously. Otherwise, a mismatch triggers a category reset and a match tracking procedure takes place. Particularly, the global vigilance criterion is satisfied if all channels meet their individual vigilance criteria, i.e., if $\bigwedge\limits_{k=1}^K \mathds{1}_{VR_J}^k(\bm{x})=1$. If this condition is not satisfied, fusion ART's match tracking mechanism simultaneously raises all vigilance parameters until a mismatch is triggered in one of the channels. The search then continues until a resonant category is found or created. Then, learning takes place as
\begin{equation}
\bm{w}_J^k(new) = (1-\beta^k)\bm{w}_J^k(old) + \beta^k(\bm{x}^k \wedge \bm{w}_J^k(old)),~\forall k,
\label{Eq:fusion_4}
\end{equation}
\noindent where $\beta^k \in (0,1]$ is the learning parameter of layer $F_1^k$. When a new input is presented, \mbox{$\rho^k = \bar{\rho}_k$}, where $\bar{\rho}_k$ is the baseline vigilance of layer $F_1^k$. Additionally, if an input to a channel is not present, then it is set to~$\Vec{\bm{1}}$ to enable the prediction/recovery of missing values. 

Notably, fusion ART generalizes some other ART models, i.e., by appropriately designing fusion ART, it can reduce to different ART models and perform distinct machine learning modalities: (i) 1 channel (samples) fusion ART reduces to ART~\cite{Carpenter1991} (Sec.~\ref{Sec:FA}) and performs match-based unsupervised learning, (ii) 2 channels (samples and class labels) fusion ART reduces to adaptive resonance associative map - ARAM~\cite{Tan1995b} (Sec.~\ref{Sec:ARAM}) and performs association-based supervised learning and (iii) 3 channels (states, actions and rewards) fusion ART reduces to fusion architecture for learning, cognition, and navigation - FALCON~\cite{tan2004} (Secs.~\ref{Sec:RFALCON} and~\ref{Sec:TDFALCON}) and performs reinforcement learning. Additionally, fusion ART can perform instruction-based learning by rule-based knowledge integration (generation of IF-THEN rules mapping antecedents and consequents from one channel to another, and rule insertion capability). 

\subsubsection{Biclustering ARTMAP}
\label{Sec:BARTMAP}

Biclustering ARTMAP (BARTMAP) ~\cite{xu2011, Xu.2012a} is based on fuzzy ARTMAP~\cite{carpenter1992} (Sec.~\ref{Sec:FAM}) and was designed to find correlation-based subspace clustering. It uses two Fuzzy ART modules (\ARTa\ and \ARTb) connected through a regulatory inter-ART module to achieve a biclustering of the data matrix on both the input space (rows) and the feature space (columns). The \ARTb\ module is used to cluster the feature vectors and create a set of feature clusters. Then, the samples are presented to the \ARTa\ module while using the inter-ART module to integrate the clustering results on both the feature and input spaces and create biclusters that capture the local relations between the inputs and features. Note that BARTMAP learns in offline mode. This architecture was shown to perform fast and stable biclustering of gene expression data~\cite{xu2011} and later modified to build a collaborative filtering recommendation system~\cite{Elnabarawy2016}.

The BARTMAP algorithm begins by presenting all the feature vectors to \ARTb\ (which is a standard fuzzy ART module), using it to build clusters of the feature vectors. Next, it begins presenting the input vectors to \ARTa\ and allows it to build clusters in the input space. If \ARTa\ places an input in a previously committed category, the inter-ART module then computes the similarity between the new sample and the samples in the existing cluster, but only within each feature cluster from \ARTb, thereby testing the correlation between the new sample and each of the existing biclusters. If any of the biclusters passes a user-defined correlation threshold $\eta$, the cluster is updated with the new sample. However, if none of the current biclusters passes, the \ARTa\ vigilance threshold is temporarily increased (match tracking mechanism, see Sec.~\ref{Sec:AM}), and the sample is presented again to find a new cluster. If no suitable cluster is found that also satisfies the correlation threshold, the \ARTa\ vigilance will eventually be increased enough to force the creation of a new cluster.

Consider the data matrix $\bm{X} = [x_{i,j}]_{N \times d}$, encompassing $N$ samples in a $d$-dimensional feature space. After \ARTb\ detects $N_b$ clusters of features, the $k^{th}$ input to \ARTa\ becomes \mbox{$\bm{x}_k = [\bm{x}_k^{c_1^b},...,\bm{x}_k^{c_i^b},...,\bm{x}_k^{c_{N_b}^b}] \in \mathbb{R}^d$}, where $\bm{x}_k^{c_i^b}$ comprises the subset of components of $\bm{x}_k$ associated with the $i^{th}$ feature cluster identified by \ARTb\ ($c_i^b$). The similarity between the input sample $\bm{x}_k$ and an \ARTa\ cluster $c_j^a$ with $n_j^a$ samples, across an \ARTb\ feature cluster $c_i^b$ with $n_i^b$ features, is defined using the average Pearson correlation coefficient~\cite{Bain1992} as follows:
\begin{equation}
\bar{r}_{c_j^a, c_i^b}(\bm{x}_k) = \dfrac{1}{n_j^a} \sum\limits_{l=1,\bm{x}_l \in c_j^a}^{n_j^a} r_{c_j^a, c_i^b}(\bm{x}_k^{c_i^b}, \bm{x}_l^{c_i^b}),
\end{equation}
\noindent where
\begin{equation}
r_{c_j^a, c_i^b}(\bm{x}_k^{c_i^b}, \bm{x}_l^{c_i^b}) = \dfrac{\sum\limits_{t = 1}^{n_i^b} (x_{k,t}^{c_i^b} - \bar{x}_{k}^{c_i^b}) (x_{l,t}^{c_i^b} - \bar{x}_{l}^{c_i^b})}{\sqrt{\sum\limits_{t = 1}^{n_i^b}(x_{k,t}^{c_i^b} - \bar{x}_{k}^{c_i^b})^2}\sqrt{\sum\limits_{t = 1}^{n_i^b}(x_{l,t}^{c_i^b} - \bar{x}_{l}^{c_i^b})^2}}.
\label{eq:correlation}
\end{equation}

Here, $x_{m,t}^{c_i^b}$ refers to the value for sample $\bm{x}_m$ at feature $t$ within the \ARTb\ cluster $c_i^b$ ($m=k,l$). Similarly, $\bar{x}_{m}^{c_i^b}$ denotes the average value of $\bm{x}_m$ across all the features in \ARTb's cluster $c_i^b$:
\begin{equation}
\bar{x}_{m}^{c_i^b} = \dfrac{1}{n_i^b} \sum\limits_{t = 1}^{n_i^b} x_{m,t}^{c_i^b}.
\end{equation}

\subsubsection{Generalized heterogeneous fusion ART}
\label{Sec:GHF_ART}

The generalized heterogeneous fusion ART (GHF-ART)~\cite{Meng2014} is a model designed to perform co-clustering of heterogeneous data (i.e., mixed data types). It extends the heterogeneous fusion ART (HF-ART)~\cite{Meng.2012a}, which is a two-channels fusion ART-based model, to a multiple channel architecture. The distinctive characteristic of the generalized heterogeneous fusion ART is that its learning functions vary according to each data type, i.e., when a winner node~$J$ satisfies the vigilance criterion, different channels are adapted following different learning functions~$f_L^k(\cdot)$. For instance, if the input $\bm{x}^k$ corresponds to a visual feature from image data or a text feature from a document, then the corresponding weight vector is updated following Eq.~(\ref{Eq:fusion_4}). Alternately, if $\bm{x}^k$ is a feature from data meta-information, then the weight vector of the corresponding channel $k$ is adapted using the recursive mean formula 
\begin{equation}
\bm{w}_J^k(new) =  \left(1-\frac{1}{n_J(new)}\right)\bm{w}_J^k(old) +\frac{1}{n_J(new)} \bm{x}^k,
\label{Eq:GHF_ART_1}
\end{equation}
\begin{equation}
n_J(new)  =  n_J(old) + 1,
\label{Eq:GHF_ART_2}
\end{equation}
\noindent where $n_J$ corresponds to the number of samples encoded by node $J$.

Another key characteristic of the generalized heterogeneous fusion ART is the adaptive channel weighting: the contribution parameters are initially uniformly initialized, and then, during learning, undergo self-adaptation using
\begin{equation}
\gamma^k(new) = \dfrac{R^k}{\sum\limits_{k=1}^K R^k},~\forall k,
\label{Eq:GHF_ART_3}
\end{equation}
\noindent where 
\begin{equation}
R^k = exp\left(-\frac{1}{N}\sum_{j=1}^N D_j^k \right),
\label{Eq:GHF_ART_4}
\end{equation}
\begin{equation}
D_j^k = \frac{\frac{1}{n_j}\sum\limits_{l=1}^{n_j} \norm{\bm{w}_j^k - \bm{x}_l^k}_1}{\norm{\bm{w}_j^k}_1}.
\label{Eq:GHF_ART_5}
\end{equation}

The variable $R$ is a robustness measure used to estimate the discriminative power of each channel given the intra-cluster scatter. In practice, performing the offline computations in Eq.~(\ref{Eq:GHF_ART_5}) can be expensive. Therefore, since only $D_J^k$ needs to be updated after the presentation of each sample, then $\gamma^k(new)$ can be estimated incrementally. Particularly, when there is a resonant committed node $J$, if $\bm{x}^k$ is a meta-information feature, then 
\begin{equation}
\begin{split}
D_J^k(new) & = \dfrac{n_J(old)}{n_J(new)\norm{\bm{w}_J^k(new)}_1} \\
& \left( \norm{\bm{w}_J^k(old)}_1 D_J^k(old)  
 -  \norm{\bm{w}_J^k(new) - \dfrac{n_J(old)}{n_J(new)}\bm{w}_J^k(old) }_1
+ \dfrac{1}{n_J(old)} \norm{\bm{w}_J^k(new) - \bm{x}^k}_1 \right),
\end{split}
\label{Eq:GHF_ART_6}
\end{equation}
\noindent otherwise,
\begin{equation}
\begin{split}
D_J^k(new) & = \dfrac{n_J(old)}{n_J(new)\norm{\bm{w}_J^k(new)}_1}  \\
& \left( \norm{\bm{w}_J^k(old)}_1 D_J^k(old) 
 -  \norm{\bm{w}_J^k(old) - \bm{w}_J^k(new)}_1 
 +  \dfrac{1}{n_J(old)} \norm{\bm{w}_J^k(new) - \bm{x}^k}_1 \right).
\end{split}
\label{Eq:GHF_ART_7}
\end{equation}

If a new category is created, regardless of $\bm{x}^k$ type, the contribution parameters are updated via a proportionality change  
\begin{equation}
\gamma^k(new) = \dfrac{\left(R^k\right)^{\frac{N}{N+1}}}{\sum\limits_{k=1}^K \left(R^k\right)^{\frac{N}{N+1}}},~\forall k,
\label{Eq:GHF_ART_8}
\end{equation}
\noindent where $N$ is the number of categories.

Note that the generalized heterogeneous fusion ART can also include prior knowledge by appropriate initialization of the network.

\subsubsection{Hierarchical Biclustering ARTMAP}
\label{Sec:HBARTMAP}

Hierarchical Biclustering ARTMAP (H-BARTMAP)~\cite{Kim2016} uses BARTMAP (\ref{Sec:BARTMAP}) iteratively to obtain a hierarchy of biclusters. The algorithm begins by running BARTMAP on the complement coded data with low vigilance values, which produces a relatively small number of larger-sized biclusters. In the following step, H-BARTMAP uses a bicluster matching threshold and a correlation fitness function to build and evaluate the biclusters at the current level. After that, the BARTMAP algorithm is used again on each of the resulting clusters with increased  vigilance and correlation thresholds. These are adjusted by small values that are a function of the number of samples as well as the number of features and average correlation in each bicluster. The H-BARTMAP algorithm repeats those two steps recursively for a specified number of times. Then, the best layer in the recursive tree that optimizes the desired cluster validity index~\cite{xu2009} or any other user-specified criteria is chosen.

\subsection{Summary}

Table~\ref{Tab:BasicARTs} summarizes the nature of the category representations of the ART elementary models described in the previous subsections, during activation, match and learning stages. Particularly, it lists if winner-takes-all (WTA) or distributed (D) coding is employed by these networks.

\begin{table*}[!ht]
\centering
\caption{Summary of the code representations used by the unsupervised learning ART models.}
\begin{threeparttable}
\begingroup\setlength{\fboxsep}{0pt}
\colorbox{lightgray}{
\begin{tabular*}{\textwidth}{@{\extracolsep{\fill}}lllll@{}}
\toprule
ART model & Activation & Match & Learning & Reference(s) \\
\midrule
\midrule
ART 1                   & WTA & WTA & WTA & \cite{Carpenter1987} \\
ART 2-A                 & WTA & WTA & WTA & \cite{Carpenter1991b} \\
Fuzzy ART               & WTA & WTA & WTA & \cite{Carpenter1991} \\
Fuzzy Min-Max           & WTA & WTA & WTA & \cite{Simpson1993} \\
ARTtree                 & WTA & WTA & WTA & \cite{Wunsch1993} \\
SMART                   & WTA & WTA & WTA & \cite{Bartfai1994} \\
ArboART                 & WTA & WTA & WTA & \cite{Ishihara1995} \\
Distributed ART         & D   & D   & D   & \cite{Carpenter.1996a, Carpenter.1996b, Carpenter.1997a} \\
Gaussian ART            & WTA & WTA & WTA & \cite{williamson1996} \\
HART-J/S                & WTA & WTA & WTA & \cite{Bartfai1996, Bartfai1997} \\
Hypersphere ART         & WTA & WTA & WTA & \cite{anagnostopoulos2000} \\
Ellipsoid ART           & WTA & WTA & WTA & \cite{anagnostopoulos2001, anagnostopoulos2001b} \\
Quadratic neuron ART    & WTA & WTA & WTA & \cite{ChunSu2002,ChunSu2005} \\
Bayesian ART            & WTA & WTA & WTA & \cite{vigdor2007} \\
Fusion ART              & WTA & WTA & WTA & \cite{tan2007} \\
Fuzzy ART-GL            & WTA & WTA & WTA & \cite{Isawa2007} \\
GramART                 & WTA & WTA & WTA & \cite{Meuth2009} \\
TopoART                 & WTA & WTA & D   & \cite{Tscherepanow2010} \\
BARTMAP                 & WTA & WTA & WTA & \cite{xu2011} \\
GH Fusion ART           & WTA & WTA & WTA & \cite{Meng2014} \\
Hierarchical BARTMAP    & WTA & WTA & WTA & \cite{Kim2016} \\
CVIFA                   & WTA & WTA & WTA & \cite{leonardo2017} \\
DVFA                    & WTA & WTA & WTA & \cite{leonardo.2018b} \\
DDVFA                   & D   & D   & WTA & \cite{leonardo.2018c} \\
\bottomrule
\end{tabular*}
}\endgroup
\begin{tablenotes}[normal,flushleft]
\item[]WTA: winner-takes-all code.
\item[]D: distributed code.
\end{tablenotes}
\end{threeparttable}
\label{Tab:BasicARTs}
\end{table*}

\section{ART models for supervised learning}
\label{Sec:SL}

\subsection{Architectures for classification}

\begin{figure*}[!t]
\centerline{
\includegraphics[width=\textwidth]{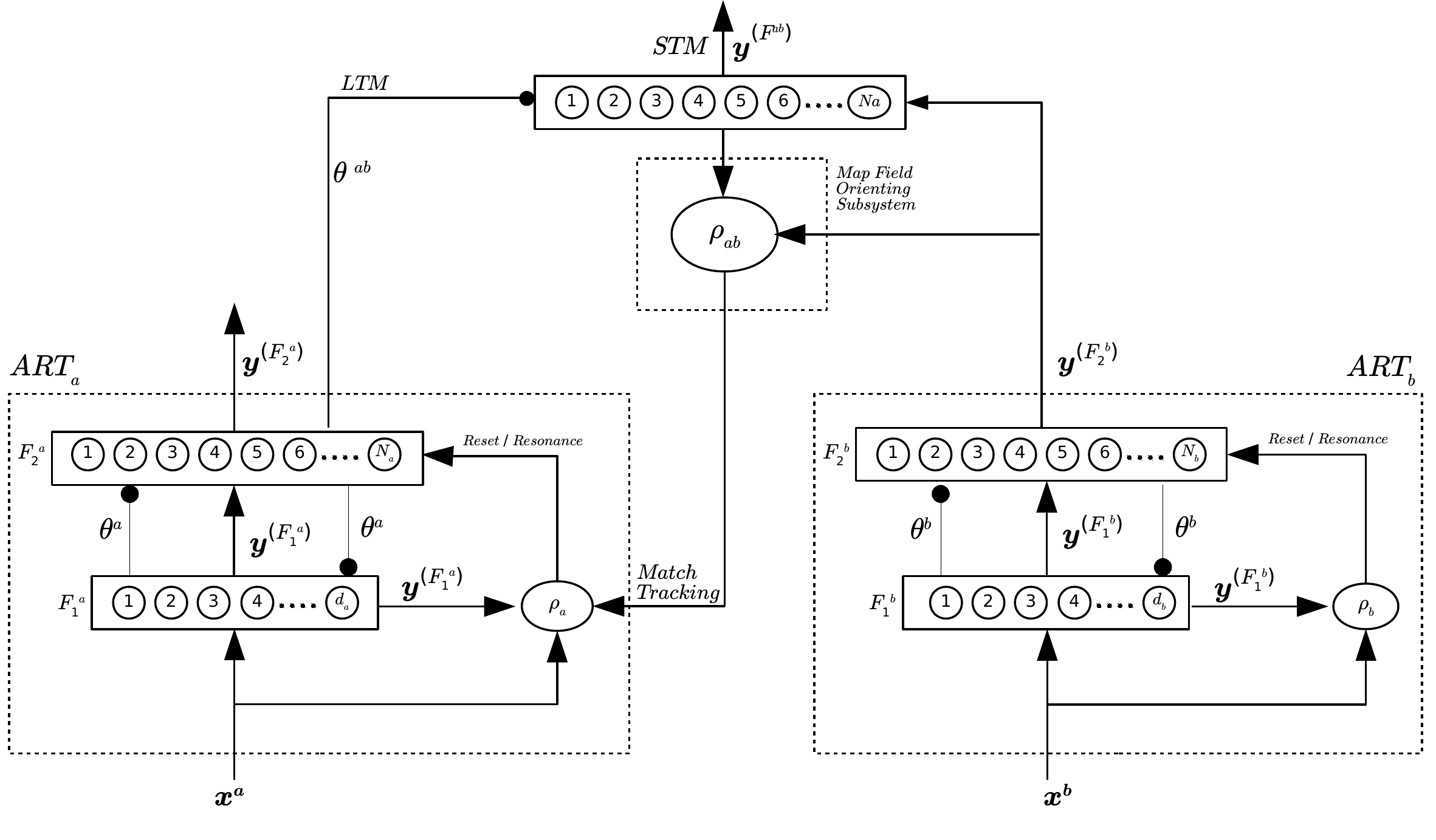}}
\caption{Elementary ARTMAP model.}
\label{Fig:gen_ARTMAP}
\end{figure*}

ART models used for supervised learning applications typically follow an ARTMAP architecture (Fig.~\ref{Fig:gen_ARTMAP}), which consists of two elementary ART units (\ARTa \ and \ARTb) interconnected by an associative learning network, namely the map field, that performs multidimensional mappings between categories of both such units, as well as allowing for associative recalls when the input to one of the ART modules is missing. Notably, ARTMAP models usually inherit the properties of their elementary ART building blocks. This section describes the main characteristics of members of the supervised ART family in terms of their map field  LTM units, dynamics (which encompasses activation, match, resonance criterion and learning) and user-defined parameters. For clarity, Table~\ref{Tab:Notation_SL} summarizes the notation used in the following subsections.

\begin{table}[!ht]
\centering
\caption{Supervised ART models notation.}
\begin{threeparttable}
\begingroup\setlength{\fboxsep}{0pt}
\colorbox{lightgray}{
\begin{tabular*}{\columnwidth}{@{}ll@{\extracolsep{\fill}}}
\toprule
Notation & Description \\
\midrule
\midrule
$\bm{x}^l$          & input sample to ART\textsubscript{$l$} \\
$d_l$               & input data dimensionality ($\bm{x}^l \in \mathbb{R}^{d_l}$)\\
$F_1^l$             & feature representation field of ART\textsubscript{$l$}\\
$F_2^l$             & category representation field of ART\textsubscript{$l$}\\
F\textsuperscript{ab}   & map field \\
$\bm{y}^{(F_1^l)}$  & $F_1^l$ activity (STM) \\
$\bm{y}^{(F_2^l)}$  & $F_2^l$ activity (STM) \\
$N_l$               & number of categories in ART\textsubscript{$l$} \\
$c^l$               & a category in ART\textsubscript{$l$} \\
$\bm{y}^{(F^{ab})}$ & F\textsuperscript{ab} activity (STM) \\
$\bm{\theta}^{ab}$  & map field parameters (LTM unit) \\
$M^{ab}$            & map field match function \\
$J$                 & \ARTa \ chosen category index (via WTA) \\
$K$                 & \ARTb \ chosen category index (via WTA) \\
$\rho_l$            & vigilance parameter of ART\textsubscript{$l$} \\
$\bar{\rho}$        & \ARTa \ baseline vigilance parameter \\
\bottomrule
\end{tabular*}
}\endgroup
\begin{tablenotes}[normal,flushleft]
\item[] Variable $l$ indexes the elementary ART module: $l=a,b$.
\end{tablenotes}
\end{threeparttable}
\label{Tab:Notation_SL}
\end{table}

When an ARTMAP architecture is used for pattern recognition or classification tasks, typically \ARTa\ clusters data samples while \ARTb\ clusters class labels in parallel. Therefore, while ART maps samples to categories, an ARTMAP architecture goes one step further and maps categories to classes. During training, \ARTa\ is subjected to a certain level of agreement with \ARTb's activity, given that the latter is encodes the target labels. This is performed by a second vigilance test that uses \ARTb's supervisory signal (i.e., response) to trigger a mismatch or allow learning given an incorrect or correct prediction, respectively. Specifically, when \ARTa's prediction is disproven by \ARTb's, then the map field triggers a match tracking mechanism in which \ARTa's resonating category is inhibited, the baseline vigilance is temporarily changed and the search process restarts, causing \ARTa\ to select another category. Therefore, the map field is a critic, i.e., its purpose is to assess the quality of the mapping between both ART modules and also the necessity of adding a new node based on a supervised signal. Specifically, by engaging the match tracking mechanism, ARTMAP trades generalization for specificity to decrease training error. 
Often, \ARTb \ is  omitted and an $N_b$-dimensional vector of labels is used in its place (since \ARTb's vigilance parameter would typically be set to~$1$, which would correspond to the number of categories being equal to the number of classes). Moreover, \ARTa's baseline vigilance parameter, which controls the granularity of the input space, is usually set to small values since this correlates with improved generalization capabilities and a higher level of compression, i.e., network complexity. Algorithm~\ref{Alg:ARTMAP} summarizes the dynamics of an elementary ARTMAP model.

\begin{algorithm}[!ht]
\DontPrintSemicolon
\SetKwInOut{Input}{Input}\SetKwInOut{Output}{Output}
\BlankLine
\Input{$\{\bm{x}^a,\bm{x}^b\}$, \{\ARTa \ and \ \ARTb \ parameters\}, $\{\bm{\beta}_{ab}, \bm{\gamma}_{ab}, \bm{\rho}_{ab}, \bm{\lambda}_{ab}\}$ (map field parameters).} 
\Output{$\bm{y}^{(F^{ab})}$ (map field activity).}
\algrule
\tcc{Notation\;
$\mathcal{C}_l$: set of ART\textsubscript{$l$} nodes ($l=a,b$). \;
$\bm{\theta}^{ab}$: map field LTM unit.\; 
$\bm{\beta}_{ab}$: map field learning function parameter(s).\; 
$\bm{\gamma}_{ab}$: map field match function parameter(s).\; 
$\bm{\rho}_{ab}$: map field vigilance parameter(s).\; 
$\bm{\lambda}_{ab}$: map field initialization parameter(s).\; 
$f_M^{ab}(\cdot)$: map field match function.\;
$f_L^{ab}(\cdot)$: map field learning function.\;
$f_V^{ab}(\cdot)$: map field vigilance function.\;
$f_N^{ab}(\cdot)$: map field initialization function.\;
$f_I^{ab}(\cdot)$: map field inference function.
}
\tcc{Training}
\nl \label{step1AM}Present input $\bm{x}^b \in \bm{X}^b$ to \ARTb.\;
\nl Perform the dynamics of \ARTb \ and find its resonating category $K$ (Alg.~\ref{Alg:ART}). \;
\nl \label{stepMT}Present input $\bm{x}^a \in \bm{X}^a$ to \ARTa. \;
\nl Perform the dynamics of \ARTa \ and find its resonating category $J$ (Alg.~\ref{Alg:ART}). \;
\nl Compute the map field match function: $M_J^{ab} = f_M^{ab}(J,K, \bm{\theta}^{ab}, \bm{\gamma}_{ab})$. \;
\nl Perform the map field vigilance test: $V_J = f_V^{ab} = \mathds{1}_{VR_J^{ab}}(\bm{x}^a)$. \;
\nl \uIf{$V_J$ is TRUE}{
\nl Update \ARTa's and \ARTb's categories $J$ and $K$ (Alg.~\ref{Alg:ART}).\;

\nl \uIf{\ARTa \ OR \ARTb \ created a new node}{
\nl $\theta^{ab}_{|\mathcal{C}_a|+1} = f_N^{ab}(J,K,\bm{\lambda}_{ab})$.\;
}
\nl \Else{ 
\nl Update the map field: $\theta^{ab}_J(new) = f_L^{ab}(\bm{x},\bm{\theta}_J^{ab}(old), \bm{\beta}_{ab})$.\;
}
}
\nl \Else{ 
\nl Inhibit \ARTa's category $J$ for $\bm{x}^a$.\;
\nl Go to step~\ref{stepMT}.\;
}
\nl Go to step~\ref{step1AM}.\;
\tcc{Inference}
\nl Present input $\bm{x}^a \in \bm{X}^a$ to \ARTa. \;
\nl Perform the dynamics of \ARTa \ (Alg.~\ref{Alg:ART}). \;
\nl Compute the degree of association to each \ARTb \ node $k$ according to \ARTa's activity(s): $\sigma_k = f_I^{ab}(\bm{y}^{F_2^a},\bm{\theta}^{ab})$. \;
\nl Set output:
$y_j^{(F^{ab})} = 
\begin{cases} 
1, & \mbox{if } j = \argmax\limits_k(\sigma_k)\\ 
0, & \mbox{otherwise}
\end{cases}$.\;
\caption{Elementary ARTMAP algorithm.}
\label{Alg:ARTMAP}  
\end{algorithm}

\subsubsection{ARTMAP}
\label{Sec:AM}

The first adaptive resonance theory supervised predictive mapping (predictive ART or ARTMAP) model~\cite{Carpenter1991a} consists of two binary ART~1 modules (Section~\ref{Sec:ART1}), \ARTa\ and \ARTb, connected via an inter-ART associative memory, namely the map field \Fab. The latter performs multidimensional mappings between the binary input samples clustered by modules A and B. Moreover, when the input of a module is missing, it can be recalled by such associative memory. The map field LTM is represented by a matrix \mbox{$\bm{W}^{ab} = [w^{ab}_{ij}]_{N_a \times N_b}$} such that $w^{ab}_{ij}=1$ if there is an association between category~$i$ of \ARTa\ and category~$j$ of \ARTb\ and zero otherwise, and $N_a$ and $N_b$ are the number of nodes in \ARTa\ and \ARTb, respectively. The matrix $\bm{W}^{ab}$ is initialized as $\bm{1}$ (i.e., the row vector $\bm{w}^{ab}_1 = \Vec{\bm{1}}$). The bottom-up and top-down weight vectors of both ART~1's are initialized as described in Section~\ref{Sec:ART1}.

\textbf{Training.} 
The map field F\textsuperscript{ab} activity is defined as
\begin{equation}
\bm{y}^{{(F^{ab})}} = 
\begin{cases} 
\bm{y}^{{(F_2^{b})}} \cap \bm{w}_J^{ab}, & \mbox{if both ARTs are active (training)}\\
\bm{w}_J^{ab}, & \mbox{if only ART\textsubscript{a} is active (prediction)} \\
\bm{y}^{{(F_2^{b})}}, & \mbox{if only ART\textsubscript{b} is active} \\
\Vec{\bm{0}}, & \mbox{otherwise} \\
\end{cases}.
\label{Eq:AM_1}
\end{equation} 
\noindent where \mbox{$\bm{w}_J^{ab} = (w_{J1},...,w_{JN_{b}})$} is the $J^{th}$ row of $\bm{W}^{ab}$, which is associated with ART\textsubscript{a}'s resonant category~$J$. 

After resonant nodes for both ART modules have been selected following the presentation of a sample pair $(\bm{x}^a, \bm{x}^b)$, the map field match function is computed as
\begin{equation}
M_J^{ab} = \frac{\|\bm{y}^{{(F^{ab})}}\|_1}{\|\bm{y}^{{(F_2^{b})}}\|_1} = \frac{\| \bm{y}^{{(F_2^{b})}} \cap \bm{w}_J^{ab} \|_1}{\|\bm{y}^{{(F_2^{b})}}\|_1},
\label{Eq:AM_2}
\end{equation}
\noindent where the vigilance test is satisfied if $M_J^{ab} \geq \rho_{ab}$. During training, if \ARTa's prediction is correct (i.e., confirmed by \ARTb's supervised signal feedback), all three modules learn. Otherwise, a match tracking mechanism (MT+) is engaged, such that ART\textsubscript{a}'s vigilance parameter is temporarily raised by an amount small enough to inhibit the resonant category 
\begin{equation}
\rho_a = M_J^a + \epsilon,~0<\epsilon \ll 1,
\label{Eq:AM_3}
\end{equation}
\noindent and the search process restarts. Either another resonant category is found or a new one is created, and the vigilance returns to its baseline value ($\rho_a = \bar{\rho}_a$) upon the presentation of a new input pair. Complement coding is usually employed to avoid cases in which \ARTa's vigilance is raised to a value greater than one.

Now consider that the resonant categories of \ARTa\ and \ARTb\ are $J$ and $K$, respectively. When the map field vigilance test is satisfied ($M_J^{ab} \geq \rho_{ab}$), then \ARTa\ and \ARTb\ are updated as described in Sec.~\ref{Sec:ART1}, and the map field weight vector associated with category $J$ is updated as
\begin{equation}
w_{Jk}^{ab}(new) = \bm{y}^{(F_2^b)} \cap \bm{w}_J^{ab}(old) =
\begin{cases} 
1, & \mbox{if } k=K \\ 
0, & \mbox{otherwise}
\end{cases} 
\label{Eq:AM_4}
\end{equation}
\noindent such that it becomes permanently associated with ART\textsubscript{b}'s category $K$. Note that the \Foa, \Fta\ and \Fab\ layers may be viewed as input, hidden and output layers, respectively.

\textbf{Inference.} 
In prediction mode, it is sufficient to track the map field's weight vector $\bm{w}_J^{ab}$ and set it as the systems' output, i.e., when an \ARTa's resonant category~$J$ is found, the predicted class $K$ is obtained as
\begin{equation}
K = \argmax\limits_{k} \left(\sigma_k \right),    
\label{Eq:AM_5}
\end{equation}
\noindent 
\noindent where
\begin{equation}
\sigma_k = \sum\limits_{j=1}^{N_a} w^{ab}_{jk}y_j^{(F_2^a)}.
\label{Eq:AM_5a}
\end{equation}

A simplified ARTMAP version, namely the simple ARTMAP~\cite{Gotarredona.1998a}, replaces \ARTb\ (and thus its \Ftb\ activity  $\bm{y}^{(F_2^b)}$) with a binary vector $\bm{y}^b$ indicating the class membership of the input sample $\bm{x}^a$ (i.e., $y^b_k=1$ if $\bm{x}^a$ belongs to class $k$, and $y^b_i=0$ $\forall i \neq k$).

\subsubsection{Fuzzy ARTMAP}
\label{Sec:FAM}

Fuzzy ARTMAP (FAM)~\cite{carpenter1992} is to ARTMAP what fuzzy ART is to ART 1: it extends the capabilities of ARTMAP to enable the processing of real-valued data by replacing logical with fuzzy AND intersection. Thus, fuzzy ARTMAP also consists of two fuzzy ART modules, \ARTa\ and \ARTb, connected by a map field \Fab\ that maps the categories of one ART to another via a matrix of weights $\bm{W}^{ab}$, as described in Sec.~\ref{Sec:AM}. 

\textbf{Training.} 
The map field \Fab\ activity is defined as
\begin{equation}
\bm{y}^{{(F^{ab})}} = 
\begin{cases} 
\bm{y}^{{(F_2^{b})}} \wedge \bm{w}_J^{ab}, & \mbox{if both ARTs are active (training)}\\
\bm{w}_J^{ab}, & \mbox{if only ART\textsubscript{a} is active (prediction)} \\
\bm{y}^{{(F_2^{b})}}, & \mbox{if only ART\textsubscript{b} is active} \\
\Vec{\bm{0}}, & \mbox{otherwise} \\
\end{cases}.
\label{Eq:FAM_1}
\end{equation} 

During training, \ARTa\ and \ARTb\ perform their dynamics (Section~\ref{Sec:FA}) simultaneously and independently, with their respective inputs, until both establish resonant nodes $J$ and $K$, respectively. Then, the map field computes its activity vector using these two pieces of information, as defined in Eq.~(\ref{Eq:FAM_1}). Next, a second (map field) vigilance test is performed to assess the mapping correctness using
\begin{equation}
M_J^{ab} = \frac{\norm{\bm{y}^{{(F^{ab})}}}_1}{\norm{\bm{y}^{{(F_2^{b})}}}_1} = \frac{\norm{\bm{y}^{{(F_2^{b})}} \wedge \bm{w}_J^{ab}}_1}{\norm{\bm{y}^{{(F_2^{b})}}}_1},
\label{Eq:FAM_2}
\end{equation}
\noindent and, if it satisfies $M_J^{ab} \geq \rho_{ab}$ ($\rho_{ab} \in [0,1]$), then learning takes place. Otherwise, in response to a mismatch, the match tracking mechanism (M+) is triggered: the current resonating category~$J$ is inhibited (lateral reset), \ARTa's vigilance parameter is raised by a small constant (Eq.~(\ref{Eq:AM_3})) and the search continues with the remaining nodes until a resonant category that satisfies both $\rho_a$ and $\rho_{ab}$ is either found or created. Finally, $\rho_a$ is reset to its baseline value $\rho_a = \bar{\rho}_a$ for the presentation of the following sample. However, the study in~\cite{Anagnostopoulos2003a} indicates that not using match tracking (MT+) reduces the computational burden and model complexity while improving generalization capabilities~\cite{Andonie2006}.

In both fuzzy ART modules learning is ensued as described in Section~\ref{Sec:FA}, whereas in the map field, its weight vectors are updated such that a permanent association is made between the active nodes of \ARTa\ and \ARTb
\begin{equation}
w_{Jk}^{ab}(new) =  \bm{y}^{(F_2^b)} \wedge \bm{w}_J^{ab}(old) = 
\begin{cases} 
1, & \mbox{if } k=K \\ 
0, & \mbox{otherwise}
\end{cases}. 
\label{Eq:FAM_3}
\end{equation}

Note that uncommitted nodes participate in the WTA competition. They are initialized as $\Vec{\bm{1}}$, and the \ARTa's ones are mapped to all \ARTb \ nodes. A slow-learning mode was introduced in~\cite{Carpenter1995c}:
\begin{equation}
\bm{w}_J^{ab}(new) = \left( 1 - \beta_{ab} \right)\bm{w}_J^{ab}(old) + \beta_{ab} \left[ \bm{y}^{{(F_2^{b})}} \wedge \bm{w}_J^{ab}(old) \right],
\label{Eq:FAM_4}
\end{equation}
\noindent where $\beta_{ab}$ is the map field's learning rate, and the conditional probability $p(c_K^b | c_J^a)$ can be estimated nonparametrically as
\begin{equation}
\hat{p}(c_K^b | c_J^a) = \dfrac{w_{JK}^{ab}}{\sum\limits_{i=1}^{N_b}w_{Ji}^{ab}}.
\label{Eq:FAM_5}
\end{equation}

\textbf{Inference.} 
In testing mode only \ARTa\ is active. Its output is used to make a prediction and concretely retrieve the labels from \ARTb\ via the \Fab's weight matrix (Eqs.~(\ref{Eq:AM_5}) and~(\ref{Eq:AM_5a})). Note that training, prediction/inference and learning are all WTA (based on a single category).

The simplified fuzzy ARTMAP (SFAM)~\cite{kasuba1993} is a simplification of the original fuzzy ARTMAP specifically devised for classification tasks, in which, like simple ARTMAP in Sec.~\ref{Sec:AM}, \ARTb\ is replaced by vectors indicating the class labels. Another simplified design is discussed in~\cite{Baghmisheh2003}. 

\subsubsection{Fuzzy Min-Max}
\label{Sec:fuzzy_minmax_SL}

Fuzzy Min-Max~\cite{Simpson1992} is a supervised learning neural network classifier that uses fuzzy sets for its internal categories, like its clustering counterpart (Sec.~\ref{sec:fuzzy-min-max-art}). It is composed of three layers of neurons: an input layer F\textsubscript{A}, a layer of hyperbox nodes F\textsubscript{B} and a layer of class nodes F\textsubscript{C}. The hyperbox fuzzy sets are adjusted using an expansion-and-contraction-based fuzzy min-max classification learning algorithm that adjusts the fuzzy associations between the inputs and classes. It accomplishes that by identifying which hyperbox to expand for each input and expanding it correspondingly. Then, it identifies any resulting overlap between hyperboxes of different classes and minimally adjusts these hyperboxes to eliminate the overlap.

\subsubsection{Fusion ARTMAP}
\label{Sec:FusionAM}

Fusion ARTMAP~\cite{Asfour.1993} is a modular neural network model designed to classify data originating from multiple sources (i.e., to perform sensor fusion). It generalizes fuzzy ARTMAP (Sec.~\ref{Sec:FAM}) by incorporating multiple ART modules, one for each sensor. The outputs of these local ART modules are fed to a fuzzy ARTMAP, specifically, to the latter's \ARTa\ module, since \ARTb\ receives the class labels. Another key feature of fusion ARTMAP is the parallel match tracking. Following an incorrect prediction, the vigilance parameter of each ART module is raised (individual ARTs and fuzzy ARTMAP's \ARTa) 
\begin{equation}
\rho_k = \bar{\rho}_k + \Delta \rho,~\forall k,
\label{Eq:fusion_AM_1}
\end{equation}
\begin{equation}
\Delta \rho = \left(M_J^n - \bar{\rho}_n \right) + \epsilon,    
\label{Eq:fusion_AM_2}
\end{equation}
\begin{equation}
n = \argmin\limits_k \left(M_J^k\right),    
\label{Eq:fusion_AM_3}
\end{equation}
\noindent where $\rho_k$ and $\bar{\rho}_k$ are the vigilance and baseline vigilance of ART module $k$, respectively. Each ART module can have its own baseline vigilance parameter, or the entire fusion ARTMAP system can have a single common baseline vigilance. The variable $M_J^k$ is the match function value of ART module $k$'s category $J$. Note that ART module $n$ yielded the smallest match value and is therefore deemed the least predictive.

The vigilance values of the local ART modules and fuzzy ARTMAP's \ARTa\ are increased by the same value, which is enough to promote a mismatch in ART module $n$. Therefore, the latter is forced to promote a new search, while the other modules maintain their output. This procedure enables credit assignment to specific modules instead of uniformly blaming all modules regardless of their predictive power. Fusion ARTMAP improves memory compression (compared to single-ART module systems that concatenate all sensor data into a single large vector) given the sharing of the local ART's weight vectors across fuzzy ARTMAP.

The generalized symmetric fusion ARTMAP~\cite{Asfour.1993} replaces fuzzy ARTMAP with a global ART module that receives the outputs of all local ART modules and is responsible for the decision-making process. This model can handle multiple input sensors and multiple supervised inputs. In cases consisting of only one supervised input, the functionality is reduced to fusion ARTMAP. 

\subsubsection{LAPART}
\label{Sec:LAPART}

The LAPART~1~\cite{Healy1993} and LAPART~2~\cite{Healy1998} neural networks are two ART-based logic inference and supervised learning architectures. The LAPART 1 architecture uses two ART 1 networks $A$ and $B$ to learn logic inference and association, wherein if network $A$ assigns its input sample to a category, that results in network $B$ assigning its input to the corresponding category. It then uses the learned inference associations between the two networks to test hypotheses and classification decisions. The LAPART~2 algorithm uses the same architecture but introduces a lateral reset procedure and builds a rule extraction network that was shown to converge in two passes through the training data.

\subsubsection{ART-EMAP}
\label{Sec:ARTEMAP}

Adaptive resonance theory with spatial and temporal evidence integration (ART-EMAP)~\cite{Carpenter1995} augments fuzzy ARTMAP with a number of features to deal with noisy or ambiguous data: distributed representation during inference, integration of spatial-time information, extension of the map field into a multiple field EMAP module and a fine-tuning unsupervised learning stage (rehearsal). 

\textbf{Training.} ART-EMAP training is identical to fuzzy ARTMAP's (Sec.~\ref{Sec:FAM}).

\textbf{Inference.} ART-EMAP introduces two contrast enhancement procedures for distributed activation: the normalized power rule defined as
\begin{equation}
y_j^{(F_2^a)} = \frac{(T_j^a)^p}{\sum\limits_{i=1}^{N_a} (T_i^a)^p}~,~p>1, \label{Eq:EMAP_1}  
\end{equation}
\noindent and the threshold rule
\begin{equation}
y_j^{(F_2^a)} = \frac{[T_j^a - T]^+}{\sum\limits_{i=1}^{N_a} [T_i^a - T]^+} 
\end{equation}
\noindent where $T$ is a threshold parameter, and $[\xi]^+ = \max\{0, \xi\}$ is a rectifier operation. The activity of the first map field $F_1^{ab}$ is then defined as
\begin{equation}
\bm{y}^{(F_1^{ab})} = \bm{S}^{ab}
\label{Eq:EMAP_3}  
\end{equation}
\noindent where 
\begin{equation}
S^{ab}_k = \sum\limits_{j=1}^{N_a} w_{jk}^{ab}y_j^{(F_2^a)},
\label{Eq:EMAP_4}  
\end{equation}

A class is predicted using such distributed representation via the second map field activity $F_2^{ab}$
\begin{equation}
y_k^{(F_2^{ab})} =  
\begin{cases} 
1, & \mbox{if } k=K \\ 
0, & \mbox{otherwise}
\end{cases}, 
\label{Eq:EMAP_5}  
\end{equation}
\noindent where 
\begin{equation}
K = \argmax\limits_k \left[ y_k^{(F_1^{ab})} \right].    
\label{Eq:EMAP_6}  
\end{equation}

To address ambiguity (i.e., categories with similar activation values), the $F_2^{ab}$ activity can be redefined as:
\begin{equation}
y_k^{(F_2^{ab})} =  
\begin{cases} 
1, & \mbox{if } y_k^{(F_1^{ab})} > (DC)y_j^{(F_1^{ab})}~~\forall j \neq k \\ 
0, & \mbox{otherwise}
\end{cases} 
\label{Eq:EMAP_7}  
\end{equation}
\noindent where $DC \geq 1$ is a decision criterion. While $\bm{y}_k^{(F_2^{ab})} = \Vec{\bm{0}}$, the system waits for another input (i.e., data samples from the same and yet unknown class) until the inequality in  Eq.~(\ref{Eq:EMAP_7}) is satisfied. Moreover, the power rule can also be applied to the $F_1^{ab}$ activity 
\begin{equation}
y_k^{(F_1^{ab})} = \frac{(S_k^{ab})^q}{\sum\limits_{i=1}^{N_b} (S_i^{ab})^q}~,~q>1, 
\label{Eq:EMAP_8}  
\end{equation}
\noindent where the $q$ is the power parameter.

To handle noisy environments, ART-EMAP uses a map evidence accumulation field $F_E^{ab}$ that combines information from multiple $F_1^{ab}$ activities over time:
\begin{equation}
T_k^{ab}(new) = T_k^{ab}(old) + y_k^{(F_1^{ab})},
\label{Eq:EMAP_9}  
\end{equation}
\noindent where $T_k^{ab}$ is the evidence accumulating MTM. It is initialized as zero ($\bm{T}^{ab} = \Vec{\bm{0}}$) and reset once the DC is satisfied. The $F_2^{ab}$ activity can then be redefined as
\begin{equation}
y_k^{(F_2^{ab})} =  
\begin{cases} 
1, & \mbox{if } T_k > (DC)T_j~~\forall j \neq k \\ 
0, & \mbox{otherwise}
\end{cases}, 
\label{Eq:EMAP_2}  
\end{equation}
\noindent where improved accuracy correlates with larger $DC$ values and a greater number of samples~\cite{Carpenter1995}. 

Finally, to learn from the samples used to disambiguate prediction, an unsupervised learning stage (``rehearsal'') takes place. In this fine-tuning stage, the LTMs of \ARTa, \ARTb\ and the map field maintain their values, whereas another set of weights from \Fta\ to $\text{F}_\text{E}^\text{ab}$ is adapted when such samples are re-presented to the system. 

\subsubsection{Adaptive resonance associative map}
\label{Sec:ARAM}

The fuzzy adaptive resonance associative map (ARAM)~\cite{Tan1995b} extends ART autoassociative to heteroassociative mappings by connecting two ARTs (A and B) via a common category representation field \Fout.

\textbf{LTM.} Fuzzy ARAM has two \Fin\ layers connected to a single \Fout\ layer whose LTM unit is \mbox{$\bm{\theta} = \{\bm{w} = [\bm{w}^a, \bm{w}^b]\}$}.

\textbf{Activation.} 
When normalized and complement coded inputs (\mbox{$\bm{x} = [\bm{x}^a, \bm{x}^b]$}) are presented, the activation function is computed as
\begin{equation}
T_j = \gamma \dfrac{|\bm{x}^a \wedge \bm{w}_j^a|}{\alpha_a +|\bm{w}_j^a|} + (1 - \gamma) \dfrac{|\bm{x}^b \wedge \bm{w}_j^b|}{\alpha_b +|\bm{w}_j^b|},    \end{equation}
\noindent where $\gamma \in [0,1]$ is the contribution parameter. Note that there is an independent set of parameters for each module: choice parameters $\alpha_m>0$, learning parameters $\beta_m \in [0,1]$ and vigilance parameters $\rho_m \in [0,1]$, where $m \in \{ a, b \}$. 

\textbf{Match and resonance.}
Consider that node $J$ has been selected via a WTA competition. \Fin\ and \Fout\ activities are defined as:
\begin{equation}
y_j^{(F_1^m)} = 
\begin{cases} 
\bm{x}^m, & \mbox{if $F_2^m$ is inactive} \\ 
\bm{x}^m \wedge \bm{w}_J^m, & \mbox{otherwise}
\end{cases},
\end{equation}
\noindent where $m \in \{ a, b \}$, and
\begin{equation}
y_j^{(F_2)} = 
\begin{cases} 
1, & \mbox{if } j=J\\ 
0, & \mbox{otherwise}
\end{cases}.
\end{equation}

The match functions are computed for node $J$ as
\begin{equation}
M_J^m = \frac{\norm{\bm{y}^{(F_1^m)}}_1}{\norm{\bm{x}^m}_1} = \frac{\norm{\bm{x}^m \wedge \bm{w}_J^m}_1}{\norm{\bm{x}^m}_1},
\end{equation}
\noindent and resonance occurs if $M_J^m \geq \rho_m$ for both $m \in \{ a, b \}$ simultaneously.  Thus, $VR_J = \{[\bm{x}^a,\bm{x}^b] :$ \mbox{$M_J^a(\bm{x}^a) \geq \rho_a$}  and  \mbox{$M_J^b(\bm{x}^b) \geq \rho_b  \}$}. In this case, learning is ensued such that the weights $\bm{w}_J^m$ are updated using
fuzzy ART's learning rule (Eq.~(\ref{Eq:FA_6}) in Sec.~\ref{Sec:FA}). Otherwise, a match tracking mechanism temporarily raises the baseline $\bar{\rho}_a$ (which is reset at the start of each sample presentation) as in fuzzy ARTMAP (Sec.~\ref{Sec:FAM}), and the search for another resonant category continues. If an uncommitted category is recruited, then another one is initialized as $\bm{w}^m = \Vec{\bm{1}}$. Specifically, when such dynamics take place and $\gamma=1$, fuzzy ARAM is functionally equivalent to fuzzy ARTMAP~\cite{Tan1995b}.

\subsubsection{Gaussian ARTMAP}
\label{Sec:GAM}

The Gaussian ARTMAP (GAM)~\cite{williamson1996} is a discriminative model~\cite{vigdor2007} that uses Gaussian ART elementary units (Sec.~\ref{Sec:GA}) as building blocks. 

\textbf{Training.} Training follows the standard ARTMAP dynamics (Sec.~\ref{Sec:AM}), where the match tracking mechanism is triggered following a predictive error. 

\textbf{Inference.} During testing mode, predictions are made considering the total probability of each classes, i.e., by using Eqs.~(\ref{Eq:AM_5}) and~(\ref{Eq:AM_5a}) with $y_j^{(F_2^a)}=T_j^a$ (Eq.~(\ref{Eq:GA_1})).

\subsubsection{Probabilistic fuzzy ARTMAP}
\label{Sec:PFAM}

The probabilistic fuzzy ARTMAP (PFAM)~\cite{Lim1997, Lim2000a} combines fuzzy ARTMAP's code compression ability (Sec.~\ref{Sec:FAM}) with the probability density function estimation of probabilistic neural networks (PNN)~\cite{Specht1990} in a hybrid system: during training, a fuzzy ARTMAP variant is used to generate prototypes in a supervised manner, whereas during inference, the PNN uses Bayes decision theory to make predictions.

\textbf{Training.}  
Training is similar to fuzzy ARTMAP, except for the following:
\begin{enumerate}
\item Map field dynamics: the activity of F\textsuperscript{ab} used to compute the match function (Eq.~(\ref{Eq:FAM_2}) in Sec.~\ref{Sec:FAM}) is defined as
\begin{equation}
\bm{y}^{{(F^{ab})}} = \bm{y}^{{(F_2^{b})}} \wedge \frac{\bm{w}_J^{ab}}{\norm{\bm{w}_J^{ab}}_1},
\label{Eq:PFAM_1}
\end{equation} 
\noindent and when learning is ensued, $\bm{W}^{ab}$ is updated using
\begin{equation}
\bm{w}_J^{ab}(new) = \bm{w}_J^{ab}(old) + \bm{y}^{{(F^{ab})}};
\label{Eq:PFAM_2}
\end{equation}
\item If the match tracking mechanism is engaged, then the condition 
\begin{equation}
0 \leq \rho_a \leq \min\left( 1, M_J^a + \epsilon \right),~0<\epsilon \ll 1,
\label{Eq:PFAM_mt}
\end{equation} 
\noindent is enforced to enable identical categories to be associated with different classes~\cite{Lim1997b};
\item Centroids $\bm{\mu}_j^{a}$ are embedded in \ARTa\ (i.e., the LTM unit is $\bm{\theta}=\{\bm{w}, \bm{\mu}\}$). These are initialized as \mbox{$\bm{\mu}_j^{a} = \Vec{\bm{0}}$} and recursively estimated using 
\begin{equation}
\bm{\mu}_j^{a}(new) = \bm{\mu}_j^{a}(old) + \frac{1}{\norm{\bm{w}_J^{ab}}_1}\left( \bm{x}^a - \bm{\mu}_j^{a}(old) \right),
\label{Eq:PFAM_3}
\end{equation}
\noindent where $\bm{x}^a$ is complement coded for fuzzy ARTMAP categories $\bm{w}$ but not for the centroids $\bm{\mu}$.
\end{enumerate}

\textbf{Inference.} Prediction is accomplished using the maximum a posteriori (MAP) or minimum-risk estimate:
\begin{equation}
\hat{p}(c_k^b | \bm{x}^a) = \hat{p}(\bm{x}^a | c_k^b) \hat{p}(c_k^b)l(c_{jk}),
\label{Eq:PFAM_4}
\end{equation}
\noindent where $l(c_{jk})$ represents the cost of selecting $c_k^b$ when in fact the true class is $c_j^b$. The prior probability estimate of a given class $k$ is given by the ratio of the number of samples encoded by \ARTa's prototypes that are mapped to class $k$ to the total number of samples presented to PFAM:
\begin{equation}
\hat{p}(c_k^b) = \dfrac{\sum\limits_{j=1}^{N_a} w_{jk}^{ab}}{\sum\limits_{k=1}^{N_b} \sum\limits_{j=1}^{N_a} w_{jk}^{ab}},
\label{Eq:PFAM_5}
\end{equation}
\noindent and $p(\bm{x}^a|c_k^b)$ is estimated using the Parzen-window method~\cite{Parzen1962, Cacoullos1966} with isotopic Gaussians kernels ($\bm{\Sigma}_j=\sigma_j^2\bm{I}$)
\begin{equation}
\hat{p}(\bm{x}^a|c_k^b) = \sum\limits_{j=1}^{N_a} \dfrac{\mathds{1}_{c_k^b}(\bm{\mu}_j^{a})}{\sum\limits_{i=1}^{N_a} \mathds{1}_{c_k^b}(\bm{\mu}_i^{a})}  \frac{e^{\left(-\frac{\norm{ \bm{x}^a - \bm{\mu}_j^{a}}_2^2}{2 \sigma_j^2} \right)}}{\left( 2 \pi \right)^{\frac{d}{2}} \sigma_j^d},
\label{Eq:PFAM_8}
\end{equation}
\noindent where 
\begin{equation}
\mathds{1}_{c_k^b}(\bm{\mu}_j^{a}) =
\begin{cases} 
1, & \mbox{if } \bm{\mu}_j^{a} \in c_k^b \\ 
0, & \mbox{otherwise}
\end{cases}.
\label{Eq:PFAM_8a}
\end{equation}

The kernels used for the realization of the Parzen-window density estimation have heteroscedastic components, which are computed as
\begin{equation}
\sigma_j = \frac{1}{r} \min\limits_i \norm{ \bm{\mu}_j^{a} - \bm{\mu}_i^{a}}_2,
\label{Eq:PFAM_9}
\end{equation}
\noindent or determined using the $k$-nearest neighbors method~\cite{duda2000}
\begin{equation}
\sigma_j = \frac{1}{k} \sum\limits_{i=1}^k \| \bm{\mu}_j^{a} - \bm{\mu}_i^{a}\|,~ 1 \leq k \leq N_a - 1,
\label{Eq:PFAM_10}
\end{equation}
\noindent where $r$ is a user-defined overlapping parameter, and $\bm{\mu}_j^{a}$ and $\bm{\mu}_i^{a}$ belong to different classes in Eqs.~(\ref{Eq:PFAM_9}) and~(\ref{Eq:PFAM_10}).

\subsubsection{ARTMAP-IC}
\label{Sec:AMIC}

The ARTMAP-IC model~\cite{Carpenter1998b} is a fuzzy ARTMAP variant whose key characteristics are (i) a new match tracking mechanism (MT-) to reduce model complexity and (ii) the inclusion of instance counting (via a new counting field F\textsubscript{3}) for probabilistic distributed prediction.

ARTMAP-IC replaces \ARTb\ with a vector $\bm{y}^b$ encoding the classes of the classification problem, such that, for a given input $\bm{x}^a$ presented to \ARTa,
\begin{equation}
y_i^b = 
\begin{cases}
1, & \mbox{if } \bm{x}^a \in \mbox{ class } i\\
0, & \mbox{otherwise}
\end{cases},
\label{Eq:AMIC_1}
\end{equation}

The activity of the counting field F\textsubscript{3} (located in-between ART\textsubscript{a} and F\textsuperscript{ab}) is defined as
\begin{equation}
y_j^{(F_3)} = 
\begin{cases}
y_j^{(F_2^a)}, & \mbox{training}    \\
\dfrac{c_j y_j^{(F_2^a)}}{\sum\limits_{i=1}^{N_a} c_i y_i^{(F_2^a)}}, & \mbox{prediction}     
\end{cases},
\label{Eq:AMIC_2}
\end{equation}
\noindent where the instance counting weight $c_j$ records the number of samples that are encoded by  category $j$, i.e., the number of times it is activated. The map field F\textsuperscript{ab} activity can then be defined as
\begin{equation}
\bm{y}^{{(F^{ab})}} = 
\begin{cases} 
\bm{y}^b \wedge \bm{U}, & \mbox{ training}\\
\bm{U}, & \mbox{ prediction} \\
\end{cases}
\label{Eq:AMIC_3}
\end{equation} 
\noindent where the kth component of the map field's input is
\begin{equation}
U_k = \sum\limits_{j=1}^{Na} w_{jk}^{ab} y_j^{(F_3)}, k=1,...,N_b,
\label{Eq:AMIC_4}
\end{equation}
\noindent and here $N_b$ represents the number of classes. 

\textbf{Training.} During training, the match function is defined as
\begin{equation}
M_J^{ab} = \frac{\norm{\bm{y}^b \wedge \bm{U}}_1 }{\norm{\bm{y}^b}_1} = \norm{ \bm{y}^b \wedge \bm{w}_J^{ab} }_1,
\label{Eq:AMIC_5}
\end{equation}
\noindent since $\bm{U} = \bm{w}_J^{ab}$ (because $\bm{y}^{(F_2^a)} = \bm{y}^{(F_3)}$) and $\norm{\bm{y}^b}_1=1$. If the vigilance criterion is not satisfied ($M_J^{ab} < \rho_{ab}$), then the new match tracking mechanism (MT-) is engaged such that ART\textsubscript{a}'s vigilance is set to
\begin{equation}
\rho_a(new) = M_J^a + \epsilon,~\epsilon \leq 0 \mbox{ and } \|\epsilon\| \mbox{ small}, 
\label{Eq:AMIC_6}
\end{equation}
\noindent and the search proceeds as with fuzzy ARTMAP. Otherwise, if learning is ensued, then fuzzy \ARTa\ and the map field weight vectors learn as described in Secs.~\ref{Sec:FA} and~\ref{Sec:FAM}, respectively. The instance counting is updated as
\begin{equation}
c_j(new) = c_j(old) + y_j^{(F_2^a)},  
\label{Eq:AMIC_7}
\end{equation}
\noindent where $c_j$'s are initialized as 0.

\textbf{Inference.} During testing, no search occurs, and ARTMAP-IC uses the Q-max rule to distribute \Fta\ activity via the following contrast enhancement procedure:
\begin{equation}
y_j^{(F_2^a)} = 
\begin{cases} 
\dfrac{T_j}{\sum\limits_{\lambda \in \Lambda} T_{\lambda}}, & \mbox{if } j \in \Lambda \\ 
0, & \mbox{otherwise}
\end{cases},
\label{Eq:AMIC_9}
\end{equation}
\noindent where $\Lambda$ is the set formed by the $Q$ categories with the largest activation values ($Q$ is a user-defined parameter). This is similar to k-nearest neighbors~\cite{duda2000} where $Q$ assumes the role of $k$~\cite{Carpenter1998b}. Setting $Q=1$ leads to WTA mode.

Finally, the probability of class $k$ is then computed as
\begin{equation}
\sigma_k = \dfrac{U_k}{\sum\limits_{l=1}^{N_b} U_l} = \dfrac{\sum\limits_{j \in \Lambda} w_{jk}^{ab} c_j T_j}{\sum\limits_{l=1}^{N_b} \sum\limits_{j \in \Lambda} w_{jl}^{ab} c_j T_j}.    
\label{Eq:AMIC_10}
\end{equation}

\subsubsection{Distributed ARTMAP}
\label{Sec:dARTMAP}

Distributed ARTMAP (dARTMAP)~\cite{Carpenter1998a} was developed to improve supervised ART models regarding model compactness and noise robustness (i.e., reduce category proliferation) while performing fast and stable learning via distributed representation. It features distributed activation, match and learning functions. Notably, distributed ARTMAP generalizes the following supervised ART models~\cite{Carpenter2003}: ``dARTMAP $\supset$ ARTMAP-IC $\supset$ default ARTMAP $\supset$ fuzzy ARTMAP'', where $\supset$ is used to indicate containment considering this ARTMAP's ecosystem.

In case of classification problems, distributed ARTMAP uses distributed ART (Sec.~\ref{Sec:dART}) as a building block for \ARTa, while replacing \ARTb\ with a binary vector indicating the input's class membership (Eq.~(\ref{Eq:AMIC_1}) in Sec.~\ref{Sec:AMIC}). The distributed ARTMAP uses an increased-gradient content-addressable memory (IG CAM) rule for contrast enhancement. A CAM rule defines a function that yields the steady state values of the network's STM when an input sample is presented. Particularly, distributed ARTMAP's CAM rule defines a power function that is controlled by a parameter $p$. The latter has a role akin to the variance in Gaussian kernels, and, as it tends to infinity, the network converges to WTA.

\textbf{Training.} During training, the distributed ARTMAP alternates between distributed and WTA modes. Like ARTMAP-IC (Sec.~\ref{Sec:AMIC}), distributed ARTMAP features a counting field $\text{F}_\text{3}^\text{a}$ (for instance counting purposes) which is cascaded to \Fta\ and employs the MT- match tracking search algorithm. Briefly, the distributed representation undergoes the unsupervised (Eqs.~(\ref{Eq:dART_6}) to~(\ref{Eq:dART_5})) and supervised vigilance (i.e., prediction assessment) tests, and if one of them fails the system switches to WTA mode and its corresponding dynamics are carried out (in which nodes can be added incrementally). Otherwise, distributed mode dynamics take place. 

Particularly, the distributed ARTMAP uses the distributed choice-by-difference activation function (Eq.~(\ref{Eq:dART_2}) in Sec.~\ref{Sec:dART} disregarding the depletion parameters)
\begin{equation}
T_j = \sum\limits_{i=1}^{2d} \left [x_i^a \wedge (1 - \tau_i^{bu}) \right] + (1-\alpha)\sum\limits_{i=1}^{2d} \tau_i^{bu},~\alpha \in (0,1),   
\label{Eq:dARTMAP_1}
\end{equation}
\noindent and, after these are computed, the following subsets of highly active nodes are considered:
\begin{enumerate}
\item  $\Lambda = \{ j : T_j \geq T^u \}$ 
\item $\Lambda' = \{ j : T_j = (2 - \alpha)d \}$ 
\end{enumerate}
\noindent where $T^u$ is the activation function of an uncommitted node ($\bm{\tau}^{bu} = \bm{\tau}^{td} = \Vec{\bm{0}}$). The IG CAM rule specifies the following functions for the steady-state activities of distributed ARTMAP's modes
\begin{itemize}
\item Distributed mode
\begin{itemize}
\item If $\Lambda' \neq \{ \emptyset \}$, then
\begin{equation}
y_j^{(F_2^a)} = 
\begin{cases} 
\dfrac{1}{|\Lambda'|}, & \forall j \in \Lambda' \\ 
0, & \mbox{otherwise} 
\end{cases},
\label{Eq:dARTMAP_2}
\end{equation}
\noindent where $| \cdot |$ represents the cardinality of a set.

\item If $\Lambda' = \{ \emptyset \}$ and $\Lambda \neq \{ \emptyset \}$, then
\begin{equation}
y_j^{(F_2^a)} = 
\begin{cases} 
\dfrac{1}{1 + \sum\limits_{\lambda \in \Lambda, \lambda \neq j} \left[ \dfrac{(2 - \alpha)d - T_j}{(2 - \alpha)d - T_\lambda}\right]^p}, & \forall j \in \Lambda \\
0, & \mbox{otherwise} 
\end{cases}
\label{Eq:dARTMAP_3}
\end{equation}
\end{itemize}
\noindent where $p \in (0, \infty)$ is the power parameter. The \ARTa's counting field F\textsubscript{3} activity is then defined as
\begin{equation}
y_j^{(F_3^a)} = \dfrac{c_j y_j^{(F_2^a)}}{\sum\limits_{\lambda =1}^C c_\lambda y_\lambda^{(F_2^a)}},
\label{Eq:dARTMAP_4}
\end{equation}
\noindent where $C$ is the number of \ARTa's committed nodes, and $c_j$ is the instance counting of node $j$ (if uncommitted, then $c_j=0$). The signal used in the \ARTa's match function is then
\begin{equation}
\sigma_i = \sum\limits_{j=1}^C \left[ y_{j}^{(F_3^a)} - \tau^{td}_{j,i}\right]^+,~i=1,...,2d.   
\label{Eq:dARTMAP_5}
\end{equation}

\item WTA mode

\begin{itemize}
\item If $\Lambda \neq \{ \emptyset \}$, then the winner node is \mbox{$J = \argmax\limits_{j \in \Lambda}{(T_j)}$}.
\item If $\Lambda = \{ \emptyset \}$, then the uncommitted node is recruited to learn the presented input sample.
\end{itemize}
The \ARTa's counting field F\textsubscript{3} activity is then
\begin{equation}
y_j^{(F_3^a)} = y_j^{(F_2^a)} =
\begin{cases} 
1, & \mbox{if } j=J \\
0, & \mbox{otherwise} 
\end{cases},
\label{Eq:dARTMAP_6}
\end{equation}
\noindent and the signal used in the \ARTa's match function is
\begin{equation}
\sigma_i = \left( 1 - \tau^{td}_{J,i}\right),~i=1,...,2d.   
\label{Eq:dARTMAP_7}
\end{equation}
\end{itemize}

If the vigilance test of \ARTa\ is not satisfied (Eqs.~(\ref{Eq:dART_6}) to~(\ref{Eq:dART_5})) in Sec.~\ref{Sec:dART}), then distributed ARTMAP reverts to WTA mode, and the search continues until a resonant node is either found or created. Finally, the output class is then estimated using Eqs.~(\ref{Eq:AM_5}) and~(\ref{Eq:AM_5a}) with $y_j^{(F_3^a)}$ in place of $y_j^{(F_2)}$. If the prediction is incorrect, then match tracking is engaged using the MT- algorithm (Sec.~\ref{Sec:AMIC}). Otherwise, \ARTa\ adapts using the distributed ART learning laws described in Sec.~\ref{Sec:dART} (the top-down thresholds components are updated using $y_j^{(F_3^a)}$ in place of $y_j^{(F_2)}$ in Eq.~(\ref{Eq:dART_9})), and the instance countings are updated using Eq.~(\ref{Eq:AMIC_7}) in Sec.~\ref{Sec:AMIC}.

Note that if the distributed ARTMAP system enters a resonant state while in distributed mode, then, prior to learning, a credit assignment stage takes place in which the nodes permanently associated with the wrong class are inhibited, the \Fta\ activity is re-normalized (i.e., \mbox{$\norm{\bm{y}^{(F_2^a)}}_1 = 1$}) and the $\text{F}_\text{3}^\text{a}$ activity and the signal $\bm{\sigma}$ are recomputed using Eqs.~(\ref{Eq:dARTMAP_4}) and~(\ref{Eq:dARTMAP_5}), respectively.

\textbf{Inference.} To make a prediction for a new sample $\bm{x}$, distributed ARTMAP operates similarly to the training phase but always in distributed mode and with search and learning disabled (i.e., in feedforward mode).

\subsubsection{Hypersphere ARTMAP}
\label{Sec:HAM}

Hypersphere ARTMAP (HAM)~\cite{anagnostopoulos2000} closely follows the operation of fuzzy ARTMAP (Sec.~\ref{Sec:FAM}) but instead uses hypersphere ART (Sec.~\ref{Sec:HA}) modules for \ARTa\ and \ARTb. \ARTb\ is responsible for clustering the classes ($\bm{x}^b$), \ARTa\ does the data samples ($\bm{x}^a$) and the inter-ART maps the \ARTa\ categories to the \ARTb\ categories regulated by the match tracking procedure.

\subsubsection{Ellipsoid ARTMAP}
\label{Sec:EAM}

Similar to hypersphere ARTMAP, ellipsoid ARTMAP (EAM)~\cite{anagnostopoulos2001,anagnostopoulos2001b} uses ellipsoid ART (Sec.~\ref{Sec:EA}) for both its \ARTa\ and \ARTb\ modules while closely following the fuzzy ARTMAP's operation (Sec.~\ref{Sec:FAM}).

\subsubsection{\textmu ARTMAP}
\label{Sec:microAM}

The $\mu$ARTMAP model~\cite{Sanchez2000, Sanchez2002} is a fuzzy ARTMAP variant developed to reduce the type of category proliferation due to overlapping classes, consequently improving generalization capability. This is accomplished by regulating the conditional entropy between the input (\ARTa) and output (\ARTb) spaces
\begin{equation}
H(ART_b | ART_a) =  \sum\limits_{j=1}^{N_a}  h_j,
\label{Eq:microAM_1}
\end{equation}
\noindent where $h_j$ is the contribution of \ARTa's node $j$ to the total entropy:
\begin{equation}
h_j = - \hat{p}(c_j^a) \sum\limits_{k=1}^{N_b} \hat{p}(c_k^b | c_j^a) \log_2 \hat{p}(c_k^b | c_j^a),       
\label{Eq:microAM_2}
\end{equation}
\noindent and the probabilities are estimated using the map field's LTM unit, whose dynamics are similar to PROBART's (Sec.~\ref{Sec:PROBART}). This process indirectly controls the training error, which is relaxed to address overfitting.

\textbf{Training.} Training is divided into two phases, and the first one is performed online. Assuming the resonant categories of \ARTa\ and \ARTb\ are $J$ and $K$, respectively, the map field vigilance test is defined using Eq.~(\ref{Eq:microAM_2}):
\begin{equation}
M_J^{ab} = h_J,   
\label{Eq:microAM_4}
\end{equation}
\noindent where
\begin{equation}
p(c_k^b | c_j^a) = 
\begin{cases} 
\dfrac{y^{(F^{ab})}_k}{ \norm{ \bm{y}^{(F^{ab})} }_1}, & \mbox{if } j=J \\ 
\dfrac{w^{ab}_{jk}}{ \norm{ \bm{w}^{ab}_j }_1 }, & \mbox{ otherwise} 
\end{cases},
\label{Eq:microAM_5}
\end{equation}
\begin{equation}
p(c_j^a) = 
\begin{cases} 
\dfrac{\norm{ \bm{y}^{(F^{ab})} }_1}{ \norm{ \bm{y}^{(F^{ab})} }_1 + \sum\limits_{i=1 ,i \neq J}^{N_a} \norm{ \bm{w}^{ab}_i }_1}, & \mbox{if } j=J \\ 
\dfrac{\norm{ \bm{w}^{ab}_j }_1 }{ \norm{ \bm{y}^{(F^{ab})} }_1 + \sum\limits_{i=1 ,i \neq J}^{N_a} \norm{ \bm{w}^{ab}_i }_1 }, & \mbox{ otherwise} 
\end{cases}.
\label{Eq:microAM_6}
\end{equation}
\noindent Note, however, that if $J$ is an uncommitted node, then
\begin{equation}
p(c_k^b | c_J^a) = 
\begin{cases} 
1, & \mbox{if } k=K \\ 
0, & \mbox{ otherwise} 
\end{cases},
\label{Eq:microAM_7}
\end{equation}
\noindent which implies $h_J=0$. The value of $h_J$ measures the homogeneity of \ARTb\ nodes (i.e., classes) associated with \ARTa's category $J$. If $M_J^{ab} \leq h_{max}$, where $h_{max}$ is a user-defined parameter, then the map field vigilance is satisfied and learning is ensued as in PROBART (Eq.~(\ref{Eq:PROBART_2})). Otherwise, \ARTa's node $J$ is inhibited, and the search continues without changing \ARTa's vigilance parameter. Note that $h_{max}=0$ implies mapping to a single class, whereas $h_{max}>0$ allows mapping to different classes (i.e., non-zero training error).

Next, an offline training  phase is performed to measure the overlap between categories. In this second training phase no learning is permitted within the ART modules. Probabilities are re-estimated using
\begin{equation}
p(c_k^b | c_j^a) = \dfrac{v^{ab}_{jk}}{ \norm{ \bm{v}^{ab}_j }_1 },
\label{Eq:microAM_9}
\end{equation}
\begin{equation}
p(c_j^a) = \dfrac{\norm{ \bm{v}^{ab}_j }_1 }{ \sum\limits_{i=1}^{N^a} \norm{ \bm{v}^{ab}_i}_1},
\label{Eq:microAM_10}
\end{equation}
\noindent where a temporary map field co-occurrence matrix $\bm{V}^{ab}$ is updated in a unsupervised manner, i.e., without match tracking (Initialization: $\bm{V}^{ab} = \bm{0}$). The total entropy $H$ is computed using Eq.~(\ref{Eq:microAM_1}), and if $H > H_{max}$, where $H_{max}$ is a user-defined parameter, then the mapping is considered too entropic. \ARTa's category $M$ with the largest contribution $h_M$ is removed, and the baseline vigilance $\bar{\rho}_a$ is increased for all new uncommitted categories as
\begin{equation}
\bar{\rho}_a = \dfrac{\norm{ \bm{w}_M^a }_1 }{ \norm{\bm{x}^a}_1 }  + \epsilon,
\label{Eq:microAM_11}
\end{equation}
\noindent thus adaptively tuning individual vigilance parameters of \ARTa's categories. The samples that were associated with node $M$ are re-presented and the learning process resumes. This entire process is repeated until $H \leq H_{max}$, in which the training stops. Notably, if $h_{max},H_{max} \geq \log_2N_b$ then $\mu$ARTMAP behaves similarly to PROBART, whereas if $h_{max}=0$ and $H_{max} \geq \log_2N_b$, then $\mu$ARTMAP behaves similarly to fuzzy ARTMAP. 

\textbf{Inference.}  Predictions are made using Eqs.~(\ref{Eq:AM_5}) and~(\ref{Eq:AM_5a}), i.e., the class output $K$ is estimated as the one that has the largest frequency of association with \ARTa's resonant category $J$.

Under certain conditions, $\mu$ARTMAP creates large categories that lead to considerable overlaps and decrease the system's performance. The safe-$\mu$ARTMAP~\cite{Sanchez2001} variant is a generalization of $\mu$ARTMAP that adds another vigilance criterion to mediate learning. Specifically, to avoid the formation of large hyperrectangles that enclose far apart samples belonging to the same class, besides passing both the \ARTa\ and the map field vigilance tests, an \ARTa\ category also needs to undergo a distance criterion defined as
\begin{equation}
M_J^{\Delta w} = \dfrac{\norm{ \bm{w}_J^a }_1 - \norm{\bm{w}_J^a \wedge \bm{x}^a}_1 }{\norm{\bm{x}^a}_1}.   
\label{Eq:microAM_13}
\end{equation}
\noindent Only if this third vigilance test is also satisfied ($M_J^{\Delta w} \leq \delta$, $0 < \delta < 1 - \rho$), then learning takes place. This imposes a restriction on the instantaneous change of a category size, which is upper bounded by~$\norm{\bm{x}^a}_1\delta$. Particularly, safe-$\mu$ARTMAP reduces to $\mu$ARTMAP when $\delta = 1$ (which effectively implies the absence of a constraint).

\subsubsection{Default ARTMAPs}
\label{Sec:defAM}

The default ARTMAP 1 model~\cite{Carpenter2003} is characterized by the usage of a distributed representation to perform continuously-valued predictions, as opposed to binary and fuzzy ARTMAP models (Secs.~\ref{Sec:AM} and~\ref{Sec:FAM}), which use WTA code representation.

\textbf{Training.} 
Default ARTMAP~1's training is akin to fuzzy ARTMAP's, except for (i)~the absence of \ARTb\ (default ARTMAP 1 is a simplified architecture), (ii)~its \ARTa\ module employs the choice-by-difference activation function defined as~\cite{Gjaja.1994a}
\begin{equation}
T_j = \norm{\bm{x} \wedge \bm{w}_j^a}_1 + (1 - \alpha)(d - \norm{\bm{w}_j^a}_1), \alpha \in (0,1),
\label{Eq:defAM_1}
\end{equation}
\noindent and (iii)~the match tracking algorithm, which is MT- search~\cite{Carpenter1998b}.

\textbf{Inference.} 
As opposed to fuzzy ARTMAP, default ARTMAP 1 uses a distributed representation for inference, where two subsets of highly active neurons are selected as:
\begin{enumerate}
\item  $\Lambda = \{ \lambda = 1,...,N_a : T_\lambda > \alpha d\}$ 
\item $\Lambda' = \{ \lambda = 1,...,N_a : T_\lambda = d~(i.e.,~\bm{w}_\lambda = \bm{x}^a)\}$ 
\end{enumerate}

Next, the IG CAM rule is applied:
\begin{itemize}
\item If $\Lambda' \neq \{ \emptyset \}$, then
\begin{equation}
y_j = 
\begin{cases} 
\dfrac{1}{|\Lambda'|}, & \forall j \in \Lambda' \\ 
0, & \mbox{otherwise} 
\end{cases},
\label{Eq:defAM_2}
\end{equation}
\noindent where $| \cdot |$ represents the cardinality of a set.
\item If $\Lambda' = \{ \emptyset \}$, then
\begin{equation}
y_j = 
\begin{cases} 
\frac{\left[ \dfrac{1}{d-T_j}\right]^p}{\sum\limits_{\lambda \in \Lambda} \left[ \dfrac{1}{d-T_\lambda}\right]^p}, & \forall j \in \Lambda \\
0, & \mbox{otherwise} 
\end{cases}.
\label{Eq:defAM_3}
\end{equation}
\end{itemize}

Finally, the predictions for each class are obtained using Eqs.~(\ref{Eq:AM_5}) and~(\ref{Eq:AM_5a}) in Sec.~\ref{Sec:AM}.

In a WTA system, such as fuzzy ARTMAP, after learning a sample, an immediate re-presentation is guaranteed to yield a correct prediction, i.e., it passes the ``next-input-test''. However, the default ARTMAP 1 WTA prediction during training might not be the same as the distributed one. To overcome this problem, the default ARTMAP 2 model~\cite{Amis2007} introduces the ``distributed-next-input-test'' during training, to assure that a correct prediction would also be performed under a distributed representation. Briefly, in order to anticipate an error, after learning from a sample in a WTA mode, the prediction is verified again using a distributed representation. If the distributed prediction is correct, then learning resumes by returning to WTA mode and presenting the next sample. Otherwise, the match tracking mechanism is engaged, the system reverts to WTA mode, the resonant category is inhibited and the network restarts the search to learn more from that sample.

\subsubsection{Boosted ARTMAP}
\label{Sec:boostedAM}

Boosted ARTMAP~\cite{Verzi1998} is a variant of fuzzy ARTMAP (Sec.~\ref{Sec:FAM}) closely related to PROBART~\cite{Marriott1995} (Sec.~\ref{Sec:PROBART}). It is inspired by Boosting theory~\cite{Schapire.1990a} and was developed to improve the fuzzy ARTMAP's generalization capability (since it is prone to overfitting the training data) and to create less complex networks (i.e., to reduce the type of category proliferation caused by overlapping classes). These are addressed by regulating the training error, which is allowed to be non-zero. Particularly, boosted ARTMAP's \ARTa\ and \ARTb\ modules are boosted ART models (which are identical to fuzzy ART, except that the categories are endowed with individual vigilance parameters), and its map field dynamics are equal to PROBARTs'.  

\textbf{Training.} Boosted ARTMAP learning is offline. After a first pass through the data, the error of \ARTa's category~$j$ is estimated as
\begin{equation}
\varepsilon_j =  p_j e_j = \dfrac{ \norm{\bm{w}^{ab}_j}_1 - \max\limits_k\left(w_{jk}^{ab}\right)}{\sum\limits_{m=1}^{N_a}\sum\limits_{n=1}^{N_b} w^{ab}_{mn}},
\label{Eq:boosted_2}
\end{equation}
\noindent where
\begin{equation}
p_j = p \left( \bm{x}~\mbox{selects}~c_j^a \right) = p(c_j^a) =  \dfrac{\norm{\bm{w}^{ab}_j }_1}{\sum\limits_{m=1}^{N_a}\sum\limits_{n=1}^{N_b} w^{ab}_{mn}},
\label{Eq:boosted_3}
\end{equation}
\begin{equation}
e_j = p \left( c^*~\mbox{not predicted by}~c_j^a \right) = 1-\frac{\max\limits_k\left( w_{jk}^{ab}\right)}{\norm{\bm{w}^{ab}_j}_1 },
\label{Eq:boosted_4}
\end{equation}
\noindent and the total error is given by
\begin{equation}
\varepsilon_T = \sum\limits_{j=1}^{N_a} \varepsilon_j = \frac{\sum\limits_{j=1}^{N_a} \left[ \norm{\bm{w}^{ab}_j}_1 - \max\limits_k\left( w_{jk}^{ab}\right)\right]}{\sum\limits_{m=1}^{N_a}\sum\limits_{n=1}^{N_b} w^{ab}_{mn}},
\label{Eq:boosted_5}
\end{equation}
\noindent where $c^*$ is the true class. Then, the vigilance parameters of \ARTa's nodes are raised by a user-defined parameter $\delta$:
\begin{equation}
\rho_{\lambda}(new) = \rho_{\lambda}(old) + \delta,~\lambda \in \Lambda, 
\label{Eq:boosted_1}
\end{equation}
\noindent where $\Lambda = \{\lambda : \varepsilon_{\lambda} > \varepsilon_{max} \}$, i.e., $\Lambda$ is the subset of nodes $\lambda$ with contributions $\varepsilon_{\lambda}$ to the total error $\varepsilon_T$ larger than the desired error $\varepsilon_{max}$. If $\Lambda = \{ \emptyset \}$ but the total error $\varepsilon_T$ is above the desired error $\varepsilon_{max}$ (i.e., if $\varepsilon_T > \varepsilon_{max}$), then the vigilances of all nodes $j$ with the largest contribution $\varepsilon_j$ are increased following Eq.~(\ref{Eq:boosted_1}). Note that when new nodes are added to the system, their initial vigilance parameter is set to a relaxed baseline value $\bar{\rho}$.

\textbf{Inference.} In prediction mode, when a sample is presented, the corresponding class label is obtained using the map field weight vector associated with \ARTa's resonant category $J$
\begin{equation}
K = \argmax\limits_k \left[ w_{Jk}^{ab} \right].
\label{Eq:boosted_6}
\end{equation}

As discussed in~\cite{Sanchez2002}, due to the lack of a match tracking mechanism, this version of boosted ARTMAP cannot handle ``populated exceptions'', i.e., when samples from one class surrounds another and it is necessary to create a category inside another category. The second version of boosted ARTMAP~\cite{Verzi2006} augments its predecessor with a match tracking mechanism to regulate the training error, whose map field dynamics are discussed next.

\textbf{Training.} During learning, when a sample pair is presented and \ARTa's and \ARTb's resonant nodes are $J$ and $K$, respectively, the map field match function is given by
\begin{equation}
M_J^{ab} = (1-e_J') \frac{\norm{ \bm{y}^{F_2^b} \wedge \bm{w}_J^{ab^{'}}}_1}{\norm{ \bm{y}^{F_2^b} }_1},     
\label{Eq:boosted_7}
\end{equation}
\noindent and resonance occurs if the winning category satisfies $M_J > (1-\epsilon)\rho_{ab}$, where $\epsilon \in [0,1]$ is the error tolerance parameter that binds the training error. The map field then learns as in PROBART (Eq.~(\ref{Eq:PROBART_2})). Otherwise, the match tracking mechanism is engaged. The temporary variables $e_J'$ and $\bm{w}_J^{ab^{'}}$ in Eq.~(\ref{Eq:boosted_7}) are computed as if category~$J$ were allowed to learn:
\begin{equation}
w_{Jl}^{ab^{'}} = 
\begin{cases} 
1, & \mbox{if } l=\argmax\limits_k \left( w_{Jk}^{ab} \right) \\
\ceil{0 + \epsilon}, & \mbox{otherwise}
\end{cases},
\label{Eq:boosted_8}
\end{equation}
\begin{equation}
e_J' = 1 - \frac{\max\limits_k \left( w_{Jk}^{ab^{''}} \right)}{\norm{\bm{w}^{ab^{''}}_J}_1 }, 
\label{Eq:boosted_9}
\end{equation}
\begin{equation}
w_{Jl}^{ab^{''}} = 
\begin{cases} 
w_{Jl}^{ab} + 1, & \mbox{if } l=K \\
w_{Jl}^{ab}, & \mbox{otherwise}
\end{cases},
\label{Eq:boosted_10}
\end{equation}
\noindent where $\ceil{\cdot}$ is the ceiling function. If node~$J$ is uncommitted, then $\bm{w}_J^{ab^{'}} = \Vec{\bm{1}}$ and $e_J'=0$ (no mismatch will take place).

\textbf{Inference.} Predictions are made using Eq.~(\ref{Eq:boosted_6}).

Note that boosted ART generalizes fuzzy ART, and boosted ARTMAP reduces in functionality to 
fuzzy ARTMAP by setting $\varepsilon_d = 0$ and $\rho^{ab}>0.5$ and to PROBART by setting $\varepsilon_d = 1$. Note that boosted ARTMAP performs empirical risk minimization, however, variants of boosted ARTMAP, such as~\cite{Verzi2001, Verzi2002, Verzi2006}, perform structural risk minimization and use Rademacher penalization~\cite{Koltchinskii2001}.

\subsubsection{Fuzzy ARTMAP with input relevances}
\label{Sec:FAMR}

The fuzzy ARTMAP with input relevances (FAMR) model~\cite{Andonie2003a, Andonie2003b, Andonie2006} is a fuzzy ARTMAP variant that modifies the map field dynamics, while maintaining the remaining dynamics of fuzzy ARTMAP. Thus, the incremental and non-parametric estimation of posterior probabilities based on the map field is augmented to reflect the degree of importance of incoming samples, especially when these are arriving from multiple heterogeneous sources corrupted by different noise levels. 

\textbf{Training.} Particularly, a sample arriving at time $t>0$ has a relevance factor $q_t \in (0,\infty)$. It is a user-defined or computed parameter, e.g., samples may be ranked based on their source noise level or have their relevance factors made proportional to its importance. Assuming the resonant categories of \ARTa\ and \ARTb\ are $J$ and $K$, respectively, then the map field recursive update equations are based on the stochastic approximation procedure~\cite{Andonie1990}: 
\begin{equation}
w_{jk}^{ab}(new) = 
\begin{cases} 
w_{jk}^{ab}(old), & j \neq J \\
\left( 1 - A_t \right)w_{jk}^{ab}(old) + A_t, & j = J,k = K \\
\left( 1 - A_t \right)w_{jk}^{ab}(old), & j = J,k \neq K
\end{cases},
\label{Eq:FAMR_1}
\end{equation}
\noindent where
\begin{equation}
A_t = \dfrac{q_t}{Q_J(new)},  
\label{Eq:FAMR_2}
\end{equation}
\begin{equation}
Q_J(new) = Q_J(old) + q_t,  
\label{Eq:FAMR_3}
\end{equation}
\noindent and $\bm{Q} = [Q_1 ... Q_{N_a}]$. Thus, an entry $w_{i,j}^{ab}$ of the map field matrix $\bm{W}^{ab}$ is an estimate of $p(c_k^b|c_k^a)$. If a new category~$K$ is created in \ARTb, then the map field weights $w_{jk}^{ab}$ are adapted as: 
\begin{equation}
w_{jk}^{ab}(new) = 
\begin{cases} 
\dfrac{q_0}{N_b(new) Q_j}, & \forall j , k=K \\
w_{jk}^{ab}(old) - \dfrac{w_{jK}^{ab}(new)}{N_b(new) - 1}, & \forall j , k \neq K 
\end{cases},
\label{Eq:FAMR_4}
\end{equation}
\noindent where $N_b(new) = N_b(old) + 1$, is the new number of nodes in ART\textsubscript{b}. If a new category is created in \ARTa\ \mbox{($J=N_a+1$)}, then $Q_J$ is set as $q_0 \geq 0$ (initial relevance parameter) and $w_{Jk}^{ab} = 1/N_b$, $\forall k$. Finally, the map field's vigilance test is redefined as
\begin{equation}
M_J^{ab} = N_b w_{JK}^{ab},    
\label{Eq:FAMR_5}
\end{equation}
\noindent such that $M_J^{ab} \geq \rho_{ab}$ must be satisfied for resonance to occur.

\textbf{Inference.} Predictions are made similarly to fuzzy ARTMAP (Sec.~\ref{Sec:FAM}).

\subsubsection{Bayesian ARTMAP}
\label{Sec:BAM}

Bayesian ARTMAP (BAM)~\cite{vigdor2007} is a generative model based on Bayes' decision theory~\cite{vigdor2007} that uses Bayesian ART modules (Sec.~\ref{Sec:BA}) as building blocks and represents class density by Gaussian mixtures. Moreover, the posterior probabilities in Bayes' theorem are estimated within and between ART modules.

\textbf{Training.} During training, the map field LTM unit is a matrix of association frequency (sample count) \mbox{$\bm{W}^{ab}=\bm{N} = [n_{kj}]_{N_b \times N_a}$} that is used to estimate the \ARTa\ and \ARTb\ joint probability distribution
\begin{equation}
\hat{p}(c_k^b, c_j^a) = \frac{n_{kj}}{\sum\limits_{i=1}^{N_b}\sum\limits_{l=1}^{N_a}n_{il}},    
\end{equation}
\noindent such that soft and hard mappings between ART modules are possible, i.e., a deterministic many-to-one mapping or a probabilistic many-to-many mapping based on $\hat{p}(c_k^b, c_j^a)$. The match tracking mechanism is triggered by the system if the match function value for \ARTa's resonant category~$J$
\begin{equation}
M^{ab}_J = \hat{p}(c_k^b | c_J^a) = \dfrac{n_{k,J}}{\sum\limits_{i=1}^{N_b}n_{i,J}},   
\end{equation}
\noindent does not satisfy $M^{ab}_J \geq \rho_{ab}$, where $\rho_{ab}$ represents the minimum class posterior probability threshold. Note that setting $\rho_{ab}=1$ enforces a hard many-to-one mapping, and Bayesian ARTMAP reduces to Gaussian ARTMAP during inference. In case of a mismatch, \ARTa's vigilance is temporarily changed to
\begin{equation}
\rho_a = M_J^a - \delta,~ 0 \leq \delta \ll M_J^a,    
\end{equation}
\noindent where $M_J^a$ is computed using Eq.~(\ref{Eq:BA_2}). The search continues until another resonant node is found or a new one is created. When learning is finally ensued, the matrix $\bm{N}$ entry $n_{KJ}$ (class K and \ARTa's resonant node $J$ association) is updated as
\begin{equation}
n_{KJ}(new) = n_{KJ}(old) + 1. 
\end{equation}

\textbf{Inference.} During testing, the class of an unseen sample is predicted using 
\begin{equation}
K = \argmax\limits_k \left( \hat{p}(c_k^b | \bm{x}^a) \right),
\end{equation}
\noindent where
\begin{equation}
\hat{p}(c_k^b | \bm{x}^a) = \dfrac{\sum\limits_{j=1}^{N_a}\hat{p}(c_k^b|c_j^a) \hat{p}(\bm{x}^a|c_j^a) \hat{p}(c_j^a)}{\sum\limits_{i=1}^{N_b}\sum\limits_{l=1}^{N_a} \hat{p}(c_i^b|c_l^a) \hat{p}(\bm{x}^a|c_l^a) \hat{p}(c_l^a)},
\end{equation}
\begin{equation}
\hat{p}(c_j^a) = \dfrac{\sum\limits_{k=1}^{N_b}n_{kj}}{\sum\limits_{l=1}^{N_a} \sum\limits_{k=1}^{N_b} n_{kl}},    
\end{equation}
\begin{equation}
\hat{p}(c_k^b|c_j^a) = \dfrac{n_{kj}}{\sum\limits_{i=1}^{N_b}n_{ij}}.   
\end{equation}

Bayesian ARTMAP variants have been developed for various tasks, such as semi-supervised learning~\cite{Tang.2010, Nooralishahi.2018a} and associative memory~\cite{Chin.2016a}.

\subsubsection{Generalized ART}
\label{Sec:GART}

The Generalized ART (GART)~\cite{Yap2008} is a hybrid model that combines a Gaussian ARTMAP~\cite{williamson1996} (Sec.~\ref{Sec:GA}) variant to cluster samples in the input space and a generalized regression neural network (GRNN)~\cite{Specht1991} to perform prediction. In this model, the mapping is one-to-one (bijective) and thus $N_a=N_b=N$.  

\textbf{Training.}  Like Gaussian and Bayesian ARTs (Secs.~\ref{Sec:GA} and~\ref{Sec:BA}, respectively), the two modified Gaussian ART modules A and B use Bayes' theorem to compute their activation functions (posterior probability as in Eq.~(\ref{Eq:GA_1})), where the prior $\hat{p}(c_j^a)$ is estimated using Eq.~(\ref{Eq:GA_3}). Again, the evidence $\hat{p}(\bm{x}^a)$ is the same for all categories and thus does not influence the WTA competition. The conditional probability estimate $\hat{p}(\bm{x}^a | \theta_j^a )$ is given by
\begin{equation}
\hat{p}(\bm{x}^a | \theta_j^a ) \propto exp\left[ -\dfrac{1}{2} \lambda (\delta_{j}^{a} (\bm{x}^a)) \right],
\label{Eq:GART_4}
\end{equation}
\noindent where $\lambda (\delta_{j}^{a})$ is defined an $\varepsilon$-insensitive loss function to handle outliers and noisy data
\begin{equation}
\lambda (\delta_{j}^{a}) = 
\begin{cases} 
0, & \mbox{if } \delta_{j}^{a} \leq \varepsilon_a\\
\delta_{j}^{a} - \varepsilon_a, & \mbox{otherwise} \\
\end{cases},
\label{Eq:GART_5}
\end{equation} 
\noindent $\varepsilon_a \geq 0$ is a user-defined parameter (if $\varepsilon=0$, then Eq.~(\ref{Eq:GART_5}) reduces to the Laplacian loss function), and 
\begin{equation}
\delta_{j}^{a}(\bm{x}^a) = \sum\limits_{i=1}^{d} \abs[\Big]{\dfrac{\mu_{ji}^a - x_{i}}{\sigma_{ji}^a}},
\label{Eq:GART_6}
\end{equation}
\noindent the parameters $\bm{\mu}_j^a$, $\bm{\sigma}_j^a$ and $n_j^a$ correspond to the centroid, standard deviation and sample count of \ARTa's category~$j$. 

When \ARTa's BMU is selected via WTA, the following match functions are computed
\begin{equation}
M_J^a = \hat{p}(\bm{x}^a | c_j^a ),
\label{Eq:GART_7}
\end{equation}
\begin{equation}
M_J^b = \hat{p}(x^b | c_j^b ),
\label{Eq:GART_8}
\end{equation}
\noindent where the systems enters a resonant state if $M_J^m \geq \rho_m$, \mbox{$\rho_m \in [0,1]$}, $m \in \{a,b\}$, i.e., if both vigilance tests are simultaneously satisfied. If learning is ensued, then
\begin{equation}
n_J^a(new) = n_J^a(old) + 1, 
\label{Eq:GART_9}
\end{equation}
\begin{equation}
\bm{\mu}_J^a(new) =  \left[1-\frac{1}{n_J^a(new)}\right]\bm{\mu}_J^a(old) +\frac{1}{n_J^a(new)}\bm{x}^a,
\label{Eq:GART_10}
\end{equation}
\begin{equation}
\bm{\sigma}_{J}^a(new)   =  \left[1-\frac{1}{n_J^a(new)}\right]\bm{\sigma}_{J}^a(old) 
 + \dfrac{1}{n_J^a(new)}\abs[\Big]{\bm{\mu}_J^a(new) -\bm{x}^a}.
\label{Eq:GART_11}
\end{equation}
\noindent where the standard deviation update is based on the Laplacian distribution.

For a newly created category, $n_J^a(new) = 1$, $\bm{\mu}_J^a = \bm{x}^a$, $\bm{\sigma}_{J}^a = \gamma_a$, $\bm{\sigma}_{J}^a = \sigma_{init}^2\Vec{\bm{1}}$ (user-defined initial standard deviation). Similar dynamics hold for \ARTb, and for both modules $N=N+1$.

\textbf{Inference.} A prediction for an unseen sample $\bm{x}$ is made using
\begin{equation}
f(\bm{x}^a) = \dfrac{\sum\limits_{j=1}^{N} \dfrac{\hat{p}(c_j^a | \bm{x}^a)}{\sigma_j^b} \mu_j^b }{\sum\limits_{j=1}^{N} \dfrac{\hat{p}(c_j^a | \bm{x}^a)}{\sigma_j^b}}, f(\bm{x}^a) \in \mathbb{R}^1.
\label{Eq:GART_12}
\end{equation}

The enhanced GART (EGART)~\cite{Yap2010} adds network pruning and rule extraction strategies to the Generalized ART model. Moreover, $\hat{p}(\bm{x}^a | c_j^a )$ is formally defined as the Laplacian likelihood function
\begin{equation}
\hat{p}(\bm{x}^a | c_j^a ) = \dfrac{1}{2^d\prod\limits_{i=1}^d \sigma_{ji}^a} exp\left[ -\sum\limits_{i=1}^{d} \dfrac{1}{\sigma_{ij}^a} \abs[\Big]{\mu_{ij}^a - x_{i}^a} \right],
\label{Eq:GART_13}
\end{equation}
\noindent and, like Gaussian ART, \ARTa's match function is a normalized version of Eq.~(\ref{Eq:GART_13})
\begin{equation}
M_J^a = \hat{p}(\bm{x}^a | c_j^a ) = exp\left[ -\sum\limits_{i=1}^{d} \dfrac{1}{\sigma_{ij}^a} \abs[\Big]{\mu_{ij}^a - x_{i}^a} \right],
\label{Eq:GART_14}
\end{equation}
\noindent where for resonance to occur in \ARTa, $M_J^a \geq \rho_a$ must be satisfied. The match tracking mechanism compares $M_J^b$ to~$\rho_b$
\begin{equation}
M_J^b = \hat{p}(\bm{x}^a | c_j^a ) = exp\left[ -\sum\limits_{i=1}^{d} \dfrac{1}{\sigma_{ij}^a} \abs[\Big]{\mu_{ij}^a - x_{i}^a} \right],
\end{equation}
\noindent and if it is not satisfied, then the match tracking mechanism temporarily raises $\rho_a$, inhibits the current winner category $J$ and resumes the search. The learning and prediction mechanisms are the same as Generalized ART.

The improved GART (IGART)~\cite{Yap2011} builds upon the enhanced GART by incorporating an ordering algorithm~\cite{dagher1999} to determine the order of input presentation as well as providing multivariate prediction $f(\bm{x}^a) \in \mathbb{R}^L$ when in inference mode:
\begin{equation}
f_l(\bm{x}^a) = \dfrac{\sum\limits_{j=1}^{N} \dfrac{\hat{p}(c_j^a | \bm{x}^a)}{\sigma_{jl}^b} \mu_{jl}^b }{\sum\limits_{j=1}^{N} \dfrac{\hat{p}(c_j^a | \bm{x}^a)}{\sigma_{jl}^b}},~l \in \{ 1,...,L \}.
\end{equation}

\subsubsection{Self-supervised ARTMAP}
\label{Sec:SSARTMAP}

The self-supervised ARTMAP (SSARTMAP)~\cite{Amis.2010a} is a model designed for self-supervised learning applications. This machine learning modality consists of a supervised learning phase, in which only certain data features are specified, followed by an unsupervised phase, in which all the data features are specified. Similar to fuzzy ARTMAP (Sec.~\ref{Sec:FAM}), this model's LTM is defined by $\bm{\theta} = \{\bm{w} = [\bm{u}, \bm{v}^c] \}$, whose geometric interpretation are hyperrectangles in the data space. An artifact of this learning modality is the ``undercommitted'' categories, defined by the presence of ``undercommitted'' features (i.e., $\exists i $ : $u_{i} > v_{i}$).

\textbf{Training.} During the first phase, where supervised learning takes place for a pre-defined number of epochs, only $\bar{d}$ features are presented to the network. That is, a sample $\bm{x}$ carries information only with respect to a subset of features. The latter are complement coded, whereas the unspecified features are set to $1$'s:
\begin{equation}
x_i = 
\begin{cases} 
x_i, & \mbox{if } i=1,..., \bar{d} \\
1 - x_i, & \mbox{if } i= d+1,..., d+\bar{d} \\
1, & \mbox{otherwise} 
\end{cases},
\label{Eq:SSARTMAP_1}
\end{equation}
\noindent such that $\norm{\bm{x}}_1 = 2d - \bar{d}$ and $\bar{d} \leq d$. Then, an activation function based on choice-by-difference~\cite{Gjaja.1994a} is computed for each category $j$:
\begin{equation}
T_j = \dfrac{\left(2d - \norm{\bm{x}}_1 \right) - \left( \norm{\bm{w}_j}_1 - \norm{\bm{x} \wedge \bm{w}_j^a}_1 \right)  }{1-\gamma \phi_j} - \alpha\left( d - \norm{\bm{w}_j}_1 \right),
\label{Eq:SSARTMAP_2}
\end{equation}
\noindent where \mbox{$0<\alpha<1$} is the choice parameter, \mbox{$0<\gamma<1-\alpha$} is the undercommittement factor, and
\mbox{$0\leq\phi_j\leq1$} is the degree of undercommittement of category $j$, defined as
\begin{equation}
\phi_j = \dfrac{1}{d} \sum\limits_{i=1}^{d} \left[ u_{j,i} - v_{j,i} \right]^+  = \dfrac{1}{d} \sum\limits_{i=1}^{d} \left[ w_{j,i} - (1-w_{j,d+i}) \right]^+,    
\label{Eq:SSARTMAP_3}
\end{equation}
\noindent where $[\cdot]^+$ is a rectifier operator. After the activation functions are computed, a subset of highly active categories is formed: \mbox{$\Lambda = \{ j : T_j \geq T^u = \alpha d\}$}, where $T^u$ is the activation function of an uncommitted category (initialized as $\bm{w} = \Vec{\bm{1}}$). If $\Lambda = \{ \emptyset \}$, then an uncommitted category is recruited and permanently mapped to the class label paired with the current input sample. Otherwise, the mapping of the resonant committed category $J$ is assessed. If it is correct, then learning is ensued as
\begin{equation}
\bm{w}_J(new) = \bm{w}_J(old) - \beta_1 \left[ \bm{w}_J(old) - \bm{x} \right]^+,
\label{Eq:SSARTMAP_4}
\end{equation}
\noindent where $[\cdot]^+$ is a component-wise rectifier operator and \mbox{$\beta_1 \in (0,1]$} is the learning parameter of this first training phase. If the prediction is incorrect, then the match tracking mechanism (user-defined MT+ or MT-, see Sec.~\ref{Sec:AMIC}) inhibits the resonant neuron, slightly changes the baseline vigilance parameter $\bar{\rho}$ and restarts the search. 

During the second phase, unsupervised learning takes place for another pre-defined number of epochs. As opposed to the previous phase, all the data features are presented (i.e., $\bm{x} = [\bm{x}, \Vec{\bm{1}} - \bm{x}]$), and distributed representation is employed. Additionally, the network runs in slow learning mode, and no mismatches occur (the vigilance parameter is set to zero). Particularly, if $\Lambda = \{ \emptyset \}$, then no learning takes place. Next, the activation functions are computed using Eq.~(\ref{Eq:SSARTMAP_2}). The distributed activity $\bm{y}^{(F_2)}$ of layer $F_2$ is established using the IG CAM rule described in Sec.~\ref{Sec:defAM} (Eqs.~(\ref{Eq:defAM_2}) and~(\ref{Eq:defAM_3})). All weight vectors are thus updated using the distributed instar learning law
\begin{equation}
\bm{w}_j(new) = \bm{w}_j(old) - \beta_2 \left[ y_j\Vec{\bm{1}} - \left(\Vec{\bm{1}} - \bm{w}_j(old)\right) - \bm{x} \right]^+,
\label{Eq:SSARTMAP_5}
\end{equation}
\noindent where $j \in \Lambda $, and \mbox{$\beta_2 \in [0,1]$} is the learning parameter of the second training phase.

\textbf{Inference.} In inference mode, the self-supervised ARTMAP dynamics are identical to the unsupervised training stage, except that no learning takes place. Predictions are made using Eqs.~(\ref{Eq:AM_5}) and~(\ref{Eq:AM_5a}) in Sec.~\ref{Sec:AM}.

\subsubsection{Biased ARTMAP}
\label{Sec:biasAM}

Biased ARTMAP (bARTMAP)~\cite{Carpenter.2010a} augments fuzzy ARTMAP with a featural biasing mechanism to handle ordering effects that arise in fast online learning mode. Said mechanism temporarily alters the network's focus among the input sample features following a predictive error.

\textbf{Training.} During training, the choice-by-difference activation function (Eq.~(\ref{Eq:defAM_1})) is used to find the winner category $J$, whose match function is computed as
\begin{equation}
M_J = \frac{\norm{\tilde{\bm{y}}^{(F_1)}}_1}{\norm{\tilde{\bm{x}}}_1},
\label{Eq:biasAM_1}
\end{equation}
\begin{equation}
\tilde{\bm{x}} = \left[ \bm{x} - \bm{e} \right]^+,
\label{Eq:biasAM_2}
\end{equation}
\begin{equation}
\tilde{\bm{y}}^{(F_1)}= \left[ \bm{y}^{(F_1)} - \bm{e} \right]^+,
\label{Eq:biasAM_3}
\end{equation}
\noindent where $[\cdot]^+$ is a component-wise rectifier operator, $\tilde{\bm{x}}$~is the biased complement coded input vector, $\tilde{\bm{y}}^{(F_1)}$~is the biased \Fin\ activity and $\bm{e} \in \mathbb{R}^{2d}$ is the bias vector, which is set to~$\Vec{\bm{0}}$ at the beginning of each input presentation (such that $\tilde{\bm{x}} = \bm{x}$ and $\tilde{\bm{y}}^{(F_1)} = \bm{y}^{(F_1)}$). If the category $J$ successfully passes the vigilance test (i.e., if it satisfies $M_J \geq \rho$) and is mapped to the correct class, then the learning dynamics are identical to fuzzy ART's (Eq.~(\ref{Eq:FA_6}) in Sec.~\ref{Sec:FA}). Alternately, if the prediction based on the resonant category is incorrect, then the bias vector is updated using Eq.~(\ref{Eq:biasAM_4}), 

\begin{equation}
e_i(new) = 
\begin{cases} 
e_i(old), & \mbox{if } \lambda \left[ \left[ y_i^{(F_1)} - e_i(old) \right]^+ - \dfrac{\norm{\bm{y}^{(F_1)}}_1}{2d} \right] \leq 0 \\
e_i(old), & \mbox{if } e_i(old) \geq \lambda \left[ \left[ y_i^{(F_1)} - e_i(old) \right]^+ - \dfrac{\norm{\bm{y}^{(F_1)}}_1}{2d} \right] > 0 \\
\dfrac{\left[ y_i^{(F_1)} - \dfrac{\norm{\bm{y}^{(F_1)}}_1}{2d} \right]}{1 + \lambda^{-1}}, & \mbox{if } y_i^{(F_1)} > e_i(old) \mbox{ and } \lambda \left[ \left[ y_i^{(F_1)} - e_i(old) \right] - \dfrac{\norm{\bm{y}^{(F_1)}}_1}{2d} \right] > e_i(old)
\end{cases},~\lambda \geq 0,
\label{Eq:biasAM_4}
\end{equation}
\noindent the match tracking algorithm alters the vigilance parameter value (MT-, Sec.~\ref{Sec:AMIC}) and the search resumes. The bias strength parameter $\lambda$ in Eq.~(\ref{Eq:biasAM_4}) can be selected by cross-validation procedures (note that setting $\lambda=0$ implies an unbiased model, i.e., fuzzy ARTMAP).

\textbf{Inference.} In prediction mode, biased ARTMAP behaves identically to fuzzy ARTMAP (Sec.~\ref{Sec:FAM}).

\subsubsection{TopoART-C}
\label{Sec:TopoART_C}

TopoART-C~\cite{Tscherepanow2012c} is an incremental classifier based on fuzzy topoART (Sec.~\ref{Sec:TopoART}). In this architecture, each topoART module (A and B) is augmented with a classification layer F\textsubscript{3} that is connected to the category layer F\textsubscript{2}. Additionally, module B is endowed with a mask layer F\textsubscript{0} preceding its feature layer F\textsubscript{1} to handle incomplete data. 

\textbf{Training.} During training, the vigilance tests are layered: the first is unsupervised and equal to fuzzy ART's (Sec.~\ref{Sec:FA}), while the second is supervised and determines whether a correct class prediction was made. These must be simultaneously satisfied for the system to enter a resonant state and learn.  

\textbf{Inference.} 
Prediction is made using topoART B, since topoART A is only used to filter noise and is therefore disregarded. Specifically, such a prediction depends on whether or not an unknown sample is completely enclosed by at least one category (which implies alternative activation function (Eq.~(\ref{Eq:TopoART_1})) equal to $1$). In the affirmative case, the system predicts the class associated with the smallest node (measured using Eq.~(\ref{Eq:FA_5})). In the negative case, the system makes a prediction based on a subset of highly active categories. Note that if the sample has missing values, then only non-missing attributes are used in the computations.  

\subsection{Architectures for regression}
\label{Sec:SL_Reg}

The supervised ART models described so far have been primarily used for classification purposes. Although, in theory, all ARTMAP variants may be used to perform regression tasks~\cite{Sasu2013}. For instance, fuzzy ARTMAP was shown to be a universal function approximator in~\cite{Verzi.2003a}. This section reviews architectures developed specifically for incremental function approximation/interpolation. An experimental comparative study on some of these ART-based regression models can be found in~\cite{Sasu.2012a}.

\subsubsection{PROBART}
\label{Sec:PROBART}

The PROBART model~\cite{Marriott1995} is a fuzzy ARTMAP variant designed to approximate noisy continuous mappings. It has a distinct map field dynamic, whose activity is given by
\begin{equation}
\bm{y}^{{(F^{ab})}} = 
\begin{cases} 
\bm{w}_J^{ab} + \bm{y}^{{(F_2^{b})}} , & \mbox{if both ARTs are active }\\
\bm{w}_J^{ab}, & \mbox{if only ART\textsubscript{a} is active} \\
\bm{y}^{{(F_2^{b})}}, & \mbox{if only ART\textsubscript{b} is active} \\
\Vec{\bm{0}}, & \mbox{otherwise} \\
\end{cases}.
\label{Eq:PROBART_1}
\end{equation} 

This change turns the map field's weight matrix $\bm{W}^{ab}$ into a frequency counter for the co-occurrence of resonant categories in both ART modules (i.e., it records the number of associations between nodes of \ARTa\ and \ARTb), thereby storing probabilistic information. Note that, in this model it is initialized as \mbox{$\bm{W}^{ab} = \bm{0}$}. 

\textbf{Training.}  PROBART does not possess a match tracking mechanism, since it is adequate for classification tasks~\cite{Marriott1995} and rule extraction~\cite{Carpenter1995a} but not for regression~\cite{Srinivasa1997}. Moreover, it directly affects the probability estimation process. Therefore, \ARTa's vigilance remains fixed. When learning is ensued, F\textsuperscript{ab} weights are updated as
\begin{equation}
\bm{w}_J^{ab}(new) = \bm{w}_J^{ab}(old) + \bm{y}^{{(F^{ab})}},
\label{Eq:PROBART_2}
\end{equation}
\noindent considering that \ARTa's and \ARTb's resonating nodes are $J$ and $K$, respectively.

\textbf{Inference.} The $l^{th}$ component of the prediction $\hat{f}(\bm{x}^a)$, when \ARTa's resonating category is $J$, is computed as
\begin{equation}
\hat{f}_l(\bm{x}^a) = \frac{1}{\norm{\bm{w}_J^{ab}}_1} \sum\limits_{k=1}^{N_b} w_{Jk}^{ab} w^b_{kl}= \sum\limits_{k=1}^{N_b} p_{Jk} w^b_{kl},
\label{Eq:PROBART_3}
\end{equation}
\noindent where \mbox{$p_{Jk} = \hat{p}(c_k^b | c_J^a) = \frac{w_{Jk}^{ab}}{\norm{\bm{w}_J^{ab}}_1}$}, $\bm{w}_J^{ab}$ is the $J^{th}$ row of $\bm{W}^{ab}$, $\norm{\bm{w}_J^{ab}}_1$ is the total number of samples associated with \ARTa's node $J$ across all \ARTb\ nodes, $w_{Jn}^{ab}$ is the number of co-activations of \ARTa's node $J$ and \ARTb's node $n$, $l \in \{1,...,d_b\}$ and $d_b$ is the original non-complement coded dimension (number of features) of \ARTb's input samples. The prediction is thus an average weighted by the conditional probabilities. Note that, to perform accurate mappings, PROBART requires large \ARTa\ vigilance parameter values, consequently generating a large number of categories~\cite{Sanchez2002}. 

PROBART's generalization capability is limited by its WTA prediction, which is addressed by Modified PROBART~\cite{Srinivasa1997} via distributed prediction. The training process is identical for both models; the difference lies in the inference mode. Each feature $l$ of the prediction $\hat{f}'(\bm{x}^a)$ is computed as 
\begin{equation}
\hat{f}_l'(\bm{x}^a) = \frac{\sum\limits_{m \in \mathcal{S}} M_m \gamma_m \hat{f}_{m,l}(\bm{x}^a)}{\sum\limits_{m \in \mathcal{S}} M_m \gamma_m},   
\end{equation}
\noindent where $\mathcal{S}$ is the set of \ARTa's resonant nodes for input $\bm{x}^a$ (i.e., $M_m \geq \rho_a$, $M_m$ is the match function value of \ARTa's neuron $m$), $\hat{f}_{m,l}(\bm{x}^a)$ is \ARTa's neuron $m$ prediction for feature $l$ computed from Eq~(\ref{Eq:PROBART_3}) and $\gamma_m$ is \ARTa's neuron $m$'s frequency of winning. Concretely, the prediction is an average weighted by \ARTa's nodes' match function values and instance countings. The size of the set $\mathcal{S}$ considered for distributed prediction is defined for each component $l$ using a heuristic that minimizes the root mean squared error over the entire training set. 

\subsubsection{FasArt and FasBack}
\label{Sec:FasART}

FasArt~\cite{Izquierdo1996, Izquierdo2001} is a neuro-fuzzy system that reinterprets fuzzy ARTMAP (Sec.~\ref{Sec:FAM}) as a fuzzy logic system by defining categories as decomposable fuzzy sets in their data spaces (universes).

\textbf{Training.} The training dynamics are identical to fuzzy ARTMAP's (\ARTa, \ARTb, and the map field), with the exception that the activation function, now also regarded as a fuzzy membership function, is defined as
\begin{equation}
T_j = \prod\limits_{i=1}^d T_{j,i},    
\end{equation}
\noindent where $T_{j,i}$ is a triangular fuzzy membership function
\begin{equation}
T_{j,i} =  
\begin{cases} 
\left[ \dfrac{\gamma(x_i - w_{j,i}) + 1}{\gamma(c_{j,i} - w_{j,i}) + 1} \right]^+, & \mbox{if } x_i \leq c_{j,i} \\
\left[ \dfrac{\gamma(1 - x_i - w_{j,d+i}) + 1}{\gamma(1 - c_{j,i} - w_{j,d+i}) + 1} \right]^+, & \mbox{if } x_i > c_{j,i} \\
\end{cases},
\end{equation}
\noindent the parameter $\gamma$ is the fuzzification rate that controls the width of the fuzzy set support (and consequently the generalization capabilities) and $\bm{c}_j$ is the centroid associated with category $j$. The fuzzy support associated category $j$ is thus defined by $\bm{w}_j$, $\bm{c}_j$ and $\gamma$. The weight vector $\bm{w}_J$ of a resonant category $J$ is updated using fuzzy ART's learning dynamics (Eq.~(\ref{Eq:FA_6}) in Sec.~\ref{Sec:FA}), whereas the centroid is updated using
\begin{equation}
\bm{c}_J(new) =  (1 - \beta_c)\bm{c}_J(old) + \beta_c \bm{x},   
\end{equation}
\noindent where $\beta_c \in (0,1]$ is the centroid's learning parameter. This learning dynamic is the same for both ART modules. However, it should be noted that note that the LTMs of \ARTa\  are also subjected to the constraint of making a correct prediction.

\textbf{Inference.} The prediction of each feature $m$ is obtained using the following defuzzification procedure (average of fuzzy set centroids):
\begin{equation}
\hat{f}_m(\bm{x}^a) = \dfrac{\sum\limits_{k=1}^{N_b}\sum\limits_{j=1}^{N_b} c^b_{k,m} w^{ab}_{j,k}T^a_j}{\sum\limits_{k=1}^{N_b}\sum\limits_{j=1}^{N_b} w^{ab}_{j,k}T^a_j},  
\end{equation}
\noindent where $T^a_j$ is the activation of \ARTa's category $j$, $c^b_{k,m}$ is the $m^{th}$ component of \ARTb's centroid $\bm{c}^b_{k}$ associated with category $k$ and $w^{ab}_{j,k}$ is the $\{j,k\}$ entry of the map field matrix $\bm{W}^{ab}$. Note that FasArt is a universal function approximator~\cite{Izquierdo2001}.

For fine-tuning purposes, particularly to improve performance and network compactness (i.e., to reduce category proliferation), FasBack~\cite{Izquierdo1997, Izquierdo2001} enhances FasArt with error-based learning by using the gradient descent optimization method to adapt some of its parameters 
\begin{equation}
\bm{p}(new) = \bm{p}(old) - \eta \dfrac{\partial \mathcal{E}}{\partial \bm{p}(old)}, 
\end{equation}
\noindent where $\bm{p} \in \{\bm{c}_j^a, \bm{c}_k^b, w^{ab}_{i,j}\}$, $\eta$~is the learning rate, $\mathcal{E}$ is error to be minimized
\begin{equation}
\mathcal{E} = \dfrac{1}{2}\norm{\hat{f}(\bm{x}^a) - \bm{d}}_2^2,    
\end{equation}
\noindent and $\hat{f}(\bm{x}^a)$ and $\bm{d}$ are the system's prediction and the desired response, respectively. Note that two learning cycles are performed: a match-based one followed by an error-based one.

FasArt has spawned many variants including recurrent~\cite{Palmero2000}, distributed~\cite{Hernandez.2003a} and dynamic~\cite{Izquierdo2009} models.

\subsubsection{Fuzzy ARTMAP with input relevances}
\label{Sec:FAMR_R}
The fuzzy ARTMAP with input relevances (FAMR)~\cite{Andonie2006, Andonie2003c}, when used for regression applications, makes predictions similarly to PROBART (Eq.~(\ref{Eq:PROBART_3}) in Sec.~\ref{Sec:PROBART}). Particularly, PROBART is said to be a special case of FAMR with its parameters set to $q_0=0$, $q_t = q \in (0,\infty)$ (constant) and $\rho_{ab}=0$.

\subsubsection{Generalized ART}
\label{Sec:GART2}

The generalized ART and its variants (Sec.~\ref{Sec:GART}) can be used for both classification and regression problems, for instance, by setting $\rho_b=1$ for the former and $\rho_b=\rho_a$ for the latter~\cite{Yap2008}. 

\subsubsection{TopoART-R}
\label{Sec:TopoART_R}

TopoART-R~\cite{Tscherepanow2011a} is a variant of fuzzy topoART (Sec.~\ref{Sec:TopoART}) designed for regression purposes. In this model, topoART module B is endowed with an input control layer F\textsubscript{0} preceding its feature layer F\textsubscript{1} to process samples with missing attributes (i.e., make predictions).

\textbf{Training.} TopoART-R training is similar to topoART (Sec.~\ref{Sec:TopoART}); however, it does not perform topological learning. Particularly, the complement coded independent and dependent variables are concatenated as a single input vector to be presented to the network. During the vigilance test stage, two match functions are independently computed for the dependent and independent variables. 

\textbf{Inference.} Similar to topoART-C (Sec.~\ref{Sec:TopoART_C}), during testing, 
module A is disregarded, the activation function used is given by Eq.~(\ref{Eq:TopoART_1}) in Sec.~\ref{Sec:TopoART} and the prediction strategy depends on whether or not the input sample is fully enclosed by at least one ``partial'' category (i.e., a hyperrectangle in the multidimensional space formed by the non-missing attributes of the presented sample, from which a prediction is sought). In the affirmative case, a ``temporary'' category is created from the intersection of these ``partial'' categories. Then, a prediction is the center of the interval defined by the  upper and lower bound components of the ``temporary'' category that correspond to a given missing attribute (dependent and independent variables are treated as missing and non-missing, respectively). In the negative case, the ``temporary'' category is created as a weighted average of a subset of highly active nodes, and then the prediction is carried out as previously described.

\subsubsection{Bayesian ARTMAP for regression}
\label{Sec:BAM_reg}

The Bayesian ARTMAP for regression (BAR)~\cite{Sasu2013} uses two Bayesian ART modules to perform clustering on both the input and the output spaces. All the dynamics of Bayesian ARTMAP discussed in Section~\ref{Sec:BAM} hold, except for the for the prediction (i.e., the function approximation) which is given by: 
\begin{equation}
\hat{f}(\bm{x}^a) = \sum\limits_{k=1}^{N_b} \hat{p}(c_k^b | \bm{x}^a) \bm{\mu}_k^b,
\label{Eq:BAR_1}
\end{equation}
\noindent where $\hat{p}(c_k^b | \bm{x}^a)$ is computed as described in Section~\ref{Sec:BAM}. The Bayesian ARTMAP for regression was shown to be a universal function approximator~\cite{Sasu2013}.

\subsection{Summary}

Table~\ref{Tab:ARTs_SL} summarizes the architectures discussed in terms of their training, inference/testing and the map field's mapping characteristics. Particularly, it lists if winner-takes-all (WTA) or distributed (D) coding is employed by these networks and whether the learned mapping is many-to-one ($ART_a \mapsto ART_b$, surjective) or many-to-many (many-to-one and one-to-many).  

\begin{table}[!htb]
\centering
\caption{Summary of supervised ART models' key characteristics.}
\begin{threeparttable}
\resizebox{\textwidth}{!}{
\begingroup\setlength{\fboxsep}{0pt}
\colorbox{lightgray}{
\begin{tabular}{lllll}
\toprule
ART model & Training & Inference & Mapping & Reference(s) \\
\midrule
\midrule
\multicolumn{4}{l}{Classification} \\
\midrule
ARTMAP                      & WTA   & WTA   & many-to-one  & \cite{Carpenter1991a} \\
Fuzzy ARTMAP                & WTA   & WTA   & many-to-one  & \cite{carpenter1992} \\
Fuzzy Min-Max               & WTA   & WTA   & many-to-one  & \cite{Simpson1992} \\
Fusion ARTMAP               & WTA   & WTA   & many-to-many & \cite{Asfour.1993} \\
LAPART 1                    & WTA   & WTA   & many-to-one  & \cite{Healy1993} \\
ART-EMAP                    & WTA   & D     & many-to-one  & \cite{Carpenter1995} \\
ARAM                        & WTA   & WTA   & many-to-many & \cite{Tan1995b} \\
Gaussian ARTMAP             & WTA   & D     & many-to-one  & \cite{williamson1996} \\
Probabilistic fuzzy ARTMAP  & WTA   & D     & many-to-many & \cite{Lim1997} \\
ARTMAP IC                   & WTA   & D     & many-to-one  & \cite{Carpenter1998b} \\
distributed ARTMAP          & WTA/D & D     & many-to-one  & \cite{Carpenter1998a} \\
Hypersphere ARTMAP          & WTA   & WTA   & many-to-one  & \cite{anagnostopoulos2000} \\
Ellipsoid ARTMAP            & WTA   & WTA   & many-to-one  & \cite{anagnostopoulos2001,anagnostopoulos2001b} \\
$\mu$-ARTMAP                & WTA   & WTA   & many-to-many & \cite{Sanchez2002} \\
Default ARTMAP 1            & WTA   & D     & many-to-one  & \cite{Carpenter2003} \\
Boosted ARTMAP              & WTA   & WTA   & many-to-many & \cite{Verzi2006} \\
FAMR                        & WTA   & WTA & many-to-many & \cite{Andonie2006} \\
Default ARTMAP 2            & WTA/D & D     & many-to-one  & \cite{Amis2007} \\
Bayesian ARTMAP             & WTA   & D     & many-to-many & \cite{vigdor2007} \\
Generalized ART             & WTA   & D     & one-to-one   & \cite{Yap2008} \\
Self-supervised ARTMAP      & WTA/D & D     & many-to-one  & \cite{Amis.2010a} \\
Biased ARTMAP               & WTA   & WTA   & many-to-one  & \cite{Carpenter.2010a} \\
TopoART-C                   & WTA   & D     & many-to-one  & \cite{Tscherepanow2012c} \\
\midrule
\multicolumn{4}{l}{Regression} \\
\midrule
PROBART                     & WTA   & WTA   & many-to-many & \cite{Marriott1995} \\
Modified PROBART            & WTA   & D     & many-to-many & \cite{Srinivasa1997} \\
FasART/FasBack              & WTA   & D     & many-to-one  & \cite{Izquierdo2001} \\
FAMR                        & WTA   & WTA   & many-to-many & \cite{Andonie2006} \\
Generalized ART             & WTA   & D     & one-to-one   & \cite{Yap2008} \\
TopoART-R                   & WTA   & D     & many-to-many & \cite{Tscherepanow2011a} \\
Bayesian ARTMAP             & WTA   & D     & many-to-many & \cite{Sasu2013} \\
\bottomrule
\end{tabular}
}\endgroup
}
\end{threeparttable}
\label{Tab:ARTs_SL}
\end{table}

\section{ART models for reinforcement learning}
\label{Sec:RL}
The ART models described in the following subsections are used to perform reinforcement learning in which agents learn in real-time, incrementally and continuously by interacting with a complex and dynamic environment. ART-based reinforcement learning systems have found growing applications, for instance, in the computer games~\cite{Wang.2009a, Wang.2015a, Silva.2018a} and situation awareness~\cite{Brannon.2006a, brannon2009} domains.

\subsection{Reactive FALCON}
\label{Sec:RFALCON}

The reactive fusion architecture for learning, cognition, and navigation (R-FALCON)~\cite{tan2004} is a fusion ART-based model (Sec.~\ref{Sec:Fusion_ART}) that possesses three channels (or \Fin\ layers), viz., the sensory field ($\text{F}_\text{1}^\text{s}$), the motor field ($\text{F}_\text{1}^\text{a}$) and the feedback field ($\text{F}_\text{1}^\text{r}$), which are used to learn mappings across states ($\bm{s} =[s_1,...,s_n]$, where $s_j \in [0,1], \forall j$), actions ($\bm{a} = [a_1,...,a_m]$, $a_i \in [0,1], \forall i$), and rewards ($r \in [0,1]$), respectively. The general sense-act-learn dynamics of R-FALCON are described next. 

\textbf{Prediction.} Consider an agent currently at a state~$\bm{s}$. The inputs to R-FALCON's $\text{F}_\text{1}^\text{s}$, $\text{F}_\text{1}^\text{a}$ and $\text{F}_\text{1}^\text{r}$ layers are set to $\bm{x}^s = \bm{s}$, $\bm{x}^a = \Vec{\bm{1}}$ and $\bm{x}^r = [1,0]$, respectively. Note that the feedback field is modeled using $\bm{x}^r = [r, 1-r]$. A node~$J$ is then selected via a WTA competition (node~$J$ maximizes Eq.~(\ref{Eq:fusion_1}) in Sec.~\ref{Sec:Fusion_ART}). This setting of $\bm{x}^r$ used for prediction biases selection towards maximal rewards.

\textbf{Action selection policy.} The activity of layer $\text{F}_\text{1}^\text{a}$, given by
\begin{equation}
\bm{y}^{(F_1^a)} = \bm{x}^a \wedge \bm{w}_J^a = \bm{w}_J^a,
\end{equation}
\noindent is used to select the action $I$ as
\begin{equation}
I = \argmax\limits_{1 \leq i \leq m} \left( y_i^{(F_1^a)}\right).
\end{equation}

The agent performs the selected action $I$ and then enters a new state $\bm{s}'$.

\textbf{Learning.} Learning is ensued similarly to fusion ART (Sec.~\ref{Sec:Fusion_ART}) using the appropriate \Fin\ layers' inputs, which depend on the feedback received from performing the selected action:
\begin{itemize}
\item Positive feedback (reward): \Fin\ layers' inputs are set to $\bm{x}^s = \bm{s}$, $\bm{x}^a = \bm{a}$, and $\bm{x}^r = \bm{r}$.
\item Negative feedback (penalty): \Fin\ layers' inputs are set to $\bm{x}^s = \bm{s}$, $\bm{x}^a = \bar{\bm{a}} = \Vec{\bm{1}} - \bm{a}$, and $\bm{x}^r = \bar{\bm{r}} = \Vec{\bm{1}} - \bm{r}$.
\end{itemize}

R-FALCON suffers from category proliferation, so it must undergo pruning heuristics to enhance interpretability and scalability. Moreover, it can only effectively handle problems with immediate rewards.

\subsection{Temporal difference FALCON}
\label{Sec:TDFALCON}

The temporal difference fusion architecture for learning, cognition, and navigation (TD-FALCON)~\cite{tan2006, tan2008} is a fusion ART-based model developed to effectively handle not only problems with immediate rewards but also problems with delayed rewards. This is accomplished by integrating the temporal difference methods~\cite{Sutton.2018a} of Q-learning~\cite{Watkins.1992a} and state-action-reward-state-action (SARSA)~\cite{Rummery.1994a} in the learning framework. Therefore, TD-FALCON is a value iteration method that learns action policies and value functions for state-action pairs via temporal difference learning. Briefly, the TD-FALCON dynamics are as follows.

\textbf{Prediction.} For a given state $\bm{s}$, the value function of all actions in the set of actions is predicted by setting the inputs to TD-FALCON's $\text{F}_\text{1}^\text{s}$, $\text{F}_\text{1}^\text{a}$, and $\text{F}_\text{1}^\text{r}$ to $\bm{x}^s = \bm{s}$, $\bm{x}^a = \bm{a}$ and $\bm{x}^r = \Vec{\bm{1}}$, respectively. The action vector $\bm{a}$ is such that $a_I=1$ and $a_i=0$ for $i\neq I$, when taking action $I$. A node~$J$ is then selected via a WTA competition (node~$J$ maximizes Eq.~(\ref{Eq:fusion_1}) in Sec.~\ref{Sec:Fusion_ART}) for each action.

\textbf{Action selection policy.} The $\text{F}_\text{1}^\text{r}$ layer activities, given by
\begin{equation}
\bm{y}^{(F_1^r)} = \bm{x}^r \wedge \bm{w}_J^r = \bm{w}_J^r,
\end{equation}
\noindent are then used to compute the Q-values
\begin{equation}
Q(\bm{s}, \bm{a}) = \dfrac{y_1^{(F_1^r)}}{\sum\limits_{i=1}^m y_i^{(F_1^r)}}.
\end{equation}

An action is then chosen using either a decay $\epsilon$-greedy or a softmax policy, in order to address the exploration-exploitation trade-off. The agent is now in a new state~$\bm{s}'$. 

\textbf{Learning.} Finally the system acts, receives a feedback from the environment and learns using the state ($\bm{x}^s = \bm{s}$), action ($\bm{x}^a = \bm{a}$), and reward ($\bm{x}^r = [Q(\bm{s},\bm{a}), 1-Q(\bm{s},\bm{a})] $) triad. The value function used in $\bm{x}^r$ is estimated using
\begin{equation}
Q(\bm{s},\bm{a}) = Q(\bm{s},\bm{a}) + \Delta Q(\bm{s},\bm{a}) 
\end{equation}
\noindent where
\begin{equation}
\Delta Q(\bm{s},\bm{a}) = \alpha e_{TD},
\end{equation}
\noindent $e_{TD}$ is the temporal difference error and $\alpha$ is the learning rate. Particularly, the TD error for Q-learning (off-policy) is
\begin{equation}
e_{TD} = r + \gamma\max_{a'} Q(\bm{s}',\bm{a}') - Q(\bm{s},\bm{a}),    
\end{equation}
while the TD error for SARSA (on-policy) is
\begin{equation}
e_{TD} = r + \gamma Q(\bm{s}',\bm{a}') - Q(\bm{s},\bm{a}),    
\end{equation}
\noindent where $r$ is the immediate feedback and $\gamma \in [0,1]$ is the discount factor. Additionally, TD-FALCON incorporates self-scaling (Q-values $\in [0,1]$) by using
\begin{equation}
\Delta Q(\bm{s},\bm{a}) = \alpha e_{TD} \left(1 - Q(\bm{s},\bm{a}) \right).
\end{equation}

TD-FALCON trades faster learning for a less compact network (category proliferation), compared gradient-based reinforcement learning approaches, in which the training process is considerably slower but have a smaller network complexity or memory footprint (i.e., less neurons). One of the limitations of TD-FALCON is the bounded Q-values in the range $[0,1]$, which restricts the classes of problems that it can tackle. 

\subsection{Unified ART}
\label{Sec:UART}

The unified ART~\cite{seiffertt2010} is an ART model designed for mixed-modality learning, so that it seamlessly switches among the canonical machine learning modalities (UL, SL and RL). An important characteristic of this integration is the weight sharing between modalities. It uses a Markov Decision Process and Q-learning framework, and it has found application, for instance, in the field of situation awareness~\cite{Brannon.2006a, brannon2009}. 

Briefly, the unified ART consists of a fuzzy ART module (Sec.~\ref{Sec:FA}) and a controller. The latter is represented by a matrix $\bm{V} = [v_{ij}]_{N \times m}$, whose entries $v_{ij}$ estimate value functions, where $N$ and $m$ are the number of categories and available actions, respectively. 

\textbf{Prediction.} Upon presentation of an input $\bm{s}$, the fuzzy ART dynamics are performed. If an uncommitted category is selected, then the controller's matrix $\bm{V}$ need to be expanded accordingly. 

\textbf{Action selection policy.} After the output activity $\bm{y}^{(F_2)}$ of layer \Fout\ is established, it is used to select an action $I$ such that
\begin{equation}
I = \argmax\limits_{1 \leq i \leq m} \left( a_i \right).
\label{Eq:UART_1}
\end{equation}
\noindent where
\begin{equation}
\bm{a} = \bm{y}^{(F_2)~T}\bm{V} = [a_1 ... a_m].
\label{Eq:UART_2}
\end{equation}

The output activity is binary and defined using Eq.~(\ref{Eq:FA_7}) in Sec.~\ref{Sec:FA} when in WTA mode. Alternately, to reduce category proliferation, the output activity can be defined in the distributed mode by setting $y_j^{(F_2)} = T_j$, where the activation functions are computed using Eq.~(\ref{Eq:FA_1}).

\textbf{Learning.} After undertaking the selected action, the environment transitions to the next state $\bm{s}'$, and learning proceeds according to the type of signal received. Assuming WTA mode with resonant node $J$:

\begin{itemize}
\item Supervised signal: this signal has the highest priority. If the correct action was selected, then the controller learns as
\begin{equation}
v_{J,i} =     
\begin{cases} 
v_{max} , & \mbox{if } i=I\\
0, & \mbox{otherwise} \\
\end{cases},
\label{Eq:UART_3}
\end{equation}
\noindent where $v_{max}$ is the maximum allowable value. Otherwise, a mismatch triggers a search for a new resonant neuron within the fuzzy ART module. 
\item Reinforcement signal: In case of a reward, the controller learns as
\begin{equation}
v_{J,I} = v_{J,I} + \alpha r,
\label{Eq:UART_4}
\end{equation}
\noindent where $\alpha$ is a learning rate. Conversely, a penalty causes a mismatch in the fuzzy ART module, which then initiates a search for a new resonant node. The controller still learns using Eq.~(\ref{Eq:UART_4}).

\item Unsupervised signal: this scenario corresponds to the absence of a signal. No learning takes place in the controller.
\end{itemize}

Note that, for all signal types, when a resonant neuron is found within the fuzzy ART module, it is adapted according to the fast learning mode described in Sec.~\ref{Sec:FA}. 

\subsection{Extended unified ART}
\label{Sec:EUART}

The extended unified ART~\cite{seiffertt2010} is another fuzzy ART-based model designed to perform mixed-modality learning, which is accomplished via layered, modality-dependent, vigilance tests. These multiple vigilance criteria must be simultaneously satisfied for the ART system to enter a resonant state and ensue learning. Particularly, this model encodes the states in fuzzy ART's weight matrix $\bm{W}= [w_{i,j}]_{N \times n}$, and the value functions of the state-action pairs in both the critic's matrix $\bm{V} = [v_{i,j}]_{N \times m}$ and the actor's matrix $\bm{U} = [u_{i,j}]_{N \times m}$ (whose role is akin to ARTMAP's map field matrix $\bm{W}^{ab}$ (Sec.~\ref{Sec:AM})), where $N$ is the number of categories, $n$ is the dimension of the state space and $m$ is the number of available actions. Uncommitted nodes are initialized by augmenting $\bm{W}$ with a row equal to $\Vec{\bm{1}}$, while $\bm{U}$ and $\bm{V}$ are expanded with row vectors containing small random values.

\textbf{Prediction.} Upon arriving at a state $\bm{s}$, the highest active node $J$ is found following fuzzy ART's dynamics (Sec.~\ref{Sec:FA}) using the choice-by-difference activation function (Eq.~(\ref{Eq:defAM_1}) in Sec.~\ref{Sec:defAM}).

\textbf{Action selection policy.} An action is selected using
\begin{equation}
I = \argmax\limits_{1 \leq i \leq m} \left( u_{J,i} \right),
\label{Eq:EUART_1}
\end{equation}
\noindent where $\bm{u}_J$ is the $J^{th}$ row of $\bm{U}$.

\textbf{Learning.} After performing the chosen action, the environment evolves to the next state $\bm{s}'$ following its dynamics; vigilance tests and learning are then ensued in consonance with the type of signal feedback from the environment. Particularly, in unsupervised learning mode, the extended unified ART learning dynamics are akin to fuzzy ART's, where there exists only a single match function $M_J^{UL}$ (Eq.~(\ref{Eq:FA_3})) and a corresponding unsupervised vigilance test and parameter $\rho_{UL}$. In this learning mode, neither the actor nor the critic are updated. In reinforcement learning mode, besides the unsupervised vigilance test, a reinforcement vigilance test is performed, where the match function $M_J^{RL}$ is equal to the temporal difference error (Sec.~\ref{Sec:TDFALCON}) computed using $\bm{V}$; if satisfied ($M_J^{RL} > \rho_{RL}$, where $\rho_{RL} \geq 0$ is the reinforcement learning vigilance parameter), then the actor is updated as
\begin{equation}
u_{J,I} = \min \left( u_{J,I} + \alpha r, u_{max} \right),
\label{Eq:EUART_2}
\end{equation}
\noindent where $u_{max}$ is the upper bound for any entry of $\bm{U}$, and the critic is updated using Eq.~(\ref{Eq:UART_4}). If the RL test is not satisfied, a mismatch occurs, and a new search is triggered for the next highest ranking category. This process is repeated until a category satisfies the UL vigilance test while also being associated with an action (Eq.~(\ref{Eq:EUART_1})) that is different from the one taken at $\bm{s}$ (i.e., $i \neq I$), or a new category is created. Finally, supervised learning mode adds a second match function $M_J^{SL}$ on top of the unsupervised one. The former is akin to default ARTMAP's (Sec.~\ref{Sec:defAM}) and assesses if the action taken was the correct one. In the affirmative case, only the actor is updated,
\begin{equation}
u_{J,i} =     
\begin{cases} 
u_{max} , & \mbox{if } i=I\\
0, & \mbox{otherwise} \\
\end{cases},
\label{Eq:EUART_3}
\end{equation}
\noindent whereas in the negative case, a match tracking procedure (MT-)~\cite{Carpenter1998b} slightly decreases fuzzy ART's baseline vigilance parameter during this input presentation cycle, and the search restarts. Note that in all learning modes, when a category is allowed to learn, it does so by  following fuzzy ART's learning dynamics (Sec.~\ref{Sec:FA}).

\section{Advantages of ART}
\label{Sec:Advantages}

\subsection{Speed} 
\label{Sec:Speed}

One of the main advantages of ART neural network architectures is the speed with which they can process data and the relatively small number of epochs they typically require to converge. This is combined with the fact that they can be operated entirely in an online mode, which makes them very effective when working with streaming data or datasets that are too large to fit entirely in memory.

Particularly, the ART 1 (Sec.~\ref{Sec:ART1}) and fuzzy ART (Sec.~\ref{Sec:FA}) neural networks only require an amount of work linear in the number $N$ of samples in the dataset per epoch, and the amount of work performed for each input sample presentation is similarly linear in the number of features $d$ in the dataset, and the number of category templates $k$, that this sample is compared against. This leads to a running time complexity of $\mathcal{O}(Ndk)$, which means that the running time will grow linearly with the growth of any of these variables when the remaining variables are constant. In the absolute worst case, when each sample is put in its own category, this running time degrades to $\mathcal{O}(N^2d)$ since \mbox{$k = N$} in this case; although this situation is uncommon. The same running time complexity analysis applies to other ART neural architectures that faithfully follow the same learning algorithm. A thorough discussion of fuzzy ART computational complexity analysis was presented in~\cite{Granger1998}, and summarized in other studies such as~\cite{Majeed.2018a, Meng2014, meng2015}.

\subsection{Configurability}
\label{Sec:Config}

Another one of ART's main advantages is its ease of configurability~\cite{wunsch2009}. For many unsupervised learning ART neural architectures, the most influential parameter is the vigilance value $\rho$, which controls when resonance occurs between an input sample and a category and subsequently whether this category would be allowed to learn the sample or not. In this way, the ART architectures do not require the choice of the number of clusters when used as clustering algorithms, unlike many other clustering algorithms. Meanwhile, the choice of which ART architecture to use and the choice of a reasonable vigilance value can allow the discovery of many useful clusters without needing to tweak many sensitive parameter values.

\subsection{Explainability}
\label{Sec:Interp}

The way that ART builds well-behaved templates representing the categories it learns from the data is another one of its core strengths~\cite{wunsch2009}. After sufficient learning has taken place, these templates can provide the ability to interpret the results of the neural network learning~\cite{Carpenter1995a, Tan1997a, Healy.2006a} and to visualize the boundaries of each discovered category or clusters. This property is an invaluable one, since many other types of neural networks can only be used as a black-box component that cannot be explained or interpreted.

\subsection{Parallelization and hardware implementation}
\label{Sec:Parallel}

Another major strength of ART neural networks is their potential for massive parallelism and hardware implementation~\cite{wunsch2009}. Notably, early contributions include optoelectronics~\cite{Wunsch.1991a, Caudell.1992a, Wunsch1993, Blume.1995a}, analog~\cite{Ho.1994a} and VLSI~\cite{Tsay.1991a, Gotarredona.1996a, Gotarredona.1998a} systems and, more recently, an implementation in memristive hardware~\cite{Versace.2012a}. Although ART networks are incremental learners, and thus suffer from ordering effects (Sec.~\ref{Sec:ordering_effects}), the calculation of the match and activation function for each category can easily be done in parallel. Thus, ART models lend themselves well to GPU implementations, e.g., fuzzy ART in~\cite{Martinez.2007a, Martinez.2009a}, fuzzy ARTMAP in~\cite{Martinez.2011a} and ARTtree in~\cite{sejun2011}. This offers the opportunity for lower cost, energy consumption and memory footprint than other neural networks' hardware while maintaining online learning capabilities. 

\section{ART challenges and open problems}
\label{Sec:open_problems}

\subsection{Input order dependency}
\label{Sec:ordering_effects}

An important problem faced by all agglomerative clustering or incremental learning algorithms, including ART, is order-dependence of data presentation. This is especially true in fast online learning mode. Many approaches have been developed to mitigate such ordering effects, and they mostly consist of suitable pre- and post-processing strategies (c.f.~\cite{leonardo2018} and the references cited within). Particularly, for supervised ART models, these strategies include Max-Min clustering~\cite{gonzalez1974} in~\cite{dagher1998, dagher1999}; genetic algorithms~\cite{Eiben.2015} in~\cite{palaniappan2009, baek2014}; uncorrelated feature-based ordering in~\cite{oong2014}; featural biasing in~\cite{Carpenter.2010a}; and voting strategies in~\cite{carpenter1992, williamson1996, Carpenter1998b, Lim2000a, Lim2000b, Carpenter2003, Amis2007, Amis.2010a}. In regards to unsupervised ART models, examples of strategies are split, merge and delete operations in~\cite{lughofer2008}; merging heuristics in~\cite{isawa2008, Isawa2008b, Isawa2009}; cluster validity index-based vigilance tests in~\cite{leonardo2017}; and exploiting the ordering properties of visual assessment of cluster tendency (VAT)~\cite{bezdek2002,Bezdek2017} in~\cite{leonardo2018}. The presentation order of inputs still remains an open problem (even if there is meaningful temporal information embedded in the order of sample presentation (e.g., a time series) and it is much more pronounced when presentation is done in a random order), thus requiring further investigation.

\subsection{Vigilance parameter adaptation}
\label{Sec:vig_adapt}

The vigilance is the single most important parameter in any ART model. Selecting suitable values is critical to the network performance and complexity, especially in clustering applications. However, it is often set empirically in an ad hoc manner. In unsupervised learning mode, vigilance adaptation has been addressed in fuzzy ART through the activation maximization, confliction minimization and hybrid integration rules~\cite{lei2013, meng2015}; the combination with particle swarm optimization~\cite{Kennedy.1995a} and cluster validity indices~\cite{xu2009} in~\cite{smith2015}; defining the vigilance as a function of the category size~\cite{Isawa2008b, Isawa2009}; or modeling it as a fuzzy membership function~\cite{Majeed.2018a}. Despite these contributions, setting the vigilance parameter still remains a challenging task worthy of further exploration, particularly in the online learning mode.

\subsection{New metrics}
\label{Sec:metrics}

Another challenging area in the development of ART neural networks is the use of new metrics and representations that would allow ART to more robustly solve some domain-specific problems~\cite{wunsch2009}, such as grammar inference and natural language processing~\cite{Meuth2009}. Some cases require customized neural network designs, such as when the data structure is neither binary nor continuous-valued vectors or when the data has many categorical attributes with large sets of possible values for each attribute. (Notably, mixed-type data is addressed in~\cite{lam2015} in the context of unsupervised feature extraction). In such general cases, it would be highly desirable to have ART models that can deal with this data in its native form without requiring transformations while still maintaining the desirable properties that hold for many existing ART models. 

Different activation functions can endow ART-based systems with new and improved capabilities to tailor the function according to the application. Approaches discussed in~\cite{Lavoie1999}  include making the activation function a function of additional parameters (e.g., vigilance and time), defining individual activation functions for each category and dynamically varying the parameters between epochs without resetting the weights. All these modifications do not change the dynamics of the standard model; although changing the activation function implies changing the search order among the categories. Additionally, there have been some attempts at combining ART with evolutionary computing approaches and hyper-heuristics to achieve this goal (c.f.~\cite{elnabarawy2017} and the references cited within), but there remain many challenges and opportunities to be addressed in this area.

\subsection{Distributed representations}
\label{Sec:code_rep}

The winner-take-all category selection process used in the majority of ART architectures can sometimes lead to category proliferation and is one of the limiting factors of ART's capacity for mapping complex relations \cite{Hernandez.2003a,wunsch2009}. Extending the capabilities of many ART architectures toward distributed representations would lead to greater representational power for these architectures and allow them to encode more complex templates. However, the challenging aspect of this process is to maintain the desirable speed and stability of those ART systems in the presence of this distributed representation. There are examples of architectures that use distributed representations (see Tables~\ref{Tab:BasicARTs} and \ref{Tab:ARTs_SL}), especially in supervised learning, however there are still many issues to be investigated.

\subsection{Dichotomy of match- and error-based learning}
\label{Sec:learning_dichotomy}

In~\cite{wunsch2009} the conjecture is made that the dichotomy of match-based learning (i.e., Hebbian learning and ART) and error-based learning (i.e., using backpropagation in feed-forward neural networks such as deep learning architectures) is likely a false one. This still lacks a definitive resolution.  Some contributions combined the use of match-based and error-based learning such as~\cite{Izquierdo2001, ChunSu2002, ChunSu2005} by using gradient methods to optimize some of the ART parameters. However, the problem of building a system that can do both match- and error-based learning like animals appear to be capable of remains a more complex and interesting challenge, but it holds great promise for much more stable and effective machine learning. In biology, there are clear examples of learning that can happen quickly under the right circumstances, implying match-based learning, as well as incrementally improving through supervised or reinforcement learning in a way that implies error-based learning. The ability to master both types of learning and resolve this conjecture is believed to be a gateway to building machine learning systems that are fast and stable, possessing the ability for life-long learning and being resilient in the face of unpredictable changes in the environment.

\section{Code repositories}
\label{Sec:Resources}

A list of publicly available online source code/repositories is provided below: 
\begin{itemize}
\item \url{github.com/ACIL-Group}
\item \url{techlab.bu.edu/main/article/software}
\item \url{ntu.edu.sg/home/asahtan/downloads.htm}
\item \url{http://www2.imse-cnm.csic.es/~bernabe}
\item \url{ee.bgu.ac.il/~boaz/software.html}
\item \url{libtopoart.eu}
\end{itemize}

\section{Conclusions}
\label{Sec:Conclusion}

This survey presents an overview of ART models used to perform unsupervised learning (a.k.a. clustering), classification, regression and reinforcement learning tasks. It provides a description for each model focusing on the motivation behind their designs, their dynamics, as well as key characteristics such as their code representation and long-term memory. Advantages of ART are discussed as well as open problems. Although mature, the field has room to grow and is still full of opportunities. 

\section*{Acknowledgment}

This research was sponsored by the Missouri University of Science and Technology Mary K. Finley Endowment and Intelligent Systems Center; the Coordena\c{c}\~{a}o de Aperfei\c{c}oamento de Pessoal de N\'{i}vel Superior - Brazil (CAPES) - Finance code BEX 13494/13-9; the Army Research Laboratory (ARL) and the Lifelong Learning Machines program from DARPA/MTO, and it was accomplished under Cooperative Agreement Number W911NF-18-2-0260. The views and conclusions contained in this document are those of the authors and should not be interpreted as representing the official policies, either expressed or implied, of the Army Research Laboratory or the U.S. Government. The U.S. Government is authorized to reproduce and distribute reprints for Government purposes notwithstanding any copyright notation herein.

\biboptions{numbers,sort&compress}
\bibliography{mybibfile}

\end{document}